\RequirePackage{fix-cm}

\documentclass{article} 
\usepackage{iclr2027_conference,times}
\usepackage{amsmath,amsfonts,bm}

\usepackage{xcolor}
\definecolor{sentenceblue}{RGB}{0,0,160}

\newcommand{\figleft}{{\em (Left)}}
\newcommand{\figcenter}{{\em (Center)}}
\newcommand{\figright}{{\em (Right)}}
\newcommand{\figtop}{{\em (Top)}}
\newcommand{\figbottom}{{\em (Bottom)}}
\newcommand{\captiona}{{\em (a)}}
\newcommand{\captionb}{{\em (b)}}
\newcommand{\captionc}{{\em (c)}}
\newcommand{\captiond}{{\em (d)}}

\newcommand{\newterm}[1]{{\bf #1}}

\def\figref#1{figure~\ref{#1}}
\def\Figref#1{Figure~\ref{#1}}
\def\twofigref#1#2{figures \ref{#1} and \ref{#2}}
\def\quadfigref#1#2#3#4{figures \ref{#1}, \ref{#2}, \ref{#3} and \ref{#4}}
\def\secref#1{section~\ref{#1}}
\def\Secref#1{Section~\ref{#1}}
\def\twosecrefs#1#2{sections \ref{#1} and \ref{#2}}
\def\secrefs#1#2#3{sections \ref{#1}, \ref{#2} and \ref{#3}}
\def\eqref#1{equation~\ref{#1}}
\def\Eqref#1{Equation~\ref{#1}}
\def\plaineqref#1{\ref{#1}}
\def\chapref#1{chapter~\ref{#1}}
\def\Chapref#1{Chapter~\ref{#1}}
\def\rangechapref#1#2{chapters\ref{#1}--\ref{#2}}
\def\algref#1{algorithm~\ref{#1}}
\def\Algref#1{Algorithm~\ref{#1}}
\def\twoalgref#1#2{algorithms \ref{#1} and \ref{#2}}
\def\Twoalgref#1#2{Algorithms \ref{#1} and \ref{#2}}
\def\partref#1{part~\ref{#1}}
\def\Partref#1{Part~\ref{#1}}
\def\twopartref#1#2{parts \ref{#1} and \ref{#2}}

\def\ceil#1{\lceil #1 \rceil}
\def\floor#1{\lfloor #1 \rfloor}
\def\1{\bm{1}}
\newcommand{\train}{\mathcal{D}}
\newcommand{\valid}{\mathcal{D_{\mathrm{valid}}}}
\newcommand{\test}{\mathcal{D_{\mathrm{test}}}}

\def\eps{{\epsilon}}

\def\reta{{\textnormal{$\eta$}}}
\def\ra{{\textnormal{a}}}
\def\rb{{\textnormal{b}}}
\def\rc{{\textnormal{c}}}
\def\rd{{\textnormal{d}}}
\def\re{{\textnormal{e}}}
\def\rf{{\textnormal{f}}}
\def\rg{{\textnormal{g}}}
\def\rh{{\textnormal{h}}}
\def\ri{{\textnormal{i}}}
\def\rj{{\textnormal{j}}}
\def\rk{{\textnormal{k}}}
\def\rl{{\textnormal{l}}}
\def\rn{{\textnormal{n}}}
\def\ro{{\textnormal{o}}}
\def\rp{{\textnormal{p}}}
\def\rq{{\textnormal{q}}}
\def\rr{{\textnormal{r}}}
\def\rs{{\textnormal{s}}}
\def\rt{{\textnormal{t}}}
\def\ru{{\textnormal{u}}}
\def\rv{{\textnormal{v}}}
\def\rw{{\textnormal{w}}}
\def\rx{{\textnormal{x}}}
\def\ry{{\textnormal{y}}}
\def\rz{{\textnormal{z}}}

\def\rvepsilon{{\mathbf{\epsilon}}}
\def\rvtheta{{\mathbf{\theta}}}
\def\rva{{\mathbf{a}}}
\def\rvb{{\mathbf{b}}}
\def\rvc{{\mathbf{c}}}
\def\rvd{{\mathbf{d}}}
\def\rve{{\mathbf{e}}}
\def\rvf{{\mathbf{f}}}
\def\rvg{{\mathbf{g}}}
\def\rvh{{\mathbf{h}}}
\def\rvu{{\mathbf{i}}}
\def\rvj{{\mathbf{j}}}
\def\rvk{{\mathbf{k}}}
\def\rvl{{\mathbf{l}}}
\def\rvm{{\mathbf{m}}}
\def\rvn{{\mathbf{n}}}
\def\rvo{{\mathbf{o}}}
\def\rvp{{\mathbf{p}}}
\def\rvq{{\mathbf{q}}}
\def\rvr{{\mathbf{r}}}
\def\rvs{{\mathbf{s}}}
\def\rvt{{\mathbf{t}}}
\def\rvu{{\mathbf{u}}}
\def\rvv{{\mathbf{v}}}
\def\rvw{{\mathbf{w}}}
\def\rvx{{\mathbf{x}}}
\def\rvy{{\mathbf{y}}}
\def\rvz{{\mathbf{z}}}

\def\erva{{\textnormal{a}}}
\def\ervb{{\textnormal{b}}}
\def\ervc{{\textnormal{c}}}
\def\ervd{{\textnormal{d}}}
\def\erve{{\textnormal{e}}}
\def\ervf{{\textnormal{f}}}
\def\ervg{{\textnormal{g}}}
\def\ervh{{\textnormal{h}}}
\def\ervi{{\textnormal{i}}}
\def\ervj{{\textnormal{j}}}
\def\ervk{{\textnormal{k}}}
\def\ervl{{\textnormal{l}}}
\def\ervm{{\textnormal{m}}}
\def\ervn{{\textnormal{n}}}
\def\ervo{{\textnormal{o}}}
\def\ervp{{\textnormal{p}}}
\def\ervq{{\textnormal{q}}}
\def\ervr{{\textnormal{r}}}
\def\ervs{{\textnormal{s}}}
\def\ervt{{\textnormal{t}}}
\def\ervu{{\textnormal{u}}}
\def\ervv{{\textnormal{v}}}
\def\ervw{{\textnormal{w}}}
\def\ervx{{\textnormal{x}}}
\def\ervy{{\textnormal{y}}}
\def\ervz{{\textnormal{z}}}

\def\rmA{{\mathbf{A}}}
\def\rmB{{\mathbf{B}}}
\def\rmC{{\mathbf{C}}}
\def\rmD{{\mathbf{D}}}
\def\rmE{{\mathbf{E}}}
\def\rmF{{\mathbf{F}}}
\def\rmG{{\mathbf{G}}}
\def\rmH{{\mathbf{H}}}
\def\rmI{{\mathbf{I}}}
\def\rmJ{{\mathbf{J}}}
\def\rmK{{\mathbf{K}}}
\def\rmL{{\mathbf{L}}}
\def\rmM{{\mathbf{M}}}
\def\rmN{{\mathbf{N}}}
\def\rmO{{\mathbf{O}}}
\def\rmP{{\mathbf{P}}}
\def\rmQ{{\mathbf{Q}}}
\def\rmR{{\mathbf{R}}}
\def\rmS{{\mathbf{S}}}
\def\rmT{{\mathbf{T}}}
\def\rmU{{\mathbf{U}}}
\def\rmV{{\mathbf{V}}}
\def\rmW{{\mathbf{W}}}
\def\rmX{{\mathbf{X}}}
\def\rmY{{\mathbf{Y}}}
\def\rmZ{{\mathbf{Z}}}

\def\ermA{{\textnormal{A}}}
\def\ermB{{\textnormal{B}}}
\def\ermC{{\textnormal{C}}}
\def\ermD{{\textnormal{D}}}
\def\ermE{{\textnormal{E}}}
\def\ermF{{\textnormal{F}}}
\def\ermG{{\textnormal{G}}}
\def\ermH{{\textnormal{H}}}
\def\ermI{{\textnormal{I}}}
\def\ermJ{{\textnormal{J}}}
\def\ermK{{\textnormal{K}}}
\def\ermL{{\textnormal{L}}}
\def\ermM{{\textnormal{M}}}
\def\ermN{{\textnormal{N}}}
\def\ermO{{\textnormal{O}}}
\def\ermP{{\textnormal{P}}}
\def\ermQ{{\textnormal{Q}}}
\def\ermR{{\textnormal{R}}}
\def\ermS{{\textnormal{S}}}
\def\ermT{{\textnormal{T}}}
\def\ermU{{\textnormal{U}}}
\def\ermV{{\textnormal{V}}}
\def\ermW{{\textnormal{W}}}
\def\ermX{{\textnormal{X}}}
\def\ermY{{\textnormal{Y}}}
\def\ermZ{{\textnormal{Z}}}

\def\vzero{{\bm{0}}}
\def\vone{{\bm{1}}}
\def\vmu{{\bm{\mu}}}
\def\vtheta{{\bm{\theta}}}
\def\va{{\bm{a}}}
\def\vb{{\bm{b}}}
\def\vc{{\bm{c}}}
\def\vd{{\bm{d}}}
\def\ve{{\bm{e}}}
\def\vf{{\bm{f}}}
\def\vg{{\bm{g}}}
\def\vh{{\bm{h}}}
\def\vi{{\bm{i}}}
\def\vj{{\bm{j}}}
\def\vk{{\bm{k}}}
\def\vl{{\bm{l}}}
\def\vm{{\bm{m}}}
\def\vn{{\bm{n}}}
\def\vo{{\bm{o}}}
\def\vp{{\bm{p}}}
\def\vq{{\bm{q}}}
\def\vr{{\bm{r}}}
\def\vs{{\bm{s}}}
\def\vt{{\bm{t}}}
\def\vu{{\bm{u}}}
\def\vv{{\bm{v}}}
\def\vw{{\bm{w}}}
\def\vx{{\bm{x}}}
\def\vy{{\bm{y}}}
\def\vz{{\bm{z}}}

\def\evalpha{{\alpha}}
\def\evbeta{{\beta}}
\def\evepsilon{{\epsilon}}
\def\evlambda{{\lambda}}
\def\evomega{{\omega}}
\def\evmu{{\mu}}
\def\evpsi{{\psi}}
\def\evsigma{{\sigma}}
\def\evtheta{{\theta}}
\def\eva{{a}}
\def\evb{{b}}
\def\evc{{c}}
\def\evd{{d}}
\def\eve{{e}}
\def\evf{{f}}
\def\evg{{g}}
\def\evh{{h}}
\def\evi{{i}}
\def\evj{{j}}
\def\evk{{k}}
\def\evl{{l}}
\def\evm{{m}}
\def\evn{{n}}
\def\evo{{o}}
\def\evp{{p}}
\def\evq{{q}}
\def\evr{{r}}
\def\evs{{s}}
\def\evt{{t}}
\def\evu{{u}}
\def\evv{{v}}
\def\evw{{w}}
\def\evx{{x}}
\def\evy{{y}}
\def\evz{{z}}

\def\mA{{\bm{A}}}
\def\mB{{\bm{B}}}
\def\mC{{\bm{C}}}
\def\mD{{\bm{D}}}
\def\mE{{\bm{E}}}
\def\mF{{\bm{F}}}
\def\mG{{\bm{G}}}
\def\mH{{\bm{H}}}
\def\mI{{\bm{I}}}
\def\mJ{{\bm{J}}}
\def\mK{{\bm{K}}}
\def\mL{{\bm{L}}}
\def\mM{{\bm{M}}}
\def\mN{{\bm{N}}}
\def\mO{{\bm{O}}}
\def\mP{{\bm{P}}}
\def\mQ{{\bm{Q}}}
\def\mR{{\bm{R}}}
\def\mS{{\bm{S}}}
\def\mT{{\bm{T}}}
\def\mU{{\bm{U}}}
\def\mV{{\bm{V}}}
\def\mW{{\bm{W}}}
\def\mX{{\bm{X}}}
\def\mY{{\bm{Y}}}
\def\mZ{{\bm{Z}}}
\def\mBeta{{\bm{\beta}}}
\def\mPhi{{\bm{\Phi}}}
\def\mLambda{{\bm{\Lambda}}}
\def\mSigma{{\bm{\Sigma}}}

\DeclareMathAlphabet{\mathsfit}{\encodingdefault}{\sfdefault}{m}{sl}
\SetMathAlphabet{\mathsfit}{bold}{\encodingdefault}{\sfdefault}{bx}{n}
\newcommand{\tens}[1]{\bm{\mathsfit{#1}}}
\def\tA{{\tens{A}}}
\def\tB{{\tens{B}}}
\def\tC{{\tens{C}}}
\def\tD{{\tens{D}}}
\def\tE{{\tens{E}}}
\def\tF{{\tens{F}}}
\def\tG{{\tens{G}}}
\def\tH{{\tens{H}}}
\def\tI{{\tens{I}}}
\def\tJ{{\tens{J}}}
\def\tK{{\tens{K}}}
\def\tL{{\tens{L}}}
\def\tM{{\tens{M}}}
\def\tN{{\tens{N}}}
\def\tO{{\tens{O}}}
\def\tP{{\tens{P}}}
\def\tQ{{\tens{Q}}}
\def\tR{{\tens{R}}}
\def\tS{{\tens{S}}}
\def\tT{{\tens{T}}}
\def\tU{{\tens{U}}}
\def\tV{{\tens{V}}}
\def\tW{{\tens{W}}}
\def\tX{{\tens{X}}}
\def\tY{{\tens{Y}}}
\def\tZ{{\tens{Z}}}

\def\gA{{\mathcal{A}}}
\def\gB{{\mathcal{B}}}
\def\gC{{\mathcal{C}}}
\def\gD{{\mathcal{D}}}
\def\gE{{\mathcal{E}}}
\def\gF{{\mathcal{F}}}
\def\gG{{\mathcal{G}}}
\def\gH{{\mathcal{H}}}
\def\gI{{\mathcal{I}}}
\def\gJ{{\mathcal{J}}}
\def\gK{{\mathcal{K}}}
\def\gL{{\mathcal{L}}}
\def\gM{{\mathcal{M}}}
\def\gN{{\mathcal{N}}}
\def\gO{{\mathcal{O}}}
\def\gP{{\mathcal{P}}}
\def\gQ{{\mathcal{Q}}}
\def\gR{{\mathcal{R}}}
\def\gS{{\mathcal{S}}}
\def\gT{{\mathcal{T}}}
\def\gU{{\mathcal{U}}}
\def\gV{{\mathcal{V}}}
\def\gW{{\mathcal{W}}}
\def\gX{{\mathcal{X}}}
\def\gY{{\mathcal{Y}}}
\def\gZ{{\mathcal{Z}}}

\def\sA{{\mathbb{A}}}
\def\sB{{\mathbb{B}}}
\def\sC{{\mathbb{C}}}
\def\sD{{\mathbb{D}}}
\def\sF{{\mathbb{F}}}
\def\sG{{\mathbb{G}}}
\def\sH{{\mathbb{H}}}
\def\sI{{\mathbb{I}}}
\def\sJ{{\mathbb{J}}}
\def\sK{{\mathbb{K}}}
\def\sL{{\mathbb{L}}}
\def\sM{{\mathbb{M}}}
\def\sN{{\mathbb{N}}}
\def\sO{{\mathbb{O}}}
\def\sP{{\mathbb{P}}}
\def\sQ{{\mathbb{Q}}}
\def\sR{{\mathbb{R}}}
\def\sS{{\mathbb{S}}}
\def\sT{{\mathbb{T}}}
\def\sU{{\mathbb{U}}}
\def\sV{{\mathbb{V}}}
\def\sW{{\mathbb{W}}}
\def\sX{{\mathbb{X}}}
\def\sY{{\mathbb{Y}}}
\def\sZ{{\mathbb{Z}}}

\def\emLambda{{\Lambda}}
\def\emA{{A}}
\def\emB{{B}}
\def\emC{{C}}
\def\emD{{D}}
\def\emE{{E}}
\def\emF{{F}}
\def\emG{{G}}
\def\emH{{H}}
\def\emI{{I}}
\def\emJ{{J}}
\def\emK{{K}}
\def\emL{{L}}
\def\emM{{M}}
\def\emN{{N}}
\def\emO{{O}}
\def\emP{{P}}
\def\emQ{{Q}}
\def\emR{{R}}
\def\emS{{S}}
\def\emT{{T}}
\def\emU{{U}}
\def\emV{{V}}
\def\emW{{W}}
\def\emX{{X}}
\def\emY{{Y}}
\def\emZ{{Z}}
\def\emSigma{{\Sigma}}

\newcommand{\etens}[1]{\mathsfit{#1}}
\def\etLambda{{\etens{\Lambda}}}
\def\etA{{\etens{A}}}
\def\etB{{\etens{B}}}
\def\etC{{\etens{C}}}
\def\etD{{\etens{D}}}
\def\etE{{\etens{E}}}
\def\etF{{\etens{F}}}
\def\etG{{\etens{G}}}
\def\etH{{\etens{H}}}
\def\etI{{\etens{I}}}
\def\etJ{{\etens{J}}}
\def\etK{{\etens{K}}}
\def\etL{{\etens{L}}}
\def\etM{{\etens{M}}}
\def\etN{{\etens{N}}}
\def\etO{{\etens{O}}}
\def\etP{{\etens{P}}}
\def\etQ{{\etens{Q}}}
\def\etR{{\etens{R}}}
\def\etS{{\etens{S}}}
\def\etT{{\etens{T}}}
\def\etU{{\etens{U}}}
\def\etV{{\etens{V}}}
\def\etW{{\etens{W}}}
\def\etX{{\etens{X}}}
\def\etY{{\etens{Y}}}
\def\etZ{{\etens{Z}}}

\newcommand{\pdata}{p_{\rm{data}}}
\newcommand{\ptrain}{\hat{p}_{\rm{data}}}
\newcommand{\Ptrain}{\hat{P}_{\rm{data}}}
\newcommand{\pmodel}{p_{\rm{model}}}
\newcommand{\Pmodel}{P_{\rm{model}}}
\newcommand{\ptildemodel}{\tilde{p}_{\rm{model}}}
\newcommand{\pencode}{p_{\rm{encoder}}}
\newcommand{\pdecode}{p_{\rm{decoder}}}
\newcommand{\precons}{p_{\rm{reconstruct}}}

\newcommand{\laplace}{\mathrm{Laplace}} 

\newcommand{\E}{\mathbb{E}}
\newcommand{\Ls}{\mathcal{L}}
\newcommand{\R}{\mathbb{R}}
\newcommand{\emp}{\tilde{p}}
\newcommand{\lr}{\alpha}
\newcommand{\reg}{\lambda}
\newcommand{\rect}{\mathrm{rectifier}}
\newcommand{\softmax}{\mathrm{softmax}}
\newcommand{\sigmoid}{\sigma}
\newcommand{\softplus}{\zeta}
\newcommand{\KL}{D_{\mathrm{KL}}}
\newcommand{\Var}{\mathrm{Var}}
\newcommand{\standarderror}{\mathrm{SE}}
\newcommand{\Cov}{\mathrm{Cov}}
\newcommand{\normlzero}{L^0}
\newcommand{\normlone}{L^1}
\newcommand{\normltwo}{L^2}
\newcommand{\normlp}{L^p}
\newcommand{\normmax}{L^\infty}

\newcommand{\parents}{Pa} 

\DeclareMathOperator*{\argmax}{arg\,max}
\DeclareMathOperator*{\argmin}{arg\,min}

\DeclareMathOperator{\sign}{sign}
\DeclareMathOperator{\Tr}{Tr}
\let\ab\allowbreak

\usepackage{hyperref}
\hypersetup{hidelinks}
\usepackage{url}
\usepackage{longtable,booktabs,array}
\usepackage{graphicx}
\usepackage[outline]{contour}
\usepackage{wrapfig}
\usepackage{needspace}
\usepackage{tcolorbox}
\tcbuselibrary{listings,breakable,skins}
\newtcblisting{optdprompt}[1]{%
  enhanced,breakable,listing only,
  colback=white,colframe=black,colbacktitle=black,coltitle=white,
  boxrule=.6pt,arc=2mm,outer arc=2mm,
  title={#1},title after break={#1\ (continued)},
  fonttitle=\bfseries\small,
  boxsep=0pt,left=8pt,right=8pt,top=7pt,bottom=7pt,
  toptitle=4pt,bottomtitle=4pt,before skip=9pt,after skip=9pt,
  listing options={%
    basicstyle=\ttfamily\fontsize{7}{8.4}\selectfont,
    columns=fullflexible,keepspaces=true,breaklines=true,
    breakatwhitespace=false,showstringspaces=false,tabsize=2,
    aboveskip=0pt,belowskip=0pt,numbers=none,
    postbreak=\mbox{\textcolor{gray}{$\hookrightarrow$}\space}}}
\makeatletter
\newcommand{\finishwraptable}{%
  \par
  \ifnum\c@WF@wrappedlines>1
    \vskip\dimexpr\baselineskip*\c@WF@wrappedlines-\baselineskip\relax
  \fi
  \WFclear
}
\makeatother
\usepackage{placeins}

\usepackage{amssymb}
\usepackage{tikz,pgfplots,fontawesome5}
\usetikzlibrary{arrows.meta,calc}
\pgfplotsset{compat=1.18}
\DeclareUnicodeCharacter{2011}{\mbox{-}}

\title{\normalfont\bfseries Right Answers, Costly Models: The Efficiency Gap in LLM-based Optimization Modeling}

\author{%
  \begin{minipage}[t]{\dimexpr\textwidth-2\tabcolsep\relax}
  \centering\normalfont
  \setlength{\parskip}{0pt}
  {\fontsize{11}{13}\selectfont
  \textbf{Zhong Li}\textsuperscript{1*}\quad
  \textbf{Xin Huang}\textsuperscript{2*}\\[2pt]
  \textbf{Jinhui Wan}\textsuperscript{2}\quad
  \textbf{Xiangyi Wang}\textsuperscript{2}\quad
  \textbf{Shenkai Zhang}\textsuperscript{3}\\[2pt]
  \textbf{Ruiqi Chen}\textsuperscript{3}\quad
  \textbf{Wenyu Liu}\textsuperscript{4}\quad
  \textbf{Zaiwen Wen}\textsuperscript{4(\faIcon[regular]{envelope})}\quad
  \textbf{Ziyan Luo}\textsuperscript{2(\faIcon[regular]{envelope})}\endgraf}
  \vspace{5pt}
  {\normalsize
  \textsuperscript{1}Great Bay University\quad
  \textsuperscript{2}Beijing Jiaotong University\\
  \textsuperscript{3}Beihang University\quad
  \textsuperscript{4}Peking University\endgraf}
  \vspace{4pt}
  {\small
  \textsuperscript{(\faIcon[regular]{envelope})}Corresponding authors:
  \href{mailto:wenzw@pku.edu.cn}{\textbf{\texttt{wenzw@pku.edu.cn}}} (Zaiwen Wen);\\
  \href{mailto:zyluo@bjtu.edu.cn}{\textbf{\texttt{zyluo@bjtu.edu.cn}}} (Ziyan Luo)\endgraf}
  \vspace{2pt}
  {\normalsize\textsuperscript{*}Equal contributions.\endgraf}
  \end{minipage}%
}
\hypersetup{
  pdftitle={Right Answers, Costly Models: The Efficiency Gap in LLM-based Optimization Modeling},
  pdfauthor={Zhong Li, Xin Huang, Jinhui Wan, Xiangyi Wang, Shenkai Zhang, Ruiqi Chen, Wenyu Liu, Zaiwen Wen, Ziyan Luo}
}

\newcommand{\fix}{\marginpar{FIX}}
\newcommand{\new}{\marginpar{NEW}}

\iclrfinalcopy
\begin{document}

\AddToHookNext{shipout/foreground}{%
  \begin{tikzpicture}[remember picture,overlay]
    \node[anchor=center,inner sep=0pt]
      at ([xshift=\dimexpr1in+\oddsidemargin+.5\textwidth\relax,yshift=-47bp]current page.north west)
      {\makebox[\textwidth][s]{%
        \raisebox{-.5\height}{\includegraphics[width=48bp,height=15bp,keepaspectratio]{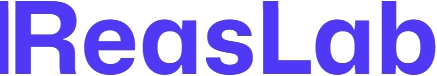}}\hfill
        \raisebox{-.5\height}{\includegraphics[width=52bp,height=18bp,keepaspectratio]{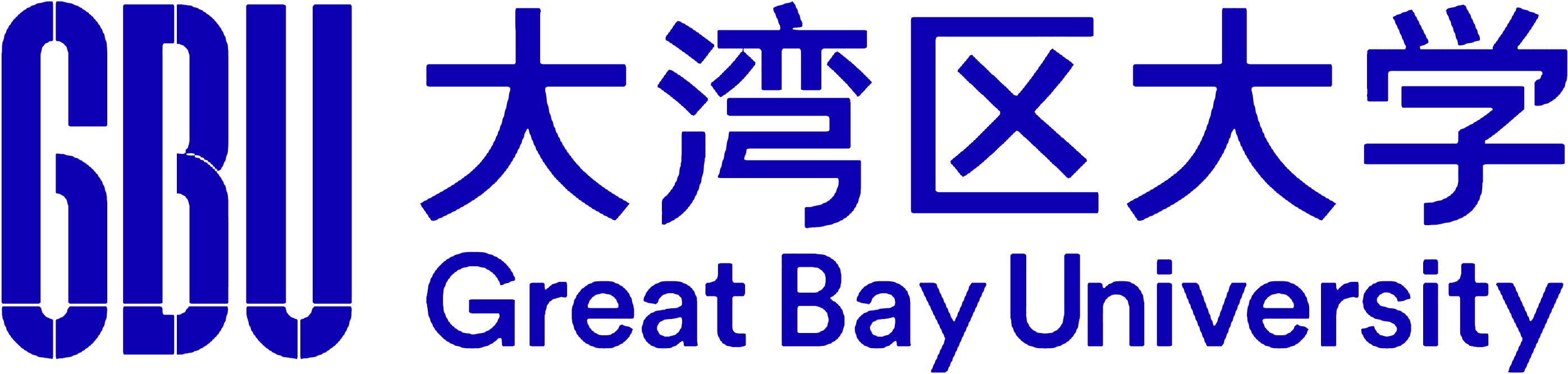}}\hfill
        \raisebox{-.5\height}{\includegraphics[width=56bp,height=18bp,keepaspectratio]{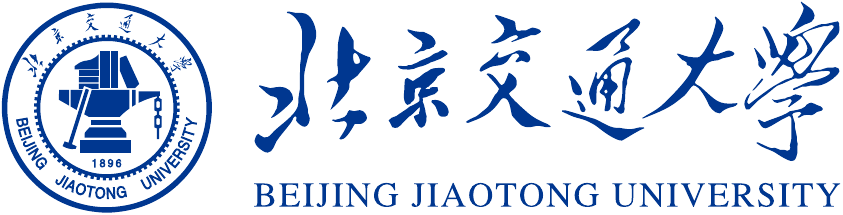}}\hfill
        \raisebox{-.5\height}{\includegraphics[width=62bp,height=18bp,keepaspectratio]{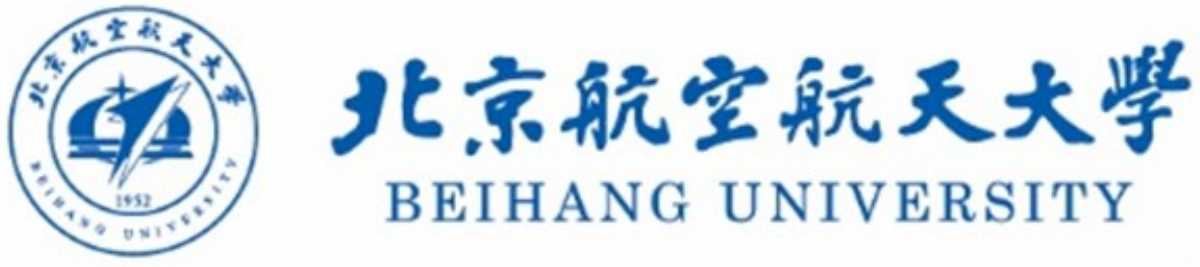}}\hfill
        \raisebox{-.5\height}{\includegraphics[width=62bp,height=18bp,keepaspectratio]{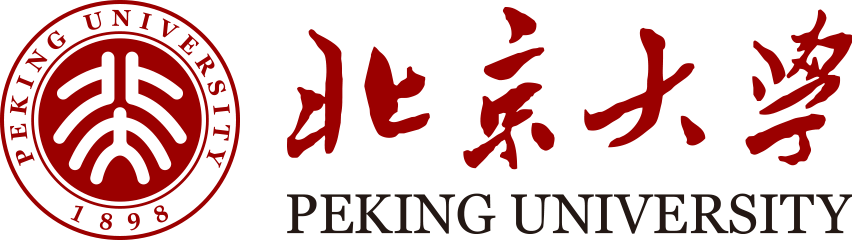}}%
      }};
    \draw[black!65,line width=.4pt]
      ([xshift=\dimexpr1in+\oddsidemargin\relax,yshift=-68bp]current page.north west)
      -- ++(\textwidth,0);
  \end{tikzpicture}%
}

{\raggedright\hyphenpenalty=10000\exhyphenpenalty=10000\maketitle}
\fancyhead{}
\renewcommand{\headrulewidth}{0pt}

\begin{abstract}
Optimization modeling formulates real-world decision problems as mathematical programs that solvers can use to find optimal decisions. Large language models (LLMs) can automate this process, but the resulting correct formulations can require substantial time and memory to construct and solve, limiting practical scalability. Therefore, we systematically investigate whether LLMs can identify problem structure from natural-language descriptions and apply suitable optimization modeling techniques to generate mathematical models and solver code that solve the problems correctly and efficiently. To this end, we first curate \textbf{OptTips}, a knowledge base of 50 expert modeling techniques in eight families. Using this knowledge, we develop \textbf{OptDachshund}, a multi-agent framework that transforms problems from existing optimization benchmarks into new tasks for evaluating LLMs' use of modeling techniques. It constructs conventional and expert mathematical models with solver code for the same task and data, providing baselines for correctness and computational cost. The resulting \textbf{EfficientOpt} benchmark contains 561 expert-reviewed tasks with paired reference implementations. Evaluation of 11 representative LLMs reveals an efficiency gap on correctly solved tasks with comparable measurements: for every LLM, most generated programs take longer to solve than their expert counterparts. Within the comparable reference-size subset, 57\% of programs with correct objective values and fewer variables and linear constraints have longer recorded solver times. Case studies show that different modeling techniques can achieve the same optimal value at similar recorded cost. Faster solving may not reduce execution time if the code takes longer to prepare data and build the model. LLM optimization modeling should therefore be evaluated for both correctness and computational efficiency.
\end{abstract}

\section{Introduction}

Optimization modeling connects real-world decision problems to mathematical solvers \citep{williams2013model}. Applications in scheduling, transportation, energy, and resource allocation depend on translating a problem into variables, an objective, and constraints \citep{pinedo2022scheduling,toth2014vehicle,conejo2010decision}. This translation requires expertise in choosing representations that solvers can handle effectively. LLMs offer a promising route to automating this work \citep{xiao2025survey}. Agent-based approaches coordinate specialized modeling, programming, and verification roles \citep{xiao2024chain,ahmaditeshnizi2024optimus}. Supervised fine-tuning (SFT) teaches models from problem--formulation pairs \citep{huang2025orlm,jiang2025llmopt,lu2025optmath}, while reinforcement learning (RL) uses solver-verified rewards \citep{chen2026sirl} or test-time policy optimization \citep{ding2026orr1} to improve modeling. This progress motivates studying the computational efficiency of LLM-generated models: the time and memory needed to build and solve them. These costs limit problem sizes under practical resource budgets \citep{klotz2013linear,vielma2015formulation}.

Executability and numerical correctness alone do not establish whether a generated optimization model is efficient to solve \citep{zhao2026sage}. The same optimization problem can be addressed through different modeling and solution approaches, with sharply different computational costs \citep{williams2013model,vielma2015formulation}. For example, loose Big-$M$ constants can weaken LP relaxations \citep{vielma2015formulation}; unnecessary integrality requirements can increase branching effort \citep{vanderhulst2026implied}. Unexploited symmetry can cause redundant search \citep{margot2010symmetry}, while separable structure can enable decomposition into coordinated subproblems \citep{dantzig1960decomposition}. Effective optimization modeling therefore requires recognizing problem structure and selecting modeling techniques that exploit it \citep{williams2013model}. However, simply asking an LLM to produce faster code does not guarantee an effective reformulation: efficiency-oriented prompting can introduce unreliable shortcuts or disrupt favorable structure \citep{wang2026formalize}. The central question is therefore: \textbf{Can LLMs recognize problem structure and apply appropriate optimization modeling techniques to obtain correct solutions at lower computational cost?}

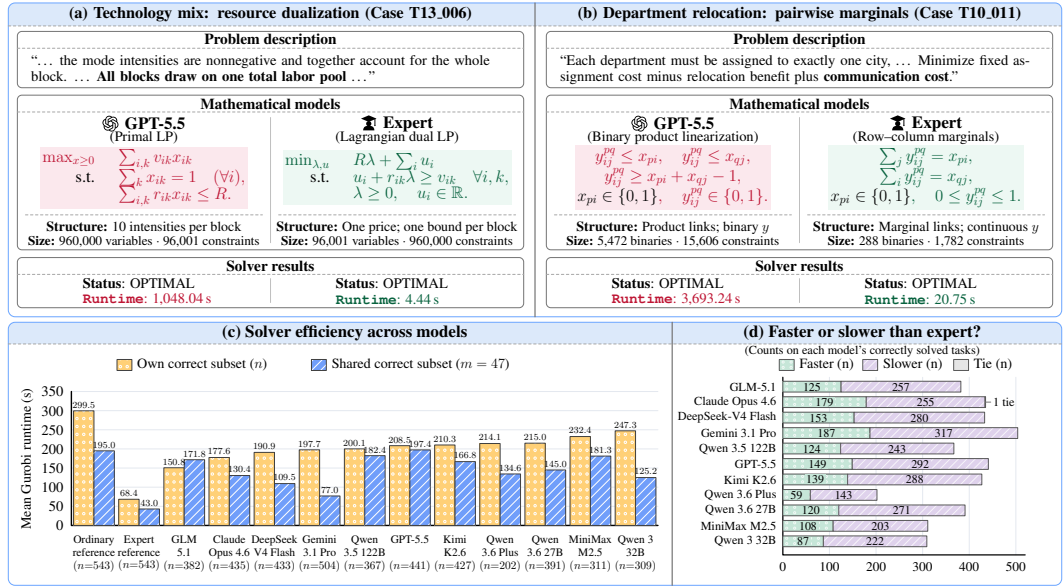
\begin{figure}[h]
\centering
\begingroup
\definecolor{figink}{HTML}{000000}
\definecolor{figblueedge}{HTML}{6AA5FF}
\definecolor{figbluehead}{HTML}{DCE9FA}
\definecolor{figblueink}{HTML}{000000}
\definecolor{figbar}{HTML}{4D72C5}
\definecolor{figfast}{HTML}{C6E6D8}
\definecolor{figtie}{HTML}{E5E5E5}
\definecolor{figslow}{HTML}{DFD2EC}
\definecolor{figgrid}{HTML}{E4E8EF}
\definecolor{figgreenhead}{HTML}{DFF2E7}
\definecolor{figgreenedge}{HTML}{68C98E}
\definecolor{figgreenink}{HTML}{07623E}
\definecolor{figredink}{HTML}{BE1534}
\definecolor{figpalered}{HTML}{FBE8ED}
\definecolor{figpalemint}{HTML}{EDF8F3}
\newcommand{\figfont}[2]{\fontsize{#1}{#2}\selectfont}
\resizebox{\textwidth}{!}{%
\begin{tikzpicture}[x=1cm,y=1cm,every node/.style={text=figink,inner sep=0pt},line cap=round,line join=round]
\path[use as bounding box] (-.32,-2.66) rectangle (32.32,15.38);
\tikzset{lzouterpanel/.style={draw=figblueedge,line width=.85pt,rounded corners=6pt}}
\begin{scope}[yshift=8.30cm]
\begin{scope}
\clip[rounded corners=6pt] (-.28,-2.62) rectangle (32.28,7.08);
\fill[figbluehead] (-.28,6.37) rectangle (32.28,7.08);
\end{scope}
\draw[lzouterpanel] (-.28,-2.62) rectangle (32.28,7.08);
\draw[black!35,line width=.4pt,dash pattern=on 2.2pt off 2.2pt]
  (-.28,6.37)--(32.28,6.37);
\begin{scope}[every node/.style={text=black,inner sep=0pt}]
\newcommand{\lzopenaiicon}[2]{%
  \begin{scope}[shift={(#1,#2)}]
    \path[fill=black,even odd rule]
      (-0.07255,0.08275)
      --(-0.07255,0.14151)
      ..controls (-0.07255,0.14631) and (-0.07015,0.14990)..(-0.06536,0.15230)
      --(0.05097,0.21946)
      ..controls (0.06776,0.22905) and (0.08694,0.23385)..(0.10613,0.23385)
      ..controls (0.17928,0.23385) and (0.22605,0.17629)..(0.22605,0.11513)
      ..controls (0.22605,0.11153) and (0.22605,0.10673)..(0.22605,0.10074)
      --(0.10373,0.17269)
      ..controls (0.09654,0.17629) and (0.08934,0.17629)..(0.08215,0.17269)
      --(-0.07255,0.08275)
      --cycle
      (0.20207,-0.14391)
      --(0.20207,-0.00480)
      ..controls (0.20207,0.00360) and (0.19847,0.00959)..(0.19128,0.01439)
      --(0.03658,0.10433)
      --(0.08694,0.13311)
      ..controls (0.09174,0.13551) and (0.09534,0.13551)..(0.09894,0.13311)
      --(0.21646,0.06476)
      ..controls (0.25004,0.04557) and (0.27282,0.00360)..(0.27282,-0.03718)
      ..controls (0.27282,-0.08395) and (0.24524,-0.12712)..(0.20207,-0.14391)
      --(0.20207,-0.14391)
      --cycle
      (-0.10853,-0.02159)
      --(-0.15890,0.00839)
      ..controls (-0.16250,0.01079) and (-0.16489,0.01439)..(-0.16489,0.01919)
      --(-0.16489,0.15470)
      ..controls (-0.16489,0.21946) and (-0.11453,0.26983)..(-0.04617,0.26983)
      ..controls (-0.02099,0.26983) and (0.00300,0.26143)..(0.02338,0.24584)
      --(-0.09774,0.17509)
      ..controls (-0.10493,0.17149) and (-0.10853,0.16549)..(-0.10853,0.15710)
      --(-0.10853,-0.02159)
      --(-0.10853,-0.02159)
      --cycle
      (0.00060,-0.08395)
      --(-0.07255,-0.04317)
      --(-0.07255,0.04197)
      --(0.00060,0.08275)
      --(0.07255,0.04197)
      --(0.07255,-0.04317)
      --(0.00060,-0.08395)
      --cycle
      (0.04617,-0.27103)
      ..controls (0.02099,-0.27103) and (-0.00300,-0.26263)..(-0.02338,-0.24704)
      --(0.09774,-0.17629)
      ..controls (0.10493,-0.17269) and (0.10853,-0.16669)..(0.10853,-0.15710)
      --(0.10853,0.02039)
      --(0.16010,-0.00959)
      ..controls (0.16369,-0.01199) and (0.16609,-0.01559)..(0.16609,-0.02039)
      --(0.16609,-0.15470)
      ..controls (0.16609,-0.22066) and (0.11453,-0.27103)..(0.04617,-0.27103)
      --(0.04617,-0.27103)
      --cycle
      (-0.09894,-0.13431)
      --(-0.21646,-0.06596)
      ..controls (-0.25004,-0.04677) and (-0.27282,-0.00480)..(-0.27282,0.03598)
      ..controls (-0.27282,0.08275) and (-0.24404,0.12592)..(-0.20087,0.14271)
      --(-0.20087,0.00360)
      ..controls (-0.20087,-0.00600) and (-0.19727,-0.01199)..(-0.19008,-0.01559)
      --(-0.03658,-0.10433)
      --(-0.08694,-0.13431)
      ..controls (-0.09054,-0.13671) and (-0.09534,-0.13671)..(-0.09894,-0.13431)
      --cycle
      (-0.10613,-0.23505)
      ..controls (-0.17569,-0.23505) and (-0.22605,-0.18228)..(-0.22605,-0.11752)
      ..controls (-0.22605,-0.11273) and (-0.22605,-0.10793)..(-0.22485,-0.10313)
      --(-0.10373,-0.17269)
      ..controls (-0.09654,-0.17749) and (-0.08934,-0.17749)..(-0.08215,-0.17269)
      --(0.07255,-0.08395)
      --(0.07255,-0.14271)
      ..controls (0.07255,-0.14750) and (0.07015,-0.15110)..(0.06656,-0.15350)
      --(-0.05097,-0.22066)
      ..controls (-0.06776,-0.23025) and (-0.08574,-0.23505)..(-0.10613,-0.23505)
      --(-0.10613,-0.23505)
      --cycle
      (0.04617,-0.30700)
      ..controls (0.12052,-0.30700) and (0.18288,-0.25424)..(0.19727,-0.18468)
      ..controls (0.26563,-0.16669) and (0.31000,-0.10193)..(0.31000,-0.03718)
      ..controls (0.31000,0.00600) and (0.29081,0.04797)..(0.25843,0.07795)
      ..controls (0.26083,0.09114) and (0.26323,0.10433)..(0.26323,0.11632)
      ..controls (0.26323,0.20507) and (0.19248,0.26983)..(0.10973,0.26983)
      ..controls (0.09294,0.26983) and (0.07735,0.26743)..(0.06176,0.26263)
      ..controls (0.03418,0.28901) and (-0.00420,0.30700)..(-0.04617,0.30700)
      ..controls (-0.12052,0.30700) and (-0.18288,0.25424)..(-0.19607,0.18348)
      ..controls (-0.26563,0.16549) and (-0.31000,0.10193)..(-0.31000,0.03598)
      ..controls (-0.31000,-0.00720) and (-0.29081,-0.04917)..(-0.25843,-0.07915)
      ..controls (-0.26083,-0.09234) and (-0.26323,-0.10433)..(-0.26323,-0.11752)
      ..controls (-0.26323,-0.20507) and (-0.19128,-0.27103)..(-0.10973,-0.27103)
      ..controls (-0.09294,-0.27103) and (-0.07735,-0.26863)..(-0.06056,-0.26383)
      ..controls (-0.03298,-0.29021) and (0.00420,-0.30700)..(0.04617,-0.30700)
      --cycle;
  \end{scope}
}
\newcommand{\lzexperticon}[2]{%
  \begin{scope}[shift={(#1,#2)},scale=.70]
    \fill[black] (-.43,.26)--(0,.44)--(.43,.26)--(0,.08)--cycle;
    \draw[black,line width=.85pt] (.34,.25)--(.34,-.10);
    \fill[black] (.34,-.12) circle (.035);
    \fill[black] (0,-.02) circle (.145);
    \fill[black] (-.31,-.39)..controls(-.30,-.15)and(-.14,-.14)..(0,-.14)
      ..controls(.14,-.14)and(.30,-.15)..(.31,-.39)--cycle;
  \end{scope}
}
\newcommand{\lzmathhighlight}[3]{%
  \begingroup\setlength{\fboxsep}{2pt}%
  \colorbox{#1}{\color{#2}$\textstyle #3$}\endgroup
}
\newcommand{\lzcasesection}[4]{%
  \draw[black!55,line width=.65pt,rounded corners=4pt]
    (#1,#2) rectangle (#1+15.65,#3);
  \draw[black!35,line width=.4pt,dash pattern=on 2.2pt off 2.2pt]
    (#1,#3-.52)--(#1+15.65,#3-.52);
  \node[font=\figfont{13.4}{15}\bfseries] at (#1+7.825,#3-.26) {#4};
}
\foreach \xx in {0,16.35}{
  \lzcasesection{\xx}{4.46}{6.20}{Problem description}
  \lzcasesection{\xx}{-.55}{4.23}{Mathematical models}
  \lzcasesection{\xx}{-2.34}{-.79}{Solver results}
  \draw[black!25,line width=.45pt] (\xx+7.825,-.33)--(\xx+7.825,3.54);
  \draw[black!25,line width=.45pt] (\xx+7.825,-2.18)--(\xx+7.825,-1.44);
  \draw[black!35,line width=.4pt,dash pattern=on 2.2pt off 2.2pt]
    (\xx+.25,.53)--(\xx+7.575,.53);
  \draw[black!35,line width=.4pt,dash pattern=on 2.2pt off 2.2pt]
    (\xx+8.075,.53)--(\xx+15.40,.53);
}
\node[font=\figfont{14.3}{16}\bfseries,align=center,text width=15.35cm] at (7.825,6.725)
  {(a) Technology mix: resource dualization (Case T13\_006)};
\node[font=\figfont{14.3}{16}\bfseries,align=center,text width=15.35cm] at (24.175,6.725)
  {(b) Department relocation: pairwise marginals (Case T10\_011)};
\node[anchor=north west,font=\figfont{12.8}{14.6},text width=14.81cm,align=left]
  at (.42,5.49) {``\ldots\ the mode intensities are nonnegative and together account for the whole block.
  \ldots\ \textbf{All blocks draw on one total labor pool} \ldots''};
\node[anchor=north west,font=\figfont{12.8}{14.6},text width=14.81cm,align=left]
  at (16.77,5.49) {``Each department must be assigned to exactly one city, \ldots\
  Minimize fixed assignment cost minus relocation benefit plus \textbf{communication cost}.''};
\foreach \xx/\role/\icon in {
  3.9125/{GPT-5.5}/\lzopenaiicon,11.7375/{Expert}/\lzexperticon,
  20.2625/{GPT-5.5}/\lzopenaiicon,28.0875/{Expert}/\lzexperticon}{
  \node[font=\figfont{15.0}{17}\bfseries] at (\xx,3.35)
    {\begin{tikzpicture}[baseline=-.65ex,scale=.80]\icon{0}{0}\end{tikzpicture}\hspace{.20cm}\role};
}
\node[font=\figfont{12.6}{14.5}] at (3.9125,2.91) {(Primal LP)};
\node[font=\figfont{12.6}{14.5}] at (11.7375,2.91) {(Lagrangian dual LP)};
\node[font=\figfont{12.6}{14.5}] at (20.2625,2.91) {(Binary product linearization)};
\node[font=\figfont{12.6}{14.5}] at (28.0875,2.91) {(Row--column marginals)};
\node[font=\figfont{14.7}{17.2}] at (3.9125,1.68) {\renewcommand{\arraystretch}{.82}\setlength{\arraycolsep}{3pt}%
\lzmathhighlight{figpalered}{figredink}{\begin{array}{@{}rl@{}}
  \max\nolimits_{x\ge0}\quad &\textstyle\sum_{i,k}v_{ik}x_{ik}\\
  \text{\color{black}s.t.}\quad &\textstyle\sum_k x_{ik}=1\quad(\forall i),\\
  &\textstyle\sum_{i,k}r_{ik}x_{ik}\le R.
\end{array}}};
\node[font=\figfont{14.7}{17.2}] at (11.7375,1.68) {\renewcommand{\arraystretch}{.82}\setlength{\arraycolsep}{3pt}%
\lzmathhighlight{figpalemint}{figgreenink}{\begin{array}{@{}rl@{}}
  \min\nolimits_{\lambda,u}\quad &\textstyle R\lambda+\sum_i u_i\\
  \text{\color{black}s.t.}\quad &u_i+r_{ik}\lambda\ge v_{ik}\quad\forall i,k,\\
  &\lambda\ge0,\quad u_i\in\mathbb{R}.
\end{array}}};
\node[font=\figfont{14.7}{17.2}] at (20.2625,1.68) {\renewcommand{\arraystretch}{.82}\setlength{\arraycolsep}{3pt}%
\lzmathhighlight{figpalered}{figredink}{\begin{array}{@{}c@{}}
  y^{pq}_{ij}\le x_{pi},\quad y^{pq}_{ij}\le x_{qj},\\
  y^{pq}_{ij}\ge x_{pi}+x_{qj}-1,\\
  \textcolor{black}{x_{pi}\in\{0,1\}},\quad y^{pq}_{ij}\in\{0,1\}.
\end{array}}};
\node[font=\figfont{14.7}{17.2}] at (28.0875,1.68) {\renewcommand{\arraystretch}{.82}\setlength{\arraycolsep}{3pt}%
\lzmathhighlight{figpalemint}{figgreenink}{\begin{array}{@{}c@{}}
  \textstyle\sum_j y^{pq}_{ij}=x_{pi},\\
  \textstyle\sum_i y^{pq}_{ij}=x_{qj},\\
  \textcolor{black}{x_{pi}\in\{0,1\}},\quad 0\le y^{pq}_{ij}\le1.
\end{array}}};
\foreach \xx/\interpretation/\size in {
  3.9125/{10 intensities per block}/{960,000 variables $\cdot$ 96,001 constraints},
  11.7375/{One price; one bound per block}/{96,001 variables $\cdot$ 960,000 constraints},
  20.2625/{Product links; binary $y$}/{5,472 binaries $\cdot$ 15,606 constraints},
  28.0875/{Marginal links; continuous $y$}/{288 binaries $\cdot$ 1,782 constraints}}{
  \node[font=\figfont{12.2}{14}] at (\xx,.14) {\textbf{Structure:} \interpretation};
  \node[font=\figfont{11.6}{13.4}] at (\xx,-.28) {\textbf{Size:} \size};
}
\foreach \xx/\runtime/\ink in {
  3.9125/{1,048.04}/{figredink},11.7375/{4.44}/{figgreenink},
  20.2625/{3,693.24}/{figredink},28.0875/{20.75}/{figgreenink}}{
  \node[anchor=west,font=\figfont{12.8}{15.0}] at (\xx-1.90,-1.59)
    {\textbf{Status}: OPTIMAL};
  \node[anchor=west,font=\figfont{12.8}{15.0},text=\ink] at (\xx-1.90,-2.08)
    {\textbf{\texttt{Runtime}}: \runtime\,s};
}
\end{scope}
\draw[black!55,line width=1.05pt] (16.0,-2.62)--(16.0,7.08);
\end{scope}
\begin{scope}[yshift=-13.59cm]
\begin{scope}
\clip[rounded corners=6pt] (-.28,10.97) rectangle (32.28,19.10);
\fill[white] (-.28,10.97) rectangle (32.28,19.10);
\fill[figbluehead] (-.28,18.43) rectangle (32.28,19.10);
\end{scope}
\draw[lzouterpanel] (-.28,10.97) rectangle (32.28,19.10);
\draw[black!35,line width=.4pt,dash pattern=on 2.2pt off 2.2pt]
  (-.28,18.43)--(32.28,18.43);
\draw[black!55,line width=1.05pt] (20.24,10.97)--(20.24,19.10);
\node[anchor=center,font=\figfont{14.9}{17}\bfseries] at (10.12,18.765) {(c) Solver efficiency across models};
\node[anchor=center,font=\figfont{14.9}{17}\bfseries] at (26.12,18.765)
  {(d) Faster or slower than expert?};
\node[anchor=center,font=\figfont{10.8}{12.5}] at (26.12,18.22)
  {(Counts on each model's correctly solved tasks)};
\definecolor{figown}{HTML}{FFD077}
\definecolor{figshared}{HTML}{709EFA}
\newcommand{\lzpatternedbar}[6]{%
  \fill[#1] (#3,#4) rectangle (#5,#6);
  \begin{scope}
    \clip (#3,#4) rectangle (#5,#6);
    \ifnum#2=0
      \foreach \dx in {.105,.315,.525}{
        \foreach \dy in {.12,.365,...,4.285}{
          \draw[white,line width=.60pt] ({#3+\dx},{#4+\dy}) circle[radius=.037cm];
        }
      }
    \else
      \foreach \dd in {-4.76,-4.48,...,.84}{
        \draw[white,line width=.65pt] ({#3+\dd},#4)--({#3+\dd+4.8},{#4+4.8});
      }
    \fi
  \end{scope}
  \draw[black!75,line width=.45pt] (#3,#4) rectangle (#5,#6);
}
\foreach \xx/\col/\pat/\txt in {
  3.08/figown/0/{Own correct subset ($n$)},
  9.09/figshared/1/{Shared correct subset ($m=47$)}}{
  \lzpatternedbar{\col}{\pat}{\xx}{17.61}{\xx+.46}{17.95}
  \node[anchor=west,font=\figfont{12.4}{14},text=black] at (\xx+.62,17.79) {\txt};
}
\def\runtimeLeft{1.67}\def\runtimeBottom{12.82}\def\runtimeHeight{4.14}\def\runtimeStep{1.395}
\node[rotate=90,font=\figfont{12.4}{14},text=black] at (.31,14.89)
  {Mean Gurobi runtime (s)};
\foreach \tick in {0,50,100,150,200,250,300,350}{
  \pgfmathsetmacro{\yy}{\runtimeBottom+\runtimeHeight*\tick/350}
  \draw[figgrid,line width=.7pt] (\runtimeLeft,\yy)--(19.80,\yy);
  \node[anchor=east,font=\figfont{12.8}{14}] at (1.43,\yy) {\tick};
}
\draw[figink,line width=1.25pt] (\runtimeLeft,16.96)--(\runtimeLeft,\runtimeBottom)--(19.80,\runtimeBottom);
\foreach \idx/\own/\shared/\model in {
0/299.5375388581889/195.04770212477825/{Ordinary\\reference\\{\color{black}$(n{=}543)$}},
  1/68.422563152857691/43.045787248205635/{Expert\\reference\\{\color{black}$(n{=}543)$}},
  2/150.78185100018666/171.76540423454122/{GLM\\5.1\\{\color{black}$(n{=}382)$}},
  3/177.58955293578663/130.44844680136822/{Claude\\Opus 4.6\\{\color{black}$(n{=}435)$}},
  4/190.92105735697317/109.45197874941725/{DeepSeek\\V4 Flash\\{\color{black}$(n{=}433)$}},
  5/197.6567739446958/77.043319139074768/{Gemini\\3.1 Pro\\{\color{black}$(n{=}504)$}},
  6/200.1010001634707/182.39595745472198/{Qwen\\3.5 122B\\{\color{black}$(n{=}367)$}},
  7/208.50061723380404/197.4411276604267/{GPT-5.5\\\strut\\{\color{black}$(n{=}441)$}},
  8/210.27955615325052/166.82965955835706/{Kimi\\K2.6\\{\color{black}$(n{=}427)$}},
  9/214.12935320575639/134.56895747590573/{Qwen\\3.6 Plus\\{\color{black}$(n{=}202)$}},
  10/215.00485030464503/144.98195743053518/{Qwen\\3.6 27B\\{\color{black}$(n{=}391)$}},
  11/232.36217281136098/181.26665958952395/{MiniMax\\M2.5\\{\color{black}$(n{=}311)$}},
  12/247.31545062358325/125.19429209891786/{Qwen 3\\32B\\{\color{black}$(n{=}309)$}}}{
  \pgfmathsetmacro{\xx}{\runtimeLeft+.6975+\idx*\runtimeStep}
  \pgfmathsetmacro{\ownTop}{\runtimeBottom+\runtimeHeight*\own/350}
  \pgfmathsetmacro{\sharedTop}{\runtimeBottom+\runtimeHeight*\shared/350}
  \lzpatternedbar{figown}{0}{\xx-.63}{\runtimeBottom}{\xx}{\ownTop}
  \lzpatternedbar{figshared}{1}{\xx}{\runtimeBottom}{\xx+.63}{\sharedTop}
  \node[anchor=south,inner sep=0pt,font=\figfont{8.2}{9.2},text=black]
    at (\xx-.315,\ownTop+.09) {\pgfmathprintnumber[fixed,precision=1,fixed zerofill]{\own}};
  \node[anchor=south,inner sep=0pt,font=\figfont{8.2}{9.2},text=black]
    at (\xx+.315,\sharedTop+.09) {\pgfmathprintnumber[fixed,precision=1,fixed zerofill]{\shared}};
  \node[anchor=north,align=center,font=\figfont{10.0}{11.2}] at (\xx,12.52) {\model};
}
\newcommand{\lzcountbar}[6]{%
  \fill[#1] (#3,#4) rectangle (#5,#6);
  \begin{scope}
    \clip (#3,#4) rectangle (#5,#6);
    \ifnum#2=0
      \foreach \dx in {.12,.365,...,8.20}{
        \foreach \dy in {.095,.26}{
          \draw[white,opacity=.75,line width=.45pt] ({#3+\dx},{#4+\dy}) circle[radius=.033cm];
        }
      }
    \else\ifnum#2=1
      \foreach \dx in {-.42,-.16,...,8.16}{
        \draw[white,opacity=.75,line width=.45pt] ({#3+\dx},#4)--({#3+\dx+.40},{#4+.40});
      }
    \fi\fi
  \end{scope}
  \draw[black!65,line width=.35pt] (#3,#4) rectangle (#5,#6);
}
\foreach \xx/\col/\pat/\txt in {23.55/figfast/0/{Faster (n)},26.13/figslow/1/{Slower (n)},28.94/figtie/2/{Tie (n)}}{
  \lzcountbar{\col}{\pat}{\xx}{17.61}{\xx+.46}{17.95}
  \node[anchor=west,font=\figfont{12.4}{14}] at (\xx+.62,17.79) {\txt};
}
\def\countLeft{23.66}\def\countScale{.0144}
\foreach \tick in {0,100,200,300,400,500}{
  \pgfmathsetmacro{\xx}{\countLeft+\countScale*\tick}
  \draw[figgrid,line width=.7pt] (\xx,11.91)--(\xx,17.40);
  \node[anchor=north,font=\figfont{11.7}{13}] at (\xx,11.70) {\tick};
}
\draw[figink,line width=1.1pt] (\countLeft,17.40)--(\countLeft,11.91)
  --({\countLeft+\countScale*520},11.91);
\foreach \idx/\model/\fast/\tie/\slow in {
0/{GLM-5.1}/125/0/257,
  1/{Claude Opus 4.6}/179/1/255,
  2/{DeepSeek-V4 Flash}/153/0/280,
  3/{Gemini 3.1 Pro}/187/0/317,
  4/{Qwen 3.5 122B}/124/0/243,
  5/{GPT-5.5}/149/0/292,
  6/{Kimi K2.6}/139/0/288,
  7/{Qwen 3.6 Plus}/59/0/143,
  8/{Qwen 3.6 27B}/120/0/271,
  9/{MiniMax M2.5}/108/0/203,
  10/{Qwen 3 32B}/87/0/222}{
  \pgfmathsetmacro{\yy}{17.11-.478*\idx}
  \pgfmathsetmacro{\fastEnd}{\countLeft+\countScale*\fast}
  \pgfmathsetmacro{\slowEnd}{\fastEnd+\countScale*\slow}
  \pgfmathsetmacro{\tieEnd}{\slowEnd+\countScale*\tie}
  \lzcountbar{figfast}{0}{\countLeft}{\yy-.167}{\fastEnd}{\yy+.167}
  \lzcountbar{figslow}{1}{\fastEnd}{\yy-.167}{\slowEnd}{\yy+.167}
  \ifnum\tie>0 \lzcountbar{figtie}{2}{\slowEnd}{\yy-.167}{\tieEnd}{\yy+.167}\fi
  \node[anchor=east,font=\figfont{10.8}{12.5}] at (23.47,\yy) {\model};
  \node[font=\figfont{10.5}{12},text=black,fill=figfast,inner xsep=1.0pt,inner ysep=.3pt]
    at ({(\countLeft+\fastEnd)/2},\yy) {\fast};
  \node[font=\figfont{10.5}{12},text=black,fill=figslow,inner xsep=1.0pt,inner ysep=.3pt]
    at ({(\fastEnd+\slowEnd)/2},\yy) {\slow};
  \ifnum\tie>0
    \draw[black!55,line width=.45pt] (\tieEnd,\yy)--(\tieEnd+.16,\yy);
    \node[anchor=west,font=\figfont{10.8}{12}] at (\tieEnd+.23,\yy) {\tie~tie};
  \fi
}
\end{scope}
\end{tikzpicture}%
}%
\endgroup
\caption{\small\textbf{Matching optimal values can conceal different computational costs.} Panels (a,b) present two case studies comparing GPT-5.5-generated and expert formulations on identical inputs; each pair reaches matching optimal objective values.
\textbf{(a) Technology mixing case.} Production blocks share a labor budget. The generated primal LP models each block--mode intensity; the expert dual LP uses one resource-price variable and one bound variable per block. The generated model requires about $236\times$ the recorded Gurobi solver time of the expert reference.
\textbf{(b) Department relocation case.} Departments are assigned to cities with communication costs between them. The generated model uses binary interaction variables; the expert links continuous interaction variables to binary assignments through row and column sums. The generated model requires about $178\times$ the recorded Gurobi solver time of the expert reference.
\textbf{(c)} Mean Gurobi runtimes compare LLM-generated models with ordinary and expert reference baselines, using correctly solved tasks in the reference cost subset (gold) and its 47 tasks solved correctly by all 11 LLMs (blue). Expert references have the lowest means in both views; the gold subsets can differ across methods.
\textbf{(d)} Green, purple, and gray bars count programs that are faster, slower, or tied with the expert on correctly solved tasks. For every LLM, most take longer than the expert, showing that numerical agreement alone does not establish efficient modeling.}
\label{fig:introduction}
\label{fig:results-561}
\end{figure}

Answering this question requires tasks that allow us to compare different modeling approaches and their computational costs. The two case studies in Figure~\ref{fig:introduction}(a,b) illustrate why: in technology mixing, the expert uses resource prices in a dual LP rather than explicitly modeling every production intensity; in department relocation, it retains binary assignments but represents their interactions with continuous variables linked through row and column sums. Both expert references attain the same optimal objective values as the corresponding GPT-5.5-generated models with substantially less recorded solver time (see Appendix~\ref{app:case-formulations} for the mathematical checks). To study such decisions systematically, we complement recent work on solver efficiency and strategy-aware optimization modeling \citep{zhao2026sage,kong2026frontieror} with tasks designed around problem structures to which specific expert modeling techniques apply. Each task includes ordinary and expert references on the same data: the ordinary reference provides a conventional mathematical model and solver code, while the expert reference applies a selected technique. Two OptTips examples in Figure~\ref{fig:opttips-structure} illustrate this distinction: replacing a loose Big-$M$ constant with a tighter valid one derived from variable bounds, and adding a valid inequality to strengthen an LP relaxation. These contrasts let us examine LLM modeling choices beyond answer correctness. Comparing the generated models and code with the paired references allows us to assess correctness, technique use, and computational cost separately.

We first curate \textbf{OptTips}, a knowledge base of 50 expert modeling techniques in eight families, drawn from optimization textbooks, research literature, and solver documentation. Here, \emph{optimization modeling techniques} are reusable ways to exploit mathematical structure when representing a problem for a solver or organizing it into coordinated submodels. OptTips contains 50 \emph{technique cards}, one structured entry per technique. Each card explains when and how to apply the technique and which inefficiency it targets, with a mathematical example. To our knowledge, OptTips is the first expert-curated technique catalog designed to construct an LLM optimization-modeling benchmark that separately evaluates technique use and computational cost. Using this knowledge, we develop \textbf{OptDachshund}, a multi-agent framework for constructing tasks that test technique use. Its six agents match techniques to source problems, redesign, audit, and repair tasks, and build and validate ordinary and expert reference implementations on the same instance. Solver validation and independent expert review yield \textbf{EfficientOpt}, a benchmark of 561 problems spanning 29 target-technique groups. Each task includes a natural-language description, numerical data, and paired references for cost comparisons. Withholding references and target-technique labels tests whether LLMs independently recognize problem structure and apply suitable techniques.

Evaluation of 11 representative LLMs on this benchmark reveals five main findings: \textit{1) Correct answers can still be costly.} Expert references have lower recorded mean solver runtimes (Figure~\ref{fig:introduction}(c)). For every LLM, most programs returning correct objective values take longer to solve than the corresponding expert implementation (Figure~\ref{fig:introduction}(d)). \textit{2) A few slow cases dominate total solver time.} Similar median solve times can hide different mean costs; across models, the slowest tenth consumes roughly 70\% of measured solver time on correctly solved tasks (Figure~\ref{fig:model-comparison-561}(b); Table~\ref{tab:absolute-costs-561}). \textit{3) Smaller formulations are not necessarily faster.} Our analysis compares variable, linear-constraint, and matrix-nonzero counts with solver runtimes (Section~\ref{sec:structural_diagnostics}). \textit{4) Technique use and observed cost are distinct.} Case studies identify different formulations attaining the same optimal value at similar recorded cost (Section~\ref{sec:structural_diagnostics}). \textit{5) Efficiency extends beyond solver runtime.} Faster solving can be offset by longer model-construction time (Section~\ref{sec:cost-shifting}). We also report memory use (Table~\ref{tab:matched-memory-561}).

Our contributions are fourfold: \textbf{1) OptTips.} We curate 50 techniques in eight families into a knowledge base linking problem structure and inefficiency symptoms to expert modeling actions. \textbf{2) OptDachshund.} We develop a multi-agent framework that uses OptTips to construct tasks in three stages: match and reformulate, audit and repair, and build and validate. \textbf{3) EfficientOpt.} We construct a 561-task benchmark with ordinary and expert references and hidden technique labels to evaluate correctness, technique use, and computational efficiency separately. \textbf{4) Experiments and findings.} Our study of 11 LLMs examines correctness, technique use, model size, and generation, construction, solver, and memory costs. We find that correct objective values and compact formulations can still come at substantial cost, and compare alternative modeling choices with expert references. The EfficientOpt dataset, evaluation code, and OptTips knowledge cards are available in our public repository.\footnote{\url{https://github.com/ZhongLIFR/EfficientOpt}}

\section{Related Work}
\label{sec:related-work}

Due to space constraints, we discuss only the most closely related work here and defer a broader review to Appendix~\ref{app:extended-related-work}. \textit{1) Expert knowledge for optimization modeling.} OptiMind \citep{zhang2025optimind} uses expert hints to guide training and inference. In concurrent work, OptSkills \citep{yang2026optskills} distills modeling and solving trajectories into reusable workflows for inference. These methods use knowledge to assist generation. Our additional contribution is to turn individual techniques into specifications for constructing evaluation tasks. Our OptTips specifies applicability conditions and modeling actions, such as deriving tighter valid Big-$M$ constants from variable bounds. Our OptDachshund turns these specifications into test problems and paired reference implementations. The target techniques and references are hidden from evaluated LLMs, so the benchmark tests independent technique selection and application. \textit{2) Evaluating modeling quality and efficiency.} SAGE \citep{zhao2026sage} trains LLMs with correctness and solver-efficiency feedback and compares formulations generated by different models. In concurrent work, FrontierOR \citep{kong2026frontieror} evaluates algorithm quality and runtime against one representative Gurobi reference per task. Their reported evaluations do not provide a conventional/expert reference pair tied to a designated modeling technique. Our EfficientOpt adds this pair for each task on identical data, making the contrast between conventional and expert modeling explicit. We inspect technique use and separately assess numerical correctness, model-construction cost, and solving cost. This distinguishes technique adoption from efficient implementation and recognizes efficient solutions using other techniques.

\section{Benchmark Construction}
\label{sec:method}

\noindent\textbf{Definition 1 (Optimization modeling).} An optimization task is $p=(q,I,\mathcal O)$, where $q$ specifies the problem in natural language, $I$ is the fixed numerical input, and $\mathcal O$ specifies the required outputs and accuracy. \emph{Optimization modeling} constructs a solution procedure $\widehat{\mathcal A}=(\widehat{\mathcal M},\widehat C)$, where $\widehat{\mathcal M}$ is a mathematical representation and $\widehat C$ is its executable implementation. The representation specifies variables, domains, objectives, and constraints, in a single formulation or coordinated submodels. The implementation handles any submodel coordination, stopping rules, and required answer recovery. A valid procedure must respect the problem semantics and satisfy $\mathcal O$: an optimal value alone is insufficient when original decisions are required. Different valid procedures may use different variables and feasible sets. Modeling techniques, such as tighter bounds, dualization, or decomposition, exploit problem structure in choosing and implementing these representations.

\noindent\textbf{Definition 2 (Modeling efficiency).} For the same task $p$ and required answer quality, \emph{modeling efficiency} concerns the time and memory needed to produce and execute a valid solution procedure. Lower cost indicates greater efficiency only for the stated resource, accounting scope, and execution conditions; time and memory are assessed separately. We distinguish \emph{generation}, which produces $\widehat{\mathcal A}$, from \emph{execution}, which runs $\widehat C$ on $I$. Execution includes data preparation, model construction, all solver calls, computation between calls, and answer recovery. Model construction specifically means executing code to instantiate variables, objectives, and constraints. Thus, faster solving can be offset by longer construction time, and neither smaller models nor technique adoption guarantees lower cost. Complete request costs also include retries, repairs, and checking. Appendix~\ref{app:formal-task} details the definitions and the stages our measurements cover.

\begin{figure}[!htbp]
\centering
\begingroup
\definecolor{pfblue}{HTML}{347AF0}
\definecolor{pfblueink}{HTML}{215BAE}
\definecolor{pfbluefill}{HTML}{E7F0FC}
\definecolor{pfgreen}{HTML}{6DAB7C}
\definecolor{pfgreenfill}{HTML}{EBF4EC}
\definecolor{pfgray}{HTML}{9298A1}
\definecolor{pfgrayfill}{HTML}{F2F3F5}
\definecolor{pfmuted}{HTML}{62666C}
\definecolor{pffeedback}{HTML}{CC4262}
\definecolor{pfpurple}{HTML}{8053A8}
\definecolor{pfredink}{HTML}{BE1534}
\definecolor{pfpalered}{HTML}{FBE8ED}
\definecolor{pfgreenink}{HTML}{07623E}
\definecolor{pfpalemint}{HTML}{EDF8F3}
\definecolor{pfA}{HTML}{1AB5BF}
\definecolor{pfD}{HTML}{6862D8}
\definecolor{pfE}{HTML}{9A56CD}
\definecolor{pfG}{HTML}{E85B86}
\definecolor{pfC}{HTML}{EE7B52}
\definecolor{pfH}{HTML}{F5C34F}
\definecolor{pfB}{HTML}{B49C58}
\definecolor{pfF}{HTML}{657885}
\newcommand{\pffont}[2]{\fontsize{#1}{#2}\selectfont}
\resizebox{\textwidth}{!}{%
\begin{tikzpicture}[x=1cm,y=1cm,line cap=round,line join=round,
  every node/.style={inner sep=0pt,outer sep=0pt,align=center,text=black},
  pfpanel/.style={line width=.65pt,dash pattern=on 3pt off 2pt,rounded corners=4pt},
  pftitle/.style={font=\bfseries\pffont{6.6}{7.5}},
  pfsmall/.style={font=\pffont{5.3}{6.1},text=pfmuted},
  pfheadrule/.style={draw=pfgray!70,line width=.35pt,dash pattern=on 1.5pt off 1.2pt},
  pfflow/.style={draw=pfblue,line width=.75pt,-{Stealth[length=3pt,width=2.8pt]}},
  pffail/.style={draw=pffeedback,line width=.7pt,densely dashed,rounded corners=2pt},
  pflabel/.style={font=\bfseries\pffont{5.0}{5.8},fill=white,inner sep=.5pt}]
\path[use as bounding box] (0,0) rectangle (14,3.03);

\draw[pfpanel,draw=pfblue,fill=pfbluefill] (2.35,.40) rectangle (4.64,2.92);
\draw[pfpanel,draw=pfgreen,fill=pfgreenfill] (4.92,.40) rectangle (9.58,2.92);
\draw[pfpanel,draw=pfgray,fill=pfgrayfill] (9.86,.40) rectangle (12.40,2.92);
\node[pftitle] at (1.08,2.74) {Knowledge base};
\node[pftitle] at (3.495,2.74) {OptDachshund};
\node[pftitle] at (7.25,2.74) {Paired references};
\node[pftitle,font=\bfseries\pffont{6.2}{7.1}] at (11.13,2.74) {Validation \& review};
\draw[pfheadrule] (2.47,2.58)--(4.52,2.58);
\draw[pfheadrule] (5.04,2.58)--(9.46,2.58);
\draw[pfheadrule] (9.98,2.58)--(12.28,2.58);
\node[pfsmall] at (3.495,2.43) {(Sec.~\ref{subsec:multi_agent_framework})};

\begin{scope}[shift={(1.08,1.51)}]
  \foreach \start/\end/\mid/\letter/\count/\fg in {
    90/133.2/111.6/A/6/black,
    133.2/198/165.6/F/9/white,
    198/234/216/B/5/black,
    234/291.6/262.8/H/8/black,
    291.6/320.4/306/C/4/black,
    320.4/363.6/342/G/6/black,
    363.6/392.4/378/E/4/white,
    392.4/450/421.2/D/8/white}{
    \filldraw[fill=pf\letter,draw=white,line width=.55pt]
      (\start:.46)--(\start:1.02)
      arc[start angle=\start,end angle=\end,radius=1.02]
      --(\end:.46) arc[start angle=\end,end angle=\start,radius=.46]--cycle;
    \node[text=\fg,font=\sffamily\bfseries\pffont{6.8}{7.5}] at (\mid:.755) {\count};
  }
  \draw[black!25,line width=.25pt] (0,0) circle (1.02);
  \draw[black!25,line width=.25pt] (0,0) circle (.46);
  \node[font=\sffamily\bfseries\pffont{5.9}{6.5}] at (0,.10) {OptTips};
  \node[pfsmall] at (0,-.14) {(Sec.~\ref{subsec:opttips})};
\end{scope}
\draw[pfflow] (2.12,1.54)--(2.33,1.54);

\node[font=\bfseries\pffont{5.3}{6.1},text=pfblueink] at (3.495,2.21) {Multi-agent framework};
\node[pfsmall] at (3.495,2.00) {Source benchmark problems};
\draw[pfflow,-{Stealth[length=2pt,width=2pt]}] (3.495,1.91)--(3.495,1.83);
\node at (2.99,1.41)
  {\includegraphics[width=.49cm,height=.58cm,keepaspectratio]{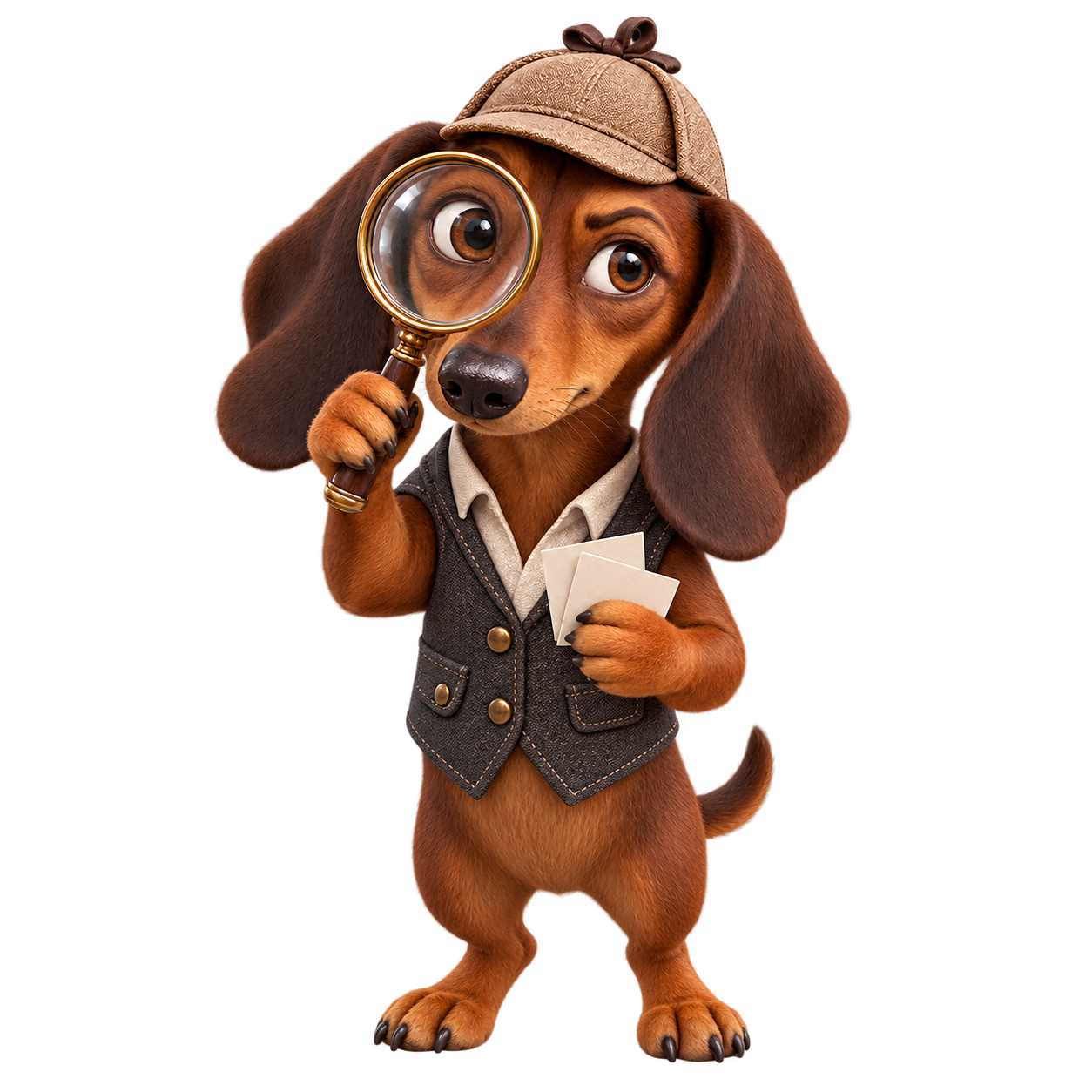}};
\node at (3.495,1.44)
  {\includegraphics[width=.70cm,height=.76cm,keepaspectratio]{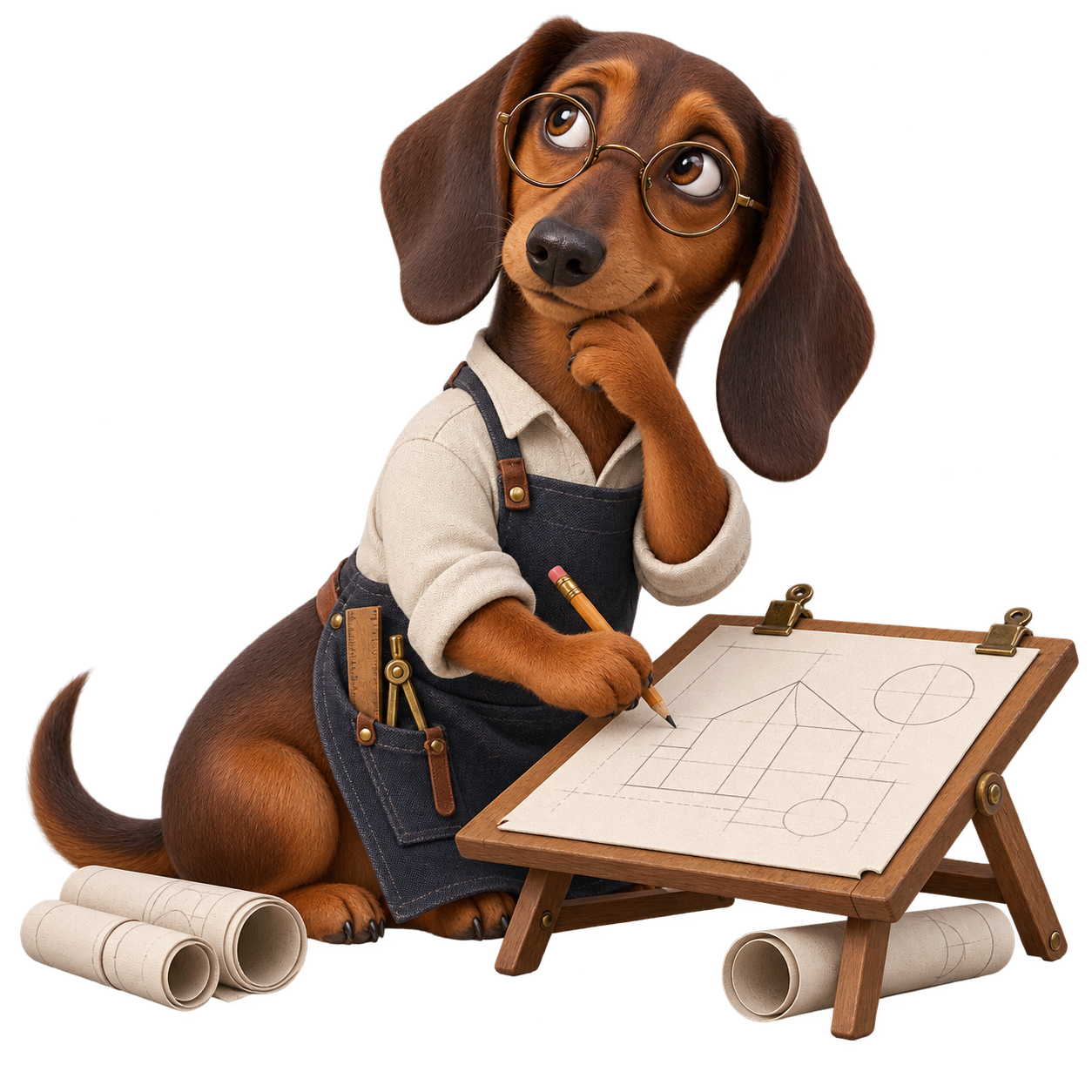}};
\node at (4.02,1.41)
  {\includegraphics[width=.51cm,height=.58cm,keepaspectratio]{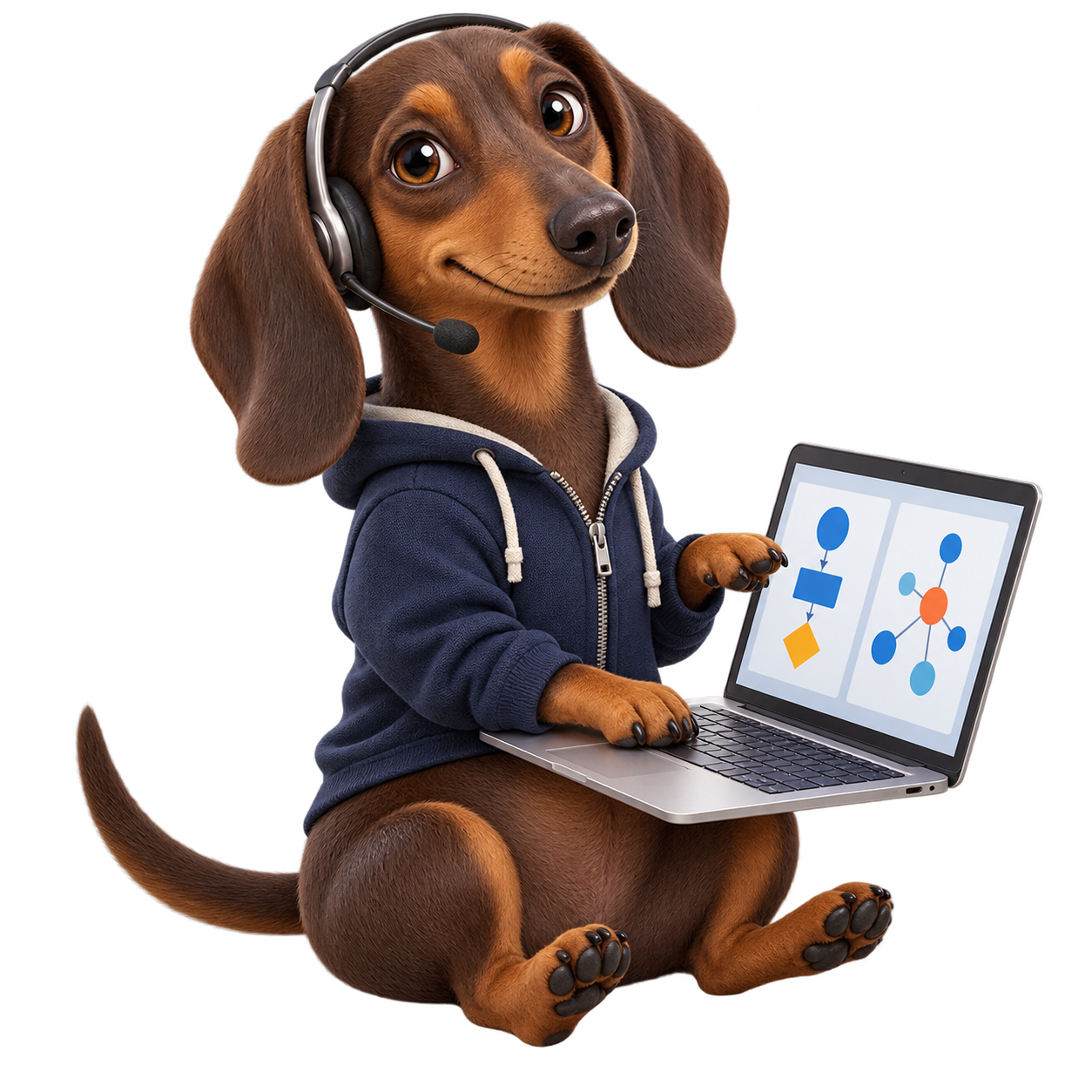}};
\node[font=\bfseries\pffont{5.9}{6.7}] at (3.495,.91) {Redesign tasks};
\node[font=\pffont{5.5}{6.3}] at (3.495,.62) {Build \& verify references};
\draw[pfflow] (4.66,1.54)--(4.90,1.54);

\node[pfsmall] at (7.25,2.45) {Same cities and travel costs};
\draw[fill=white,draw=black!45,line width=.5pt,rounded corners=3pt]
  (5.08,.53) rectangle (9.42,2.25);
\draw[black!25,line width=.3pt] (7.25,.64)--(7.25,2.13);
\node[font=\bfseries\pffont{5.8}{6.6}] at (6.165,2.05)
  {\faIcon{file-alt}\; Ordinary};
\node[font=\pffont{5.0}{5.8}] at (6.165,1.83) {(MTZ)};
\node[font=\bfseries\pffont{5.8}{6.6}] at (8.335,2.05)
  {\faIcon{user-graduate}\; Expert};
\node[font=\pffont{5.0}{5.8}] at (8.335,1.83) {(Subtour inequalities)};
\fill[pfpalered] (5.16,.90) rectangle (7.17,1.67);
\fill[pfpalemint] (7.33,.90) rectangle (9.34,1.67);
\node[font=\pffont{5.2}{6.0},text=pfredink] at (6.165,1.49)
  {\resizebox{1.85cm}{!}{$u_i-u_j+n x_{ij}\le n-1$}};
\node[font=\pffont{5.0}{5.8},text=pfredink] at (6.165,1.24)
  {\resizebox{1.80cm}{!}{$i,j\in V\setminus\{1\},\ i\ne j$}};
\node[font=\pffont{5.0}{5.8},text=pfredink] at (6.165,1.02)
  {\resizebox{1.24cm}{!}{$1\le u_i\le n-1$}};
\node[font=\pffont{5.2}{6.0},text=pfgreenink] at (8.335,1.40)
  {\resizebox{1.80cm}{!}{$\displaystyle\sum_{\substack{i,j\in S\\i\ne j}}x_{ij}\le |S|-1$}};
\node[font=\pffont{5.0}{5.8},text=pfgreenink] at (8.335,1.02)
  {\resizebox{1.88cm}{!}{$\forall S\subsetneq V,\ 2\le |S|<n$}};
\node[font=\pffont{4.9}{5.7},text=pfmuted] at (6.165,.71) {Visit-order variables};
\node[font=\pffont{4.9}{5.7},text=pfmuted] at (8.335,.71) {Exclude smaller cycles};
\draw[pfflow] (9.60,1.54)--(9.84,1.54);

\node[pfsmall] at (11.13,2.43) {(App.~\ref{app:quality-control})};
\draw[fill=white,draw=pfblue!65,line width=.5pt,rounded corners=2pt]
  (9.99,1.66) rectangle (12.27,2.29);
\begin{scope}[shift={(10.23,1.98)},scale=.28]
  \filldraw[fill=pfblueink,draw=pfblueink,line width=.5pt]
    (-.46,-.24)--(-.39,.25)--(.09,.06)--(.28,-.47)--cycle;
  \filldraw[fill=white,draw=pfblueink,line width=.5pt]
    (-.39,.25)--(-.03,.49)--(.27,.44)--(.09,.06)--cycle;
  \filldraw[fill=white,draw=pfblueink,line width=.5pt]
    (.09,.06)--(.27,.44)--(.51,-.25)--(.28,-.47)--cycle;
\end{scope}
\node[font=\bfseries\pffont{5.3}{6.1},text=pfblueink] at (11.30,2.13)
  {Solver verification};
\node[font=\pffont{4.7}{5.5}] at (11.30,1.93) {Gurobi: both references};
\node[font=\pffont{4.65}{5.4},text=pfmuted] at (11.13,1.75)
  {Status, feasibility, objective};
\node[font=\bfseries\pffont{5.15}{6.0}] at (11.13,1.49)
  {{\color{pfpurple}\faUserCheck}\; Independent review};
\draw[fill=white,draw=pfgray!40,line width=.3pt,rounded corners=2pt]
  (9.99,.53) rectangle (12.27,1.33);
\foreach \yy/\reviewtext in {
  1.20/{Task \& data validity},
  1.00/{Reference validity},
  .80/{Technique use \& leakage},
  .60/{Coverage \& duplicates}}{
  \draw[draw=pfpurple!75,line width=.4pt,rounded corners=.4pt,fill=white]
    (10.075,\yy-.035) rectangle (10.185,\yy+.075);
  \draw[draw=pfgreen!75!black,line width=.5pt]
    (10.09,\yy+.015)--(10.12,\yy-.013)--(10.185,\yy+.06);
  \node[anchor=west,font=\pffont{4.9}{5.7}] at (10.29,\yy+.015) {\reviewtext};
  \draw[pfblue!35,line width=.25pt] (10.29,\yy-.075)--(12.17,\yy-.075);
}
\draw[pfflow] (12.42,1.54)--(12.73,1.54);
\node[pflabel,text=pfblueink] at (12.62,1.73) {accept};
\node[text=pfblueink,font=\pffont{16}{18}] at (13.35,1.80) {\faDatabase};
\node[font=\bfseries\pffont{6.5}{7.5}] at (13.35,1.23) {EfficientOpt};
\node[pfsmall] at (13.35,.99) {(Sec.~\ref{subsec:efficientopt})};
\node[font=\pffont{5.8}{6.6},text=pfmuted] at (13.35,.73) {561 tasks};

\draw[pffail,-{Stealth[length=2.8pt,width=2.6pt]}]
  (11.13,.38)--(11.13,.15)--(3.495,.15)--(3.495,.38);
\node[pflabel,text=pffeedback] at (7.25,.15) {revise \& re-verify};
\draw[draw=pfgray,line width=.55pt,-{Stealth[length=2.2pt,width=2pt]}]
  (12.12,.38)--(12.12,.15)--(12.33,.15);
\node[font=\pffont{4.8}{5.6},text=pfmuted,anchor=west] at (12.38,.15) {reject};
\end{tikzpicture}%
}
\endgroup
\caption{\small\textbf{Benchmark construction overview.} Our knowledge base OptTips (Sec.~\ref{subsec:opttips}) guides our multi-agent framework OptDachshund (Sec.~\ref{subsec:multi_agent_framework}) to build tasks and paired references. Gurobi verification and independent expert review (Appendix~\ref{app:quality-control}) lead to revision, rejection, or acceptance into our benchmark EfficientOpt (Sec.~\ref{subsec:efficientopt}). The example seeks the cheapest tour visiting every city once. MTZ uses continuous visit-order variables $u_i$ (excluding reference city $1$); the expert model forbids disconnected cycles within each smaller city subset $S$. Here, $V$ contains $n$ cities and binary $x_{ij}$ selects travel from city $i$ to a different city $j$. Both models share travel costs and one-arrival/one-departure constraints, omitted for clarity.}
\label{fig:benchmark_pipeline}
\end{figure}
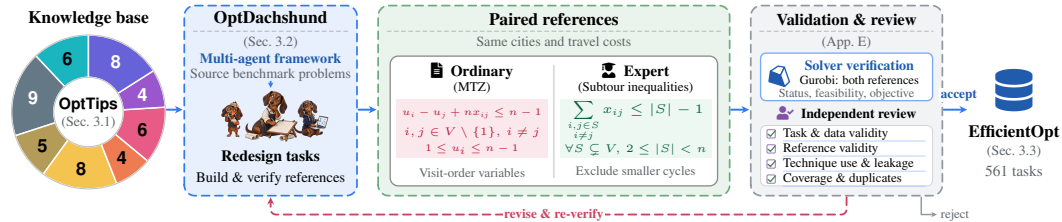

\noindent\textbf{Overview.} Figure~\ref{fig:benchmark_pipeline} summarizes the construction workflow. We first curate our knowledge base \emph{OptTips} (detailed in Sec.~\ref{subsec:opttips}), whose cards specify when and how to apply modeling techniques. Guided by these cards, our multi-agent framework \emph{OptDachshund} (detailed in Sec.~\ref{subsec:multi_agent_framework}) matches techniques to source problems, redesigns tasks, and builds paired ordinary/expert models with solver code on identical data. The resulting candidates then undergo solver verification and \emph{independent expert review} (detailed in Appendix~\ref{app:quality-control}) to check task and reference validity, technique use, leakage, and diversity. Failed checks trigger revision and re-verification; unresolved cases are rejected. Validated candidates form our benchmark \emph{EfficientOpt} (detailed in Sec.~\ref{subsec:efficientopt}).

\subsection{OptTips: Expert Modeling Knowledge}
\label{subsec:opttips}

\begin{figure}[h]
\centering
\begingroup
\definecolor{otink}{HTML}{23313D}
\definecolor{otmuted}{HTML}{566472}
\definecolor{otrule}{HTML}{BCC5CD}
\definecolor{otpanel}{HTML}{F5F7F9}
\definecolor{otaction}{HTML}{215D91}
\definecolor{otA}{HTML}{1AB5BF}
\definecolor{otD}{HTML}{6862D8}
\definecolor{otE}{HTML}{9A56CD}
\definecolor{otG}{HTML}{E85B86}
\definecolor{otC}{HTML}{EE7B52}
\definecolor{otH}{HTML}{F5C34F}
\definecolor{otB}{HTML}{B49C58}
\definecolor{otF}{HTML}{657885}

\begin{tikzpicture}[x=1pt,y=1pt,line cap=round,line join=round,
 every node/.style={inner sep=0pt,outer sep=0pt,text=otink,align=center},
 maincardtext/.style={anchor=north west,text width=125pt,align=left,font=\fontsize{5.0}{5.6}\selectfont},
 maincardmath/.style={font=\fontsize{5.7}{6.3}\selectfont},
 association/.style={line width=.65pt}]
\path[use as bounding box] (0,0) rectangle (396,100);
\begin{scope}[shift={(198,50)},x=18.18pt,y=18.18pt]
  \foreach \start/\end/\mid/\letter/\count/\fg in {
    90/133.2/111.6/A/6/otink,
    133.2/198/165.6/F/9/white,
    198/234/216/B/5/otink,
    234/291.6/262.8/H/8/otink,
    291.6/320.4/306/C/4/otink,
    320.4/363.6/342/G/6/otink,
    363.6/392.4/378/E/4/white,
    392.4/450/421.2/D/8/white}{
    \filldraw[fill=ot\letter,draw=white,line width=.9pt]
      (\start:.93)--(\start:2.75)
      arc[start angle=\start,end angle=\end,radius=2.75]
      --(\end:.93) arc[start angle=\end,end angle=\start,radius=.93]--cycle;
    \node[text=white,font=\sffamily\bfseries\fontsize{7.2}{8.0}\selectfont]
      at (\mid:1.16) {\count};
  }
  \draw[white,line width=1.1pt] (0,0) circle (1.39);
  \draw[otrule,line width=.35pt] (0,0) circle (2.75);
  \draw[otrule,line width=.35pt] (0,0) circle (.93);
  \foreach \angle/\radius/\name/\fg/\size in {
    114/2.03/{Standardize\\\& reduce}/otink/5.0,
    165.6/2.00/{Convex \&\\nonlinear}/white/5.0,
    209/2.00/{Stronger\\relaxations}/otink/4.9,
    262.8/2.06/{Uncertainty,\\objectives\\\& solvers}/otink/5.0,
    304/2.12/{Duality \&\\equilibria}/otink/4.9,
    337.0/2.00/{Decompose}/otink/4.9,
    377.8/2.01/{Discrete\\\& logical}/white/5.0,
    421.2/2.00/{Combinatorial\\structure}/white/5.0}{
    \node[text=white,font=\sffamily\fontsize{\size}{5.7}\selectfont]
      at (\angle:\radius) {\name};
  }
  \coordinate (otAedge) at (117.7:2.75);
  \coordinate (otEedge) at (18:2.75);
  \coordinate (otBedge) at (216:2.75);
  \coordinate (otCedge) at (306:2.75);
  \node[font=\sffamily\bfseries\fontsize{7.3}{8.0}\selectfont] at (0,.36) {OptTips};
  \draw[otmuted!70!white,line width=.3pt,dash pattern=on 1.3pt off 1.0pt]
    (-.64,.02)--(.64,.02);
  \node[font=\sffamily\fontsize{5.8}{6.5}\selectfont,text=otmuted]
    at (0,-.40) {50 cards\\8 families};
\end{scope}

\draw[association,otB!80!black] (otBedge) to[out=210,in=0] (137,31.5);
\draw[association,otE!80!black] (otEedge) to[out=18,in=180] (259,74.5);
\fill[otB!70!black] (otBedge) circle[radius=.9pt];
\fill[otE!70!black] (otEedge) circle[radius=.9pt];
\begin{scope}[shift={(0,0)}]
\begin{scope}\clip[rounded corners=3pt] (0,0) rectangle (137,100);
\fill[white] (0,0) rectangle (137,100);
\fill[otB!13!white] (0,90) rectangle (137,100);\end{scope}
\draw[otrule,line width=.55pt,rounded corners=3pt] (0,0) rectangle (137,100);
\draw[otB!75!black,line width=.6pt] (5,90)--(132,90);
\node[font=\bfseries\fontsize{6.3}{7.0}\selectfont] at (68.5,95) {T07\quad Valid inequalities};
\node[maincardtext] at (6,87) {\textbf{Core idea:} Cut fractional points while preserving integer solutions.};
\node[maincardtext] at (6,76) {\textbf{Applicable patterns:} Binary packing or capacity constraints.};
\node[maincardtext] at (6,65) {\textbf{Inefficiency symptoms:} A weak relaxation admits fractional packing.};
\node[maincardtext] at (6,54) {\textbf{Expert actions:} Add $x_1+x_2\le1$; retain the original row.};
\node[maincardtext,font=\bfseries\fontsize{5.0}{5.6}\selectfont] at (6,42) {Mathematical example};
\path[fill=otpanel,rounded corners=2pt] (5,16) rectangle (132,36);
\node[maincardmath,text=otmuted] at (68.5,32) {$x_1,x_2\in\{0,1\}$};
\node[maincardmath] at (68.5,25.5) {$\text{Original: }2x_1+2x_2\le3$};
\node[maincardmath,text=otaction] at (68.5,19) {$\text{Add: }x_1+x_2\le1$};
\node[maincardtext] at (6,13) {\textbf{Benchmark guidance:} Hold objective/data fixed; compare LP bounds and search nodes.};
\end{scope}
\begin{scope}[shift={(259,0)}]
\begin{scope}\clip[rounded corners=3pt] (0,0) rectangle (137,100);
\fill[white] (0,0) rectangle (137,100);
\fill[otE!13!white] (0,90) rectangle (137,100);\end{scope}
\draw[otrule,line width=.55pt,rounded corners=3pt] (0,0) rectangle (137,100);
\draw[otE!75!black,line width=.6pt] (5,90)--(132,90);
\node[font=\bfseries\fontsize{6.3}{7.0}\selectfont] at (68.5,95) {T05\quad Tight Big-$M$ bounds};
\node[maincardtext] at (6,87) {\textbf{Core idea:} Derive a valid $M$ from known variable bounds.};
\node[maincardtext] at (6,76) {\textbf{Applicable patterns:} Bounded $x$; binary $z$ controls \mbox{$x\le2$}.};
\node[maincardtext] at (6,65) {\textbf{Inefficiency symptoms:} An oversized $M$ weakens the LP relaxation.};
\node[maincardtext] at (6,54) {\textbf{Expert actions:} Use the upper bound: $M=6-2=4$.};
\node[maincardtext,font=\bfseries\fontsize{5.0}{5.6}\selectfont] at (6,42) {Mathematical example};
\path[fill=otpanel,rounded corners=2pt] (5,16) rectangle (132,36);
\node[maincardmath,text=otmuted] at (68.5,32) {$0\le x\le6,\quad z\in\{0,1\}$};
\node[maincardmath] at (68.5,25.5) {$\text{Loose: }x\le2+8(1-z)$};
\node[maincardmath,text=otaction] at (68.5,19) {$\text{Tight: }x\le2+4(1-z)$};
\node[maincardtext] at (6,13) {\textbf{Benchmark guidance:} Keep binary solutions; compare LP bounds and coefficient ranges.};
\end{scope}
\end{tikzpicture}
\endgroup
\caption{\small OptTips: 50 cards in eight families, with two abbreviated examples showing the seven-field structure. Full card examples appear in Appendix~\ref{app:opttips-catalog} (Figures~\ref{fig:opttips-eight-cards} and~\ref{fig:opttips-cards-eh}).}
\label{fig:opttips-structure}
\end{figure}
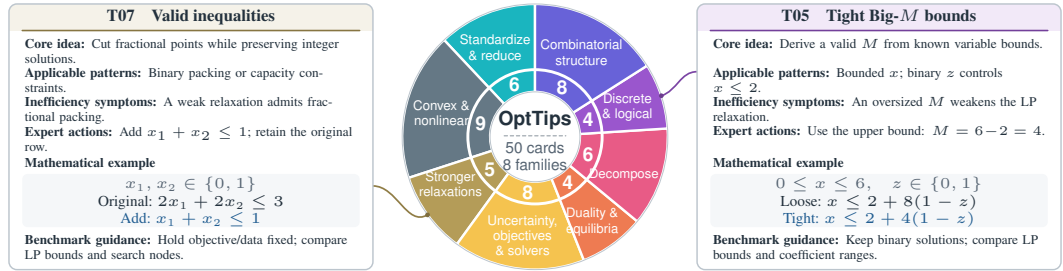

OptTips makes expert modeling knowledge explicit for task construction and evaluation. Drawing on optimization textbooks \citep{williams2013model,bertsimas1997linear,nemhauser1988integer,boyd2004convex,birge2011stochastic}, research literature \citep{dantzig1960decomposition,benders1962partitioning,mccormick1976computability,margot2010symmetry,vielma2015formulation}, and solver documentation \citep{gurobi_constraints,gurobi_numerics,gurobi_start,cplex_indicator_practices,mosek_modeling_cookbook}, we curate 50 modeling techniques, each represented by a structured \emph{technique card}. The collection spans foundational formulation choices and advanced methods across eight primary methodological families (Figure~\ref{fig:opttips-structure}). Its breadth concerns reusable modeling mechanisms, from bounds and reformulations to decomposition and structured optimization; 50 is the current collection size, not an exhaustive taxonomy. Appendix~\ref{app:opttips-catalog} provides the complete technique inventory, coverage rationale, and card schema. The full cards are available in our public repository.

Each OptTips card has seven fields. Taking the tight Big-$M$ bounds card in Figure~\ref{fig:opttips-structure} as an example, we explain what each field records. The \emph{(1)~identifier} provides a reference label, here T05. The \emph{(2)~core idea} states the mathematical principle: derive a tight, valid $M$ from known variable bounds. The \emph{(3)~applicable patterns} specify when the technique applies, such as encoding $z=1\Rightarrow x\le2$ with binary $z$ and $0\le x\le6$. The \emph{(4)~inefficiency symptoms} describe what needs improvement; here, an oversized $M$ leaves the LP relaxation unnecessarily loose. To address this, the \emph{(5)~expert actions} give the concrete modeling step: use the bound $x\le6$ to replace $M=8$ with $M=6-2=4$. The \emph{(6)~mathematical example} expresses the change in equations, comparing $x\le2+8(1-z)$ with $x\le2+4(1-z)$. Both have the same binary feasible solutions, but at relaxed $z=\tfrac12$, the upper bound on $x$ falls from $6$ to $4$. Finally, the \emph{(7)~benchmark guidance} explains how to construct and compare test models. For this technique, it calls for paired models with the same task, data, and objective: check that binary feasible solutions are unchanged, then compare LP objective bounds, coefficient ranges, and solve times. Appendix~\ref{app:opttips-catalog} provides eight full card examples, one per family.

We use OptTips to guide benchmark construction and to examine LLM modeling choices. During construction, a card specifies the problem structure needed for a technique and the change to implement in the expert reference. OptDachshund uses this information to redesign source problems and build ordinary and expert references for the same final task and data (Section~\ref{subsec:multi_agent_framework}). We assess technique use with an LLM judge across all generated programs and inspect selected cases manually (Appendices~\ref{app:judge-full} and~\ref{app:technique-cases}); Appendix~\ref{app:case-formulations} gives case-level mathematical checks and their limitations. We assess efficiency from recorded runtime and resource use, independently of technique adoption.

\FloatBarrier
\subsection{OptDachshund: Multi-Agent Benchmark Construction}
\label{subsec:multi_agent_framework}

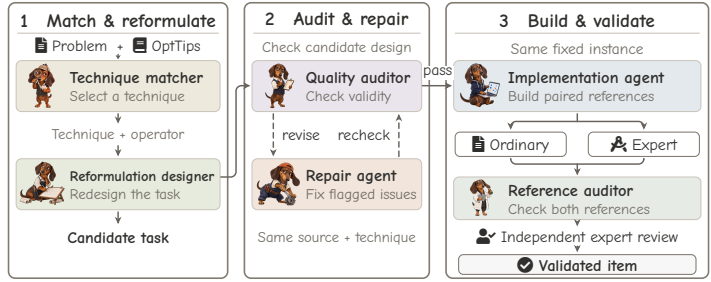
\begin{wrapfigure}{r}{9.3cm}
\vspace{-7pt}
\centering
\begingroup
\definecolor{wfink}{HTML}{302F2D}
\definecolor{wfmuted}{HTML}{706B64}
\definecolor{wfframe}{HTML}{817B72}
\definecolor{wfrule}{HTML}{C9C3BA}
\definecolor{wfflow}{HTML}{68635C}
\definecolor{wfmatch}{HTML}{EEE9DD}
\definecolor{wfdesign}{HTML}{E7EBDD}
\definecolor{wfquality}{HTML}{EAE5EC}
\definecolor{wfrepair}{HTML}{F1E5DC}
\definecolor{wfimplement}{HTML}{E3E9ED}
\definecolor{wfreference}{HTML}{E7EBE5}
\newcommand{\wfFont}{\fontencoding{OT1}\fontfamily{ComicNeue-TLF}\selectfont}
\resizebox{\linewidth}{!}{%
\begin{tikzpicture}[x=1cm,y=1cm,line cap=round,line join=round,
  every node/.style={text=wfink,align=center,inner sep=0pt,font=\wfFont},
  panel/.style={fill=white,draw=wfframe,line width=.6pt,rounded corners=2.5pt},
  rolebox/.style={draw=wfrule,line width=.35pt,rounded corners=2pt},
  flow/.style={draw=wfflow,line width=.55pt,-{Stealth[length=2.8pt,width=2.5pt]},rounded corners=2pt},
  wire/.style={draw=wfflow,line width=.55pt,rounded corners=2pt},
  feedback/.style={flow,dash pattern=on 2pt off 1.5pt},
  rolelabel/.style={font=\wfFont\fontsize{6.15}{7}\selectfont\bfseries,anchor=west},
  small/.style={font=\wfFont\fontsize{5.8}{6.6}\selectfont},
  detail/.style={font=\wfFont\fontsize{5.5}{6.3}\selectfont,text=wfmuted},
  stage/.style={font=\wfFont\fontsize{7.1}{8}\selectfont\bfseries},
  chip/.style={draw=wfframe,fill=white,line width=.4pt,rounded corners=1.8pt,
    minimum height=.32cm,font=\wfFont\fontsize{6.0}{7}\selectfont}]
\path[use as bounding box] (-.02,-.36) rectangle (9.22,3.64);
\newcommand{\wfPortrait}[3]{%
  \node[inner sep=0pt] at (#2,#3)
    {\includegraphics[width=.59cm,height=.60cm,keepaspectratio]{figure/agents_3d/#1.png}};}

\draw[panel] (0,0) rectangle (2.88,3.62);
\draw[panel] (3.10,0) rectangle (5.53,3.62);
\draw[panel] (5.75,0) rectangle (9.20,3.62);
\node[stage] at (1.44,3.39) {1\quad Match \& reformulate};
\node[stage] at (4.315,3.39) {2\quad Audit \& repair};
\node[stage] at (7.475,3.39) {3\quad Build \& validate};
\draw[wfrule,line width=.4pt] (.12,3.22)--(2.76,3.22);
\draw[wfrule,line width=.4pt] (3.22,3.22)--(5.41,3.22);
\draw[wfrule,line width=.4pt] (5.87,3.22)--(9.08,3.22);

\draw[rolebox,fill=wfmatch] (.10,2.19) rectangle (2.78,2.86);
\draw[rolebox,fill=wfdesign] (.10,.90) rectangle (2.78,1.57);
\draw[rolebox,fill=wfquality] (3.20,2.19) rectangle (5.43,2.86);
\draw[rolebox,fill=wfrepair] (3.20,.90) rectangle (5.43,1.57);
\draw[rolebox,fill=wfimplement] (5.85,2.19) rectangle (9.10,2.86);
\draw[rolebox,fill=wfreference] (5.85,.77) rectangle (9.10,1.33);

\node[small] at (1.44,3.06)
  {\faIcon{file-alt}\ Problem\; +\; \faIcon{book}\ OptTips};
\draw[flow,-{Stealth[length=1.4pt,width=1.6pt]}] (1.44,2.94)--(1.44,2.88);
\wfPortrait{matcher}{.44}{2.525}
\node[rolelabel] at (.80,2.62) {Technique matcher};
\node[detail,anchor=west] at (.80,2.38) {Select a technique};
\draw[flow] (1.44,2.16)--(1.44,2.03);
\node[detail] at (1.44,1.88) {Technique + operator};
\draw[flow] (1.44,1.74)--(1.44,1.60);
\wfPortrait{designer}{.44}{1.235}
\node[rolelabel,font=\wfFont\fontsize{5.4}{6.3}\selectfont\bfseries]
  at (.80,1.33) {Reformulation designer};
\node[detail,anchor=west] at (.80,1.09) {Redesign the task};
\draw[flow] (1.44,.87)--(1.44,.71);
\node[small,font=\wfFont\fontsize{6}{7}\selectfont\bfseries] at (1.44,.53)
  {Candidate task};
\draw[flow] (2.79,1.33)--(2.99,1.33)--(2.99,2.53)--(3.18,2.53);

\node[detail] at (4.315,3.02) {Check candidate design};
\wfPortrait{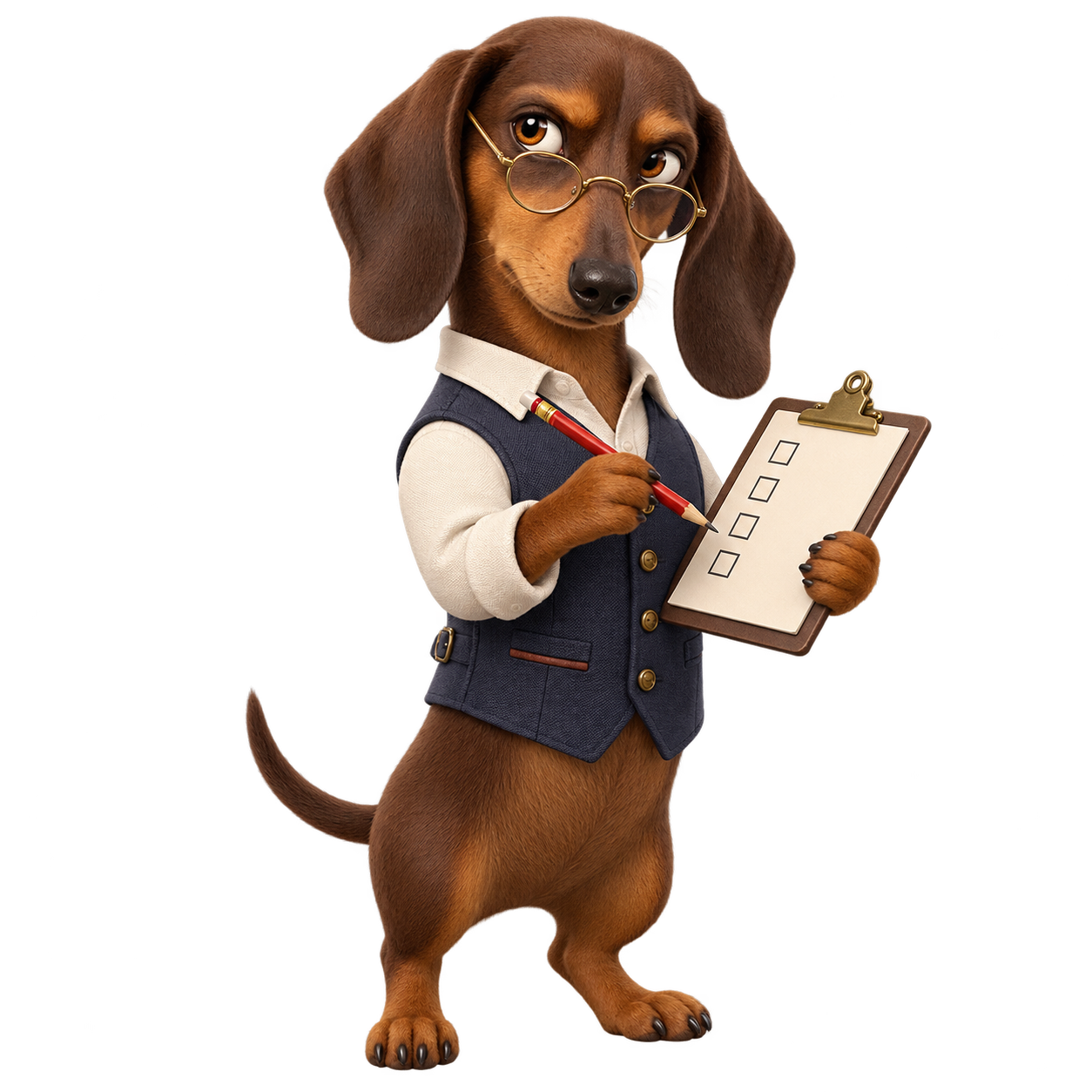}{3.54}{2.525}
\node[rolelabel] at (3.90,2.62) {Quality auditor};
\node[detail,anchor=west] at (3.90,2.38) {Check validity};
\wfPortrait{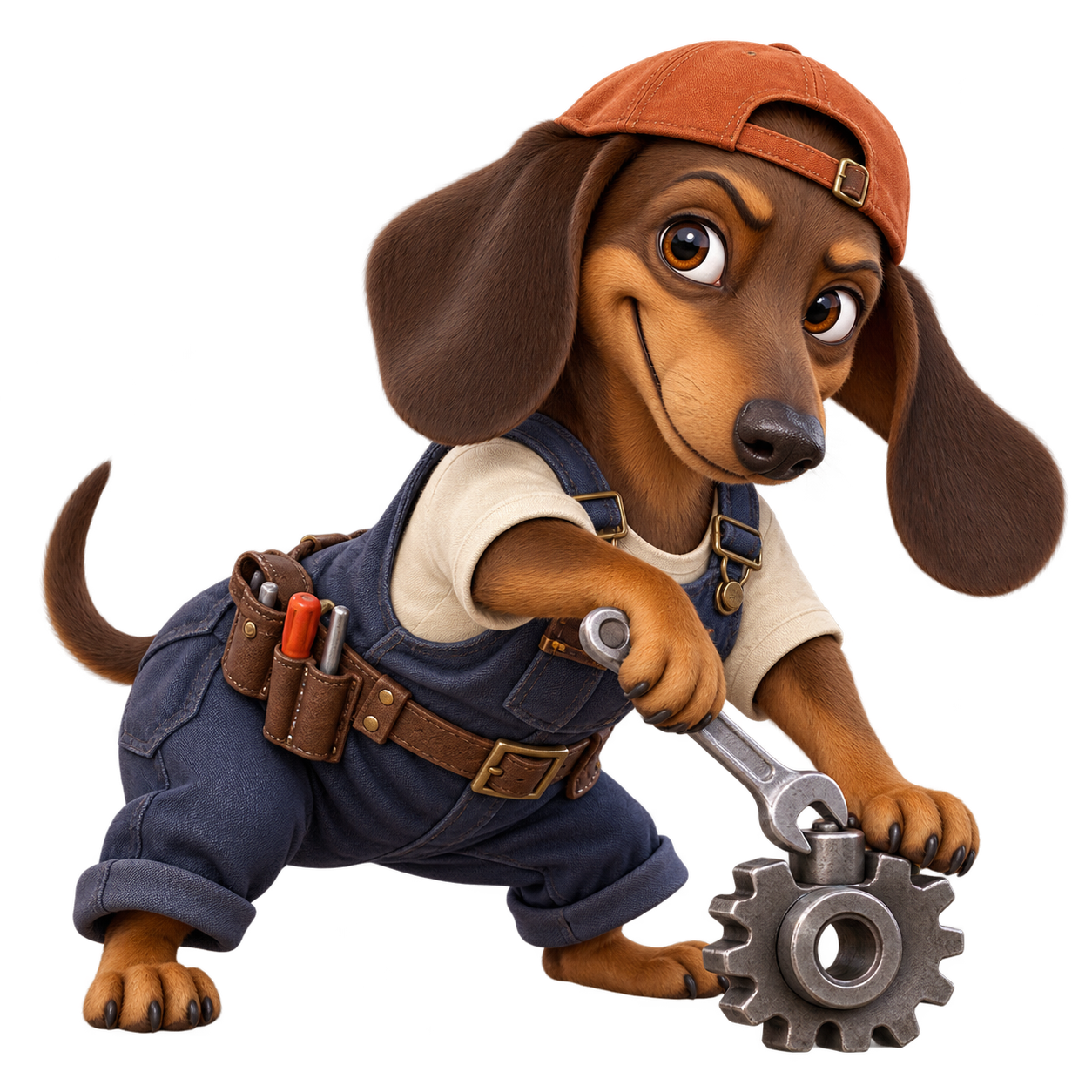}{3.54}{1.235}
\node[rolelabel] at (3.90,1.33) {Repair agent};
\node[detail,anchor=west] at (3.90,1.09) {Fix flagged issues};
\draw[feedback] (3.48,2.16)--(3.48,1.60);
\node[small,anchor=west] at (3.59,1.89) {revise};
\draw[feedback] (5.13,1.60)--(5.13,2.16);
\node[small,anchor=east] at (5.02,1.89) {recheck};
\node[detail,align=center] at (4.315,.53) {Same source + technique};
\draw[flow] (5.44,2.53)--(5.83,2.53);
\node[small,fill=white,inner sep=.5pt] at (5.64,2.70) {pass};

\node[small,text=wfmuted] at (7.475,3.02) {Same fixed instance};
\wfPortrait{implementation}{6.19}{2.525}
\node[rolelabel] at (6.56,2.62) {Implementation agent};
\node[detail,anchor=west] at (6.56,2.38) {Build paired references};
\draw[wire] (7.475,2.16)--(7.475,2.06);
\draw[flow] (7.475,2.06)--(6.60,2.06)--(6.60,1.95);
\draw[flow] (7.475,2.06)--(8.35,2.06)--(8.35,1.95);
\node[chip,minimum width=1.46cm] at (6.60,1.76)
  {\faIcon{file-alt}\ Ordinary};
\node[chip,minimum width=1.46cm] at (8.35,1.76)
  {\faIcon{drafting-compass}\ Expert};
\draw[wire] (6.60,1.57)--(6.60,1.50)--(8.35,1.50)--(8.35,1.57);
\draw[flow] (7.475,1.50)--(7.475,1.35);
\node[inner sep=0pt] at (6.19,1.05)
  {\includegraphics[width=.52cm,height=.49cm,keepaspectratio]{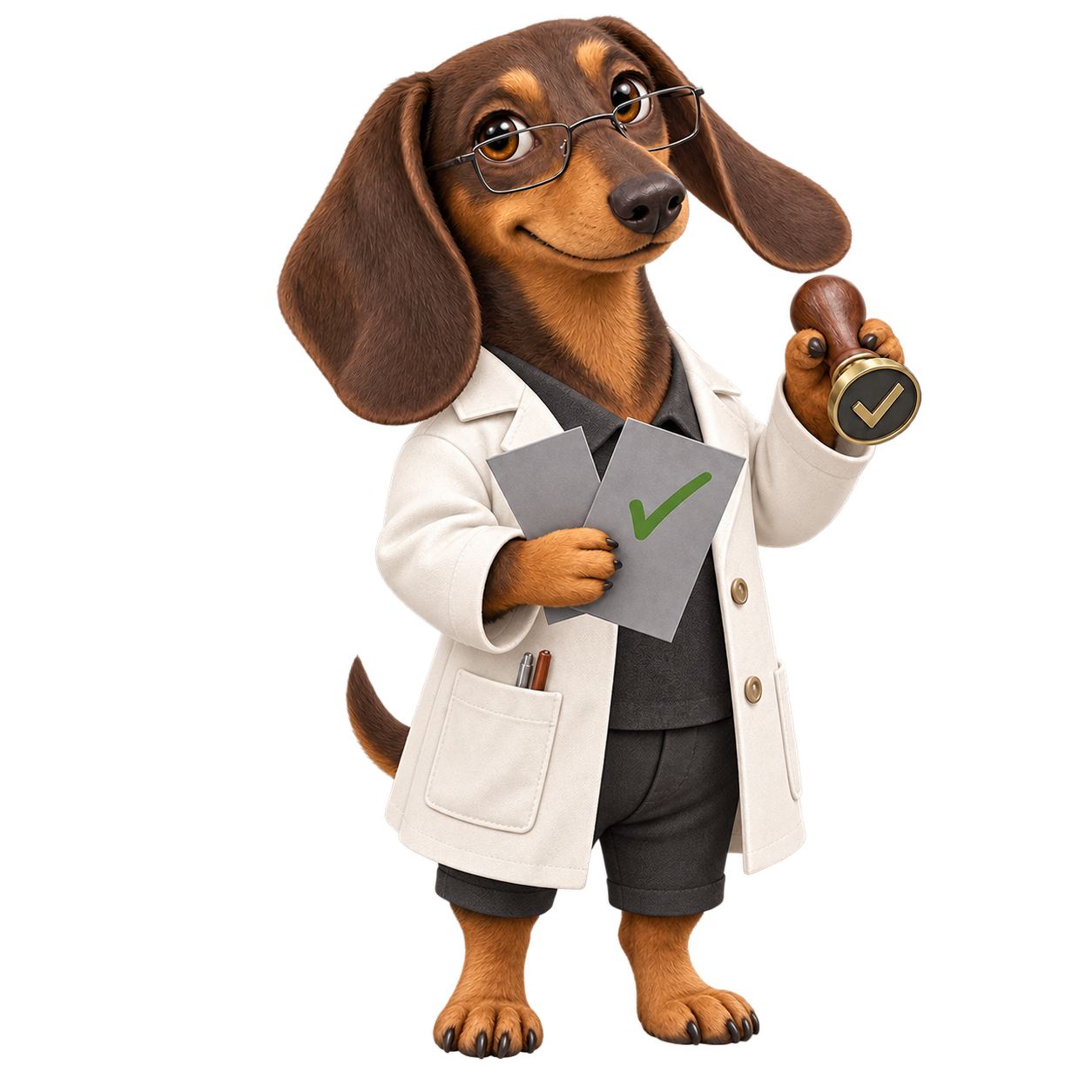}};
\node[rolelabel] at (6.56,1.15) {Reference auditor};
\node[detail,anchor=west] at (6.56,.91) {Check both references};
\draw[flow] (7.475,.74)--(7.475,.68);
\node[small] at (7.475,.55) {\faIcon{user-check}\ Independent expert review};
\draw[flow] (7.475,.42)--(7.475,.34);
\node[chip,minimum height=.26cm,minimum width=3.12cm,fill=black!3,
  font=\wfFont\fontsize{6.1}{7}\selectfont\bfseries] at (7.475,.18)
  {\faIcon{check-circle}\ Validated item};
\end{tikzpicture}}
\endgroup
\caption{\small OptDachshund workflow: from source problems to validated tasks with ordinary and expert references.}
\label{fig:multi-agent}
\vspace{-6pt}
\end{wrapfigure}

OptDachshund is a multi-agent framework that uses OptTips to transform source problems into candidate tasks with ordinary and expert reference implementations. Its six agents work in three stages: \emph{(1) match and reformulate}, \emph{(2) audit and repair}, and \emph{(3) build and validate} (Figure~\ref{fig:multi-agent}).

\noindent\textbf{Source problems.} We draw problems from MAMO EasyLP and ComplexLP \citep{huang2025llms}, and OptMATH \citep{lu2025optmath}. We select self-contained problems that can be adapted to test an OptTips technique, with solutions we can check using the benchmark's solvers. Unclear or incomplete statements and duplicate or near-duplicate templates are flagged for revision or removal.

\noindent\textbf{1) Match and reformulate.} The \emph{Technique Matcher} selects an OptTips technique and proposes how to modify a source problem to test its use. The \emph{Reformulation Designer} then writes the revised problem statement and data schema and prepares a formulation and solution for internal validation. For example, \texttt{T35\_001} tests variable elimination in an allocation problem where each group's allocations must sum to a fixed total. Once all but one allocation are chosen, the remaining allocation is determined by subtraction. Its variable can therefore be replaced by the total minus the other allocations, with its lower and upper bounds imposed on that expression. The public statement gives the task requirements without revealing the intended technique, so the evaluated LLM must choose its own modeling approach.

\noindent\textbf{2) Audit and repair.} The \emph{Quality Auditor} checks that the statement and data define a clear problem. It then checks whether the model correctly represents the stated decisions, constraints, and objective, and whether the selected technique applies. It also looks for duplicate tasks and hints that reveal the intended technique. The \emph{Repair Agent} fixes the reported problems and returns the task for another check. For minor revisions, we keep the source problem and selected technique unchanged. If a revision changes the task requirements or the technique, we repeat the full audit.

\noindent\textbf{3) Build and validate.} After the design audit, the \emph{Reference Implementation Agent} builds ordinary and expert references for the same redesigned task, data, and required outputs. The ordinary reference uses a conventional approach; the expert applies the selected technique. The \emph{Reference Audit Agent} checks their formulations, code, and solver results. Appendix~\ref{app:agents} gives implementation details and an example prompt. At least two experts review each candidate independently: one checks the public specification and another examines both references. A third expert resolves substantive disagreements. Reference runs and optimality evidence verify the answers. We require correct solutions and valid technique use. Appendix~\ref{app:quality-control} explains the review criteria and revision process.

\FloatBarrier
\Needspace{8\baselineskip}
\subsection{EfficientOpt: Benchmark Composition and Interface}
\label{subsec:efficientopt}

\begin{figure}[h]
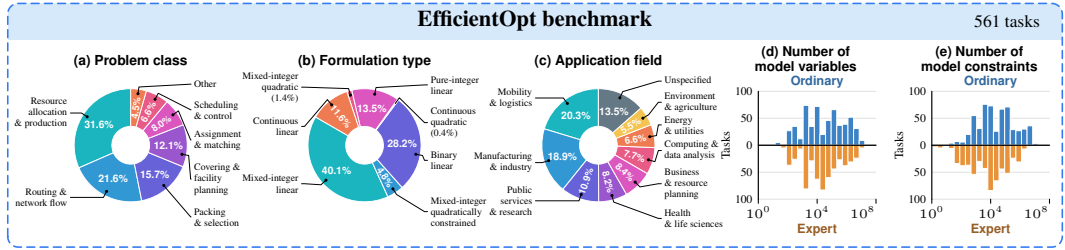

\centering
\begingroup
\definecolor{panelblue}{HTML}{D6E9FB}
\definecolor{blueedge}{HTML}{347AF0}
\resizebox{\textwidth}{!}{%
%
}
\endgroup
\caption{\small Task composition and reference model sizes for the 561 tasks in EfficientOpt. (a--c) Problem classes, formulation types of ordinary references, and application fields. (d,e) Variable and constraint counts for ordinary (blue, above) and expert (orange, below) references, shown on logarithmic horizontal axes. Appendix~\ref{app:dataset_stats} gives the classifications and counting rules. (The visual presentation is inspired by the elegant design of Fig.~1 in \citet{kong2026frontieror}.)}
\label{fig:benchmark_profile}
\end{figure}

\FloatBarrier

The resulting benchmark EfficientOpt contains 561 test tasks. These tasks span different problem classes and application fields, with reference models that vary in formulation type and size (Figure~\ref{fig:benchmark_profile}).
\emph{1) Problem classes.} The tasks cover eight classes, including resource allocation and production (31.6\%), routing and network flow (21.6\%), and scheduling and control (6.6\%). Two small classes are grouped as \emph{Other} in Figure~\ref{fig:benchmark_profile}(a).
\emph{2) Formulation types.} The ordinary reference models fall into seven types, including mixed-integer linear (40.1\%), binary linear (28.2\%), and continuous linear (11.6\%); quadratic formulations are also represented (Figure~\ref{fig:benchmark_profile}(b)).
\emph{3) Application fields.} The tasks cover eight named fields, including mobility and logistics, manufacturing and industry, and health and life sciences. Tasks without a clear application context are labeled \emph{Unspecified} (Figure~\ref{fig:benchmark_profile}(c)).
\emph{4) Reference-model sizes.} Across all 561 tasks, reference model sizes span several orders of magnitude (Figure~\ref{fig:benchmark_profile}(d,e)). We take the maximum variable and total constraint counts over each reference's models before presolve. Expert references have fewer variables in 43.1\% of pairs and fewer constraints in 51.0\%. Appendix~\ref{app:dataset_stats} details the classifications and counting rules.

\noindent\textbf{Task interface.} The task interface specifies what an LLM receives and what it must produce. Each task provides a natural-language problem statement, specifying the required outputs, and a schema for the input data. The LLM writes executable solver code for its chosen formulation. We run the code on the task's fixed numerical instance using Gurobi. The technique label, both reference implementations, and validation records are hidden from the LLM.

\noindent\textbf{Paired references.} The ordinary reference uses a conventional approach, while the expert reference applies the selected OptTips technique to the same task and data. We use verified task objectives to check the LLM's numerical answer and reference execution records to compare construction and solver costs separately. An LLM judge assesses technique use from code, with manual inspections of selected cases (Appendices~\ref{app:judge-full} and~\ref{app:technique-cases}). A different approach can also produce a correct, efficient solution. Section~\ref{sec:end-to-end} defines the numerical checks and cost measures.

\FloatBarrier

\section{Experiments}
\label{sec:experiments}

We evaluate LLM-generated programs in terms of numerical correctness, computational cost, and modeling choices. Four questions guide our analysis: \textbf{RQ1:} How does numerical accuracy vary across tasks? \textbf{RQ2:} For correctly solved tasks, how do computational costs compare with the ordinary and expert references? \textbf{RQ3:} Do smaller models solve faster, and which modeling techniques do LLMs use? \textbf{RQ4:} What time and memory costs are missed when we measure only solver time?

\subsection{Experimental Setup}
\label{sec:end-to-end}

\noindent\textbf{Tasks and evaluation.} We evaluate the 11 general-purpose LLMs in Table~\ref{tab:main-results-561} on all 561 EfficientOpt tasks.
We executed each task three times on the same machine with the same settings. Recorded results varied slightly, so the main tables report a typical run.
Each LLM generates solver code from the problem description.
We count a run as numerically correct if Gurobi returns \texttt{OPTIMAL} and its finite objective matches the verified reference value (absolute and relative tolerances: $10^{-6}$). Appendix~\ref{app:dataset_stats} details task statistics, code generation, and answer verification; Appendix~\ref{app:full-results-561} adds accuracy breakdowns and detailed cost analyses. We also evaluate five fine-tuned models. Their low accuracy limits efficiency comparisons, so we report their accuracy and failure analysis separately in Appendix~\ref{app:specialized-open-models}.

\noindent\textbf{Evaluation metrics.} We report accuracy, technique use, and code generation and execution costs:
\begin{list}{\textbullet}{\setlength{\leftmargin}{1.2em}\setlength{\labelsep}{0.4em}\setlength{\topsep}{3pt}\setlength{\partopsep}{0pt}\setlength{\itemsep}{0pt}\setlength{\parsep}{0pt}}
\item Accuracy (\%): percentage of all 561 tasks solved correctly under the numerical check above.
\item \texttt{Runtime} (s): elapsed optimization time reported by Gurobi.
\item \texttt{Work} (units): Gurobi's measure of the computation performed by the solver.
\item \texttt{Build} (s): recorded preparation time, including preprocessing and any solves within that stage.
\item \texttt{Build+opt} (s): \texttt{Build} plus elapsed time in the separately timed solver calls.
\item \texttt{API latency} (s): time for the final generation call, including service and network delays.
\item \texttt{Tokens}: provider-reported input, output, and, where available, reasoning token counts.
\item \texttt{RSS}: RAM occupied by the process, sampled during model preparation and solving.
\item \texttt{Gurobi peak}: peak memory allocated within the Gurobi environment.
\end{list}
Appendix~\ref{app:measurement-detail} gives the measurement boundaries and comparison rules.

\begin{samepage}
\subsection{Numerical Outcomes Vary Across Tasks (RQ1)}
\label{subsec:correctness_results}

\noindent\textbf{An optimal solver status is not sufficient.} Gurobi solves the supplied model, which may omit required constraints. Of 6,171 generated programs, 4,602 report \texttt{OPTIMAL}, but 338 of these return incorrect objective values. We therefore verify numerical answers before comparing costs: solving an incorrect model quickly does not establish an efficiency gain.
\end{samepage}

\noindent\textbf{Overall accuracy hides large differences across tasks.} The models are much more successful on some task types than others. Across the evaluated models, accuracy is 93.9\% on tasks that split into independent subproblems, but only 27.3\% on the five tasks targeting column generation (adding variables as needed during optimization). Success on one kind of problem therefore does not ensure reliable answers on another. Results by task group show where errors recur and closer inspection is needed (see Appendix~\ref{app:group-outcomes} for accuracy breakdowns by task group).

\subsection{Correct Answers Can Still Be Costly (RQ2)}
\label{sec:paired-costs}

\leavevmode\vspace{-2\baselineskip}
\begin{wraptable}[18]{r}{0.60\textwidth}
\vspace{-8pt}
\centering\footnotesize
\setlength{\tabcolsep}{2.0pt}
\caption{\footnotesize Numerical accuracy and computational cost. Accuracy uses all 561 tasks. Cost ratios use complete measurements on correctly solved tasks within the 543-task reference cost subset (LLM/reference; below 1 means lower cost). See Appendix~\ref{app:measurement-detail} for metric definitions and cost-ratio calculations.}
\label{tab:main-results-561}
\resizebox{\linewidth}{!}{%
\begin{tabular}{lrrrrrc}
\toprule
 & \multicolumn{2}{c}{Correct answers} & \multicolumn{4}{c}{Cost ratio (LLM/reference)} \\
\cmidrule(lr){2-3}\cmidrule(lr){4-7}
 & & & \multicolumn{2}{c}{\texttt{Runtime}} & \texttt{Work} & \texttt{Build+opt} \\
\cmidrule(lr){4-5}
Model & Count\,$\uparrow$ & Rate (\%)\,$\uparrow$ & Ordinary\,$\downarrow$ & Expert\,$\downarrow$ & Expert\,$\downarrow$ & Expert\,$\downarrow$ \\
\midrule
Gemini 3.1 Pro & 512 & 91.3 & 0.52 & 1.68 & 1.98 & 1.43 \\
GPT-5.5 & 447 & 79.7 & 0.58 & 1.79 & 2.08 & 1.64 \\
Claude Opus 4.6 & 442 & 78.8 & 0.46 & 1.49 & 1.77 & 1.42 \\
DeepSeek-V4 Flash & 441 & 78.6 & 0.49 & 1.62 & 1.78 & 1.47 \\
Kimi K2.6 & 433 & 77.2 & 0.55 & 1.79 & 2.02 & 1.59 \\
Qwen 3.6 27B & 394 & 70.2 & 0.61 & 1.94 & 1.97 & 1.99 \\
GLM-5.1 & 390 & 69.5 & 0.50 & 1.68 & 1.92 & 1.44 \\
Qwen 3.5 122B & 369 & 65.8 & 0.55 & 1.82 & 1.99 & 1.73 \\
MiniMax M2.5 & 315 & 56.1 & 0.59 & 1.80 & 2.01 & 1.77 \\
Qwen 3 32B & 313 & 55.8 & 0.66 & 2.03 & 2.08 & 1.83 \\
Qwen 3.6 Plus & 208 & 37.1 & 0.52 & 1.83 & 2.22 & 1.63 \\
\bottomrule
\end{tabular}}
\vspace{-6pt}
\end{wraptable}

\noindent\textbf{Correct answers still leave room for faster solving.} Table~\ref{tab:main-results-561} compares costs with both references on each LLM's eligible tasks. All 11 LLMs have lower aggregate \texttt{Runtime} than the ordinary reference, but higher \texttt{Runtime} than the expert (ratios 1.49--2.03). Higher \texttt{Work} ratios show that this gap also appears in measured solver effort. Including model preparation does not close it: every \texttt{Build+opt} ratio against the expert remains above one. Thus, outperforming a conventional implementation leaves room to reduce computational cost. Using both references makes this visible; an ordinary-reference comparison alone would miss the remaining gap to expert performance.

\noindent\textbf{A few slow solutions account for most solving time.} For each LLM, the slowest $\sim$10\% of runs with correct objective values and valid timings consume 68.4--76.2\% of total \texttt{Runtime} (Figure~\ref{fig:runtime-concentration}). Median runtimes are 23.1--36.0 seconds, but summed \texttt{Runtime} over each model's eligible tasks is 12.0--27.7 hours. Appendix~\ref{app:slow-task-analysis} identifies the slow tasks for all 11 models and compares them with the expert references. As a sensitivity calculation, halving only the slowest tenth would reduce total solver time by 34.2--38.1\% without changing any model's median.

\finishwraptable

\begingroup
\setlength{\intextsep}{4pt plus 1pt minus 1pt}
\begin{figure}[h]
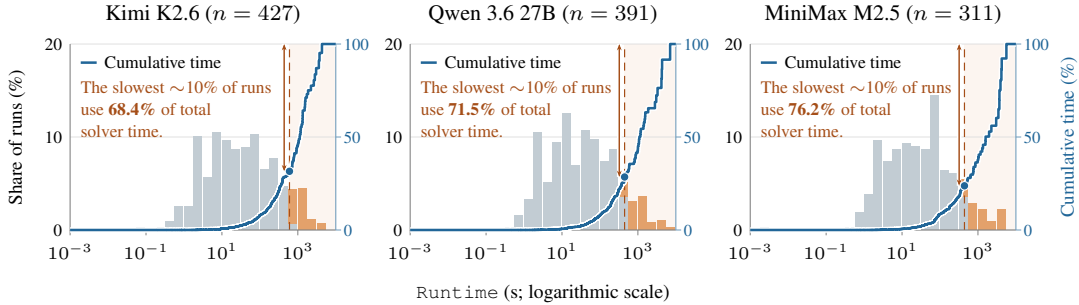

\centering
\begingroup
\definecolor{histrest}{HTML}{C2CCD3}
\definecolor{histtail}{HTML}{E3A16B}
\definecolor{histannot}{HTML}{A54E18}
\definecolor{timecurve}{HTML}{216698}

\endgroup
\caption{\footnotesize Runtime distributions and cumulative time. Bars show the percentage of runs in each time interval (left axis). Blue curves show the share of total \texttt{Runtime} used by runs at or below each runtime (right axis). Orange highlights the slowest $\sim$10\%; arrows show their contribution to total time. Only correctly solved tasks with valid timings are included. See Appendix~\ref{app:cost-distributions} for per-model runtime statistics.}
\label{fig:runtime-concentration}
\end{figure}
\endgroup

\subsection{Modeling Choices and Solver Efficiency (RQ3)}
\label{sec:structural_diagnostics}

\noindent\textbf{Smaller models do not necessarily solve faster.} Of the 735 outputs that passed our numerical checks and had smaller reported models than their expert references, 449 (61.1\%) still took longer to solve (Figure~\ref{fig:diagnostic-costs}(a)). Variable and constraint counts describe model size, but they do not fully capture how difficult the problem is for the solver. We therefore measure \texttt{Runtime} directly to assess whether a smaller model is faster to solve. Appendix~\ref{app:structure-diagnostic} reports per-model results under two size criteria and examines the magnitude of the slowdowns.

\begingroup
\begin{figure}[htbp]
\centering
\begin{minipage}[t]{0.53\textwidth}
\centering
\begingroup
\definecolor{diagradiallight}{HTML}{397AA5}
\definecolor{diagradialdark}{HTML}{08306B}
\resizebox{!}{85pt}{%
\begin{tikzpicture}[x=1cm,y=.9cm,font=\fontsize{8.2}{9.2}\selectfont]
\path[use as bounding box] (0,-.30) rectangle (8.02,-4.08);
\node[anchor=north west,inner sep=0pt,font=\bfseries\fontsize{7.5}{8.5}\selectfont] at (0.05,-.40) {(a)};
\draw[black!10,line width=.3pt] (3.9000,-.34)--(3.9000,-3.69);
\draw[black!10,line width=.3pt] (5.2000,-.34)--(5.2000,-3.69);
\draw[black!10,line width=.3pt] (6.5000,-.34)--(6.5000,-3.69);
\draw[black!10,line width=.3pt] (7.8000,-.34)--(7.8000,-3.69);
\path[fill=diagradialdark!0.0000!diagradiallight,draw=none] (2.60,-0.3910) rectangle (5.2366,-0.6090);
\node[anchor=east,inner sep=0pt,font=\fontsize{6.5}{7.3}\selectfont] at (2.45,-0.5000) {GLM-5.1};
\node[anchor=west,inner sep=0pt,text=white,font=\bfseries\fontsize{6.2}{7.0}\selectfont] at (2.78,-0.5000) {36/71 (50.7\%)};
\path[fill=diagradialdark!6.0177!diagradiallight,draw=none] (2.60,-0.6550) rectangle (5.3322,-0.8730);
\node[anchor=east,inner sep=0pt,font=\fontsize{6.5}{7.3}\selectfont] at (2.45,-0.7640) {DeepSeek-V4 Flash};
\node[anchor=west,inner sep=0pt,text=white,font=\bfseries\fontsize{6.2}{7.0}\selectfont] at (2.78,-0.7640) {31/59 (52.5\%)};
\path[fill=diagradialdark!9.6718!diagradiallight,draw=none] (2.60,-0.9190) rectangle (5.3902,-1.1370);
\node[anchor=east,inner sep=0pt,font=\fontsize{6.5}{7.3}\selectfont] at (2.45,-1.0280) {Claude Opus 4.6};
\node[anchor=west,inner sep=0pt,text=white,font=\bfseries\fontsize{6.2}{7.0}\selectfont] at (2.78,-1.0280) {44/82 (53.7\%)};
\path[fill=diagradialdark!14.0634!diagradiallight,draw=none] (2.60,-1.1830) rectangle (5.4600,-1.4010);
\node[anchor=east,inner sep=0pt,font=\fontsize{6.5}{7.3}\selectfont] at (2.45,-1.2920) {Gemini 3.1 Pro};
\node[anchor=west,inner sep=0pt,text=white,font=\bfseries\fontsize{6.2}{7.0}\selectfont] at (2.78,-1.2920) {55/100 (55.0\%)};
\path[fill=diagradialdark!15.2326!diagradiallight,draw=none] (2.60,-1.4470) rectangle (5.4786,-1.6650);
\node[anchor=east,inner sep=0pt,font=\fontsize{6.5}{7.3}\selectfont] at (2.45,-1.5560) {MiniMax M2.5};
\node[anchor=west,inner sep=0pt,text=white,font=\bfseries\fontsize{6.2}{7.0}\selectfont] at (2.78,-1.5560) {31/56 (55.4\%)};
\path[fill=diagradialdark!21.0786!diagradiallight,draw=none] (2.60,-1.7110) rectangle (5.5714,-1.9290);
\node[anchor=east,inner sep=0pt,font=\fontsize{6.5}{7.3}\selectfont] at (2.45,-1.8200) {Kimi K2.6};
\node[anchor=west,inner sep=0pt,text=white,font=\bfseries\fontsize{6.2}{7.0}\selectfont] at (2.78,-1.8200) {40/70 (57.1\%)};
\path[fill=diagradialdark!23.1572!diagradiallight,draw=none] (2.60,-1.9750) rectangle (5.6044,-2.1930);
\node[anchor=east,inner sep=0pt,font=\fontsize{6.5}{7.3}\selectfont] at (2.45,-2.0840) {GPT-5.5};
\node[anchor=west,inner sep=0pt,text=white,font=\bfseries\fontsize{6.2}{7.0}\selectfont] at (2.78,-2.0840) {26/45 (57.8\%)};
\path[fill=diagradialdark!63.1700!diagradiallight,draw=none] (2.60,-2.2390) rectangle (6.2400,-2.4570);
\node[anchor=east,inner sep=0pt,font=\fontsize{6.5}{7.3}\selectfont] at (2.45,-2.3480) {Qwen 3.6 27B};
\node[anchor=west,inner sep=0pt,text=white,font=\bfseries\fontsize{6.2}{7.0}\selectfont] at (2.78,-2.3480) {56/80 (70.0\%)};
\path[fill=diagradialdark!64.0548!diagradiallight,draw=none] (2.60,-2.5030) rectangle (6.2541,-2.7210);
\node[anchor=east,inner sep=0pt,font=\fontsize{6.5}{7.3}\selectfont] at (2.45,-2.6120) {Qwen 3.5 122B};
\node[anchor=west,inner sep=0pt,text=white,font=\bfseries\fontsize{6.2}{7.0}\selectfont] at (2.78,-2.6120) {52/74 (70.3\%)};
\path[fill=diagradialdark!91.9396!diagradiallight,draw=none] (2.60,-2.7670) rectangle (6.6970,-2.9850);
\node[anchor=east,inner sep=0pt,font=\fontsize{6.5}{7.3}\selectfont] at (2.45,-2.8760) {Qwen 3 32B};
\node[anchor=west,inner sep=0pt,text=white,font=\bfseries\fontsize{6.2}{7.0}\selectfont] at (2.78,-2.8760) {52/66 (78.8\%)};
\path[fill=diagradialdark!100.0000!diagradiallight,draw=none] (2.60,-3.0310) rectangle (6.8250,-3.2490);
\node[anchor=east,inner sep=0pt,font=\fontsize{6.5}{7.3}\selectfont] at (2.45,-3.1400) {Qwen 3.6 Plus};
\node[anchor=west,inner sep=0pt,text=white,font=\bfseries\fontsize{6.2}{7.0}\selectfont] at (2.78,-3.1400) {26/32 (81.2\%)};
\draw[black!20,line width=.3pt] (2.60,-3.335)--(7.80,-3.335);
\path[fill=diagradialdark!33.9956!diagradiallight,draw=none] (2.60,-3.391) rectangle (5.7766,-3.609);
\begin{scope}
\clip (2.60,-3.391) rectangle (5.7766,-3.609);
\foreach \xx in {2.35,2.47,...,5.83}{
  \draw[white,opacity=.22,line width=.5pt] (\xx,-3.659)--++(.35,.40);
}
\end{scope}
\node[anchor=east,inner sep=0pt,font=\bfseries\fontsize{6.5}{7.3}\selectfont] at (2.45,-3.500) {Overall};
\node[anchor=west,inner sep=0pt,text=white,font=\bfseries\fontsize{6.2}{7.0}\selectfont] at (2.78,-3.500) {449/735 (61.1\%)};
\draw[black!45,line width=.4pt] (2.60,-.34)--(2.60,-3.72)--(7.80,-3.72);
\draw[black!45,line width=.4pt] (2.6000,-3.72)--(2.6000,-3.78);
\node[anchor=north,inner sep=1pt,text=black!65,font=\fontsize{5.8}{6.6}\selectfont] at (2.6000,-3.80) {0\%};
\draw[black!45,line width=.4pt] (3.9000,-3.72)--(3.9000,-3.78);
\node[anchor=north,inner sep=1pt,text=black!65,font=\fontsize{5.8}{6.6}\selectfont] at (3.9000,-3.80) {25\%};
\draw[black!45,line width=.4pt] (5.2000,-3.72)--(5.2000,-3.78);
\node[anchor=north,inner sep=1pt,text=black!65,font=\fontsize{5.8}{6.6}\selectfont] at (5.2000,-3.80) {50\%};
\draw[black!45,line width=.4pt] (6.5000,-3.72)--(6.5000,-3.78);
\node[anchor=north,inner sep=1pt,text=black!65,font=\fontsize{5.8}{6.6}\selectfont] at (6.5000,-3.80) {75\%};
\draw[black!45,line width=.4pt] (7.8000,-3.72)--(7.8000,-3.78);
\node[anchor=north,inner sep=1pt,text=black!65,font=\fontsize{5.8}{6.6}\selectfont] at (7.8000,-3.80) {100\%};
\end{tikzpicture}%
}
\endgroup
\end{minipage}\hfill
\begin{minipage}[t]{0.45\textwidth}
\centering
\begingroup
\definecolor{judgetarget}{HTML}{009BC3}
\definecolor{judgealternative}{HTML}{F70C54}
\definecolor{judgenone}{HTML}{FDCB04}
\definecolor{judgena}{HTML}{019E61}
\begin{tikzpicture}[x=1pt,y=1pt,font=\fontsize{7.2}{8.1}\selectfont,every node/.style={inner sep=0pt,outer sep=0pt}]
\path[use as bounding box] (0,0) rectangle (170,85);
\node[anchor=north west,font=\bfseries\fontsize{7.5}{8.5}\selectfont] at (0,83) {(b)};
\begin{scope}[shift={(74,41.5)},scale=.82]
\path[fill=judgetarget,draw=white,line width=.45pt] (90.000000000:38pt) arc[start angle=90.000000000,end angle=256.728245017,radius=38pt] -- (256.728245017:12.6667pt) arc[start angle=256.728245017,end angle=90.000000000,radius=12.6667pt] -- cycle;
\path[fill=judgealternative,draw=white,line width=.45pt] (256.728245017:38pt) arc[start angle=256.728245017,end angle=345.284394750,radius=38pt] -- (345.284394750:12.6667pt) arc[start angle=345.284394750,end angle=256.728245017,radius=12.6667pt] -- cycle;
\path[fill=judgenone,draw=white,line width=.45pt] (345.284394750:38pt) arc[start angle=345.284394750,end angle=421.998055421,radius=38pt] -- (421.998055421:12.6667pt) arc[start angle=421.998055421,end angle=345.284394750,radius=12.6667pt] -- cycle;
\path[fill=judgena,draw=white,line width=.45pt] (421.998055421:38pt) arc[start angle=421.998055421,end angle=450.000000000,radius=38pt] -- (450.000000000:12.6667pt) arc[start angle=450.000000000,end angle=421.998055421,radius=12.6667pt] -- cycle;
\node[text=white,font={\sffamily\bfseries\fontsize{6.4}{7.2}\selectfont}] at (173.364122509:25.3333pt) {46.3\%};
\node[text=white,font={\sffamily\bfseries\fontsize{6.4}{7.2}\selectfont}] at (301.006319883:25.3333pt) {24.6\%};
\node[text=black!85,font={\sffamily\bfseries\fontsize{6.4}{7.2}\selectfont}] at (23.641225085:25.3333pt) {21.3\%};
\node[text=white,font={\sffamily\bfseries\fontsize{6.1}{6.8}\selectfont},rotate=75.999028] at (75.999027710:25.3333pt) {7.8\%};
\draw[black,line width=.4pt] (-28.637555,20.052232) -- (-32.372889,22.667741) -- (-48.200000,26.500000);
\fill[black] (-28.637555,20.052232) circle[radius=.65pt];
\node[anchor=east] at (-50,26.5) {Target};
\draw[black,line width=.4pt] (18.009036,-29.964583) -- (20.358041,-33.873006) -- (46.200000,-37.500000);
\fill[black] (18.009036,-29.964583) circle[radius=.65pt];
\node[anchor=west] at (48,-37.5) {Alternative};
\draw[black,line width=.4pt] (34.960000,0.000000) -- (39.520000,0.000000) -- (46.200000,11.500000);
\fill[black] (34.960000,0.000000) circle[radius=.65pt];
\node[anchor=west] at (48,11.5) {None};
\draw[black,line width=.4pt] (1.829665,34.912089) -- (2.068317,39.465839) -- (46.200000,40.000000);
\fill[black] (1.829665,34.912089) circle[radius=.65pt];
\node[anchor=west] at (48,40) {N/A};
\end{scope}
\end{tikzpicture}
\endgroup
\end{minipage}
\caption{\small Size and technique use. (a) Among correct programs with smaller formulations, how many are slower than the expert? Labels give slower/total; the striped bar combines all 11 LLMs. (b) Technique-use labels for 6,171 programs. Appendix~\ref{app:structure-diagnostic} analyzes size and runtime; Appendix~\ref{app:judge-full} details technique use.}
\label{fig:diagnostic-costs}
\end{figure}
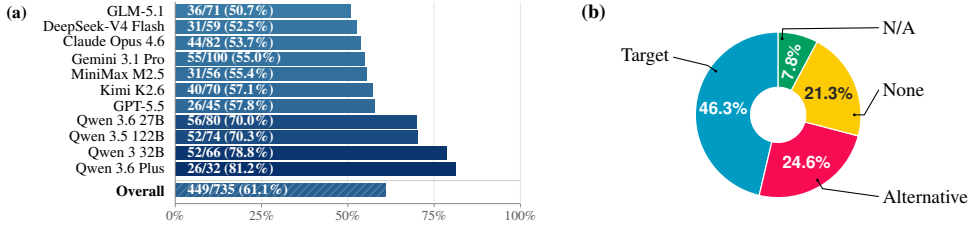
\endgroup

\noindent\textbf{The same answer can come from different modeling choices.} In department relocation (T10\_011), GPT-5.5 adds binary variables to represent pairs of department assignments. The expert represents these interactions with continuous variables linked to the original assignments, reducing the binary count from 5,472 to 288. These formulations differ in variable types and linking constraints, distinctions that objective values and runtimes alone do not reveal. Appendix~\ref{app:case-formulations} explains their integer equivalence and different LP relaxations.

\noindent\textbf{LLMs use different modeling approaches.} Each task has a designated OptTips technique. Our LLM-based assessment identifies it or an equivalent implementation in 46.3\% of programs, and another technique in 24.6\% (Figure~\ref{fig:diagnostic-costs}(b)). These labels do not establish correctness or speed. For T13\_006, Gemini models technology allocations directly, while the expert reformulates the problem using resource prices. Both return the same objective, with recorded preparation and solving times (\texttt{Build+opt}) of 16.18 and 16.27 seconds. Appendix~\ref{app:judge-full} gives technique-use breakdowns by model and task group; Appendix~\ref{app:efficiency-audit} examines the costs and limits of this comparison.

\vspace{-6pt}
\subsection{Solving Is Only Part of the Cost (RQ4)}
\label{sec:cost-shifting}

\noindent\textbf{Model preparation can change which implementation is faster.} Among 4,202 correct runs in the cost subset, 17.3\% spend longer on \texttt{Build} than on optimization. Adding \texttt{Build} to the externally timed optimization stage reverses the LLM--expert speed comparison in 8.6\% of 4,148 pairs. These reversals occur for all 11 LLMs. Appendix~\ref{app:resource-distributions} gives per-model results and sensitivity checks; Appendix~\ref{app:case-construction-costs} shows how construction offsets faster solving on T20\_011.

\noindent\textbf{Waiting for generated code can take longer than running it.} Across all available responses, per-model median \texttt{API latency} ranges from 14.0 to 231.2 seconds. Among 4,202 correct runs with both measurements, generation takes longer than recorded construction and solving (\texttt{Build+opt}) in 55.6\% of cases, rising to 76.2\% for Qwen 3.6 27B and 74.1\% for Qwen 3 32B. Appendix~\ref{app:resource-distributions} compares generation latency and token profiles across models.

\noindent\textbf{Typical memory use can hide demanding cases.} Across all 4,202 correct runs in the cost analysis, median \texttt{Gurobi peak} is 0.135~GB, while the 90th percentile is 2.314~GB and 224 runs (5.3\%) reach at least 4~GB. Appendix~\ref{app:resource-distributions} compares solver and process memory across models.

\vspace{-6pt}
\section{Discussion and Conclusion}

We introduced OptTips, which codifies 50 expert techniques in knowledge cards linking problem structure and inefficiency symptoms to concrete modeling actions. OptDachshund turns this knowledge into a six-agent workflow for task reformulation, auditing, repair, and paired reference construction and validation. EfficientOpt provides 561 tasks with ordinary and expert implementations on the same instances, allowing numerical correctness, technique use, and computational cost to be evaluated separately. Across the 11 evaluated LLMs, programs returning correct objective values use less aggregate Gurobi solver time than ordinary references but more than expert references on comparable task subsets. Smaller formulations can still solve more slowly, and model preparation can offset faster solving. Future work can extend this evaluation across solvers, instance scales, and iterative methods. Progress should be assessed by both solution correctness and computational cost.
\label{sec:main-end}

\bibliography{iclr2027_conference}

@book{williams2013model,
  author={Williams, H. Paul},
  title={Model Building in Mathematical Programming},
  edition={5th},
  publisher={John Wiley \& Sons},
  year={2013},
  isbn={9781118443330},
  sourceurl={https://www.wiley-vch.de/en/areas-interest/finance-economics-law/model-building-in-mathematical-programming-978-1-118-44333-0}
}

@book{pinedo2022scheduling,
  author={Pinedo, Michael L.},
  title={Scheduling: Theory, Algorithms, and Systems},
  edition={6th},
  publisher={Springer},
  year={2022},
  doi={10.1007/978-3-031-05921-6},
  isbn={9783031059209}
}

@book{toth2014vehicle,
  editor={Toth, Paolo and Vigo, Daniele},
  title={Vehicle Routing: Problems, Methods, and Applications},
  edition={2nd},
  publisher={Society for Industrial and Applied Mathematics},
  series={MOS-SIAM Series on Optimization},
  volume={18},
  year={2014},
  doi={10.1137/1.9781611973594},
  isbn={9781611973587}
}

@book{conejo2010decision,
  author={Conejo, Antonio J. and Carri{\'o}n, Miguel and Morales, Juan M.},
  title={Decision Making Under Uncertainty in Electricity Markets},
  publisher={Springer},
  series={International Series in Operations Research \& Management Science},
  volume={153},
  year={2010},
  doi={10.1007/978-1-4419-7421-1},
  isbn={9781441974204}
}

@article{xiao2025survey,
  title={A Survey of Optimization Modeling Meets {LLMs}: Progress and Future Directions},
  author={Xiao, Ziyang and Xie, Jingrong and Xu, Lilin and Guan, Shisi and Zhu, Jingyan and Han, Xiongwei and Fu, Xiaojin and Yu, WingYin and Wu, Han and Shi, Wei and Kang, Qingcan and Duan, Jiahui and Zhong, Tao and Yuan, Mingxuan and Zeng, Jia and Wang, Yuan and Chen, Gang and Zhang, Dongxiang},
  journal={arXiv preprint arXiv:2508.10047},
  year={2025},
  doi={10.48550/arXiv.2508.10047},
  sourceurl={https://arxiv.org/abs/2508.10047}
}

@inproceedings{ramamonjison2023nl4opt,
  title={{NL4Opt} Competition: Formulating Optimization Problems Based on Their Natural Language Descriptions},
  author={Ramamonjison, Rindranirina and Yu, Timothy and Li, Raymond and Li, Haley and Carenini, Giuseppe and Ghaddar, Bissan and He, Shiqi and Mostajabdaveh, Mahdi and Banitalebi-Dehkordi, Amin and Zhou, Zirui and Zhang, Yong},
  booktitle={Proceedings of the NeurIPS 2022 Competitions Track},
  volume={220},
  series={Proceedings of Machine Learning Research},
  pages={189--203},
  publisher={PMLR},
  year={2022},
  sourceurl={https://proceedings.mlr.press/v220/ramamonjison23a.html}
}

@inproceedings{
xiao2024chain,
title={Chain-of-Experts: When {LLM}s Meet Complex Operations Research Problems},
author={Ziyang Xiao and Dongxiang Zhang and Yangjun Wu and Lilin Xu and Yuan Jessica Wang and Xiongwei Han and Xiaojin Fu and Tao Zhong and Jia Zeng and Mingli Song and Gang Chen},
booktitle={The Twelfth International Conference on Learning Representations},
year={2024}
}

@article{lu2026pearl,
  title={PEARL: Solver-in-the-Loop Interactive Optimization Modeling from Natural Language},
  author={Lu, Hongliang and Li, Zhong and Chen, Yuxuan and Lan, Yuan and Zhang, Fan and Wen, Zaiwen},
  journal={arXiv preprint arXiv:2607.18256},
  year={2026}
}

@inproceedings{ahmaditeshnizi2024optimus,
  title={{OptiMUS}: Scalable Optimization Modeling with {(MI)LP} Solvers and Large Language Models},
  author={AhmadiTeshnizi, Ali and Gao, Wenzhi and Udell, Madeleine},
  booktitle={Proceedings of the 41st International Conference on Machine Learning},
  volume={235},
  series={Proceedings of Machine Learning Research},
  pages={577--596},
  publisher={PMLR},
  year={2024},
  sourceurl={https://proceedings.mlr.press/v235/ahmaditeshnizi24a.html}
}

@inproceedings{
jiang2025llmopt,
title={{LLMOPT}: Learning to Define and Solve General Optimization Problems from Scratch},
author={Caigao Jiang and Xiang Shu and Hong Qian and Xingyu Lu and Jun Zhou and Aimin Zhou and Yang Yu},
booktitle={The Thirteenth International Conference on Learning Representations},
year={2025},
sourceurl={https://proceedings.iclr.cc/paper_files/paper/2025/hash/fbe6dd68b0cf2b1d43b458d2b8ca31b0-Abstract-Conference.html}
}

@article{huang2025orlm,
  title={{ORLM}: A Customizable Framework in Training Large Models for Automated Optimization Modeling},
  author={Huang, Chenyu and Tang, Zhengyang and Hu, Shixi and Jiang, Ruoqing and Zheng, Xin and Ge, Dongdong and Wang, Benyou and Wang, Zizhuo},
  journal={Operations Research},
  volume={73},
  number={6},
  pages={2986--3009},
  year={2025},
  publisher={INFORMS},
  doi={10.1287/opre.2024.1233},
  sourceurl={https://pubsonline.informs.org/doi/abs/10.1287/opre.2024.1233}
}

@inproceedings{lu2025optmath,
title={{OptMATH}: A Scalable Bidirectional Data Synthesis Framework for Optimization Modeling},
author={Hongliang Lu and Zhonglin Xie and Yaoyu Wu and Can Ren and Yuxuan Chen and Zaiwen Wen},
booktitle={Proceedings of the 42nd International Conference on Machine Learning},
volume={267},
series={Proceedings of Machine Learning Research},
pages={40769--40802},
publisher={PMLR},
year={2025},
sourceurl={https://proceedings.mlr.press/v267/lu25o.html}
}

@inproceedings{yang2025optibench,
  title={{OptiBench} Meets {ReSocratic}: Measure and Improve {LLMs} for Optimization Modeling},
  author={Yang, Zhicheng and Wang, Yiwei and Huang, Yinya and Guo, Zhijiang and Shi, Wei and Han, Xiongwei and Feng, Liang and Song, Linqi and Liang, Xiaodan and Tang, Jing},
  booktitle={International Conference on Learning Representations},
  volume={2025},
  pages={24726--24759},
  year={2025},
  sourceurl={https://proceedings.iclr.cc/paper_files/paper/2025/hash/3deb687c44d3687ace0729e5db3b4efd-Abstract-Conference.html}
}

@inproceedings{zhao2026sage,
title={Strategy-Aware Optimization Modeling with Reasoning {LLM}s},
author={Ruiqing Zhao and Fengzhi Li and Yuan Zuo and Rui Liu and YanSong Liu and Yunfei Ma and Fanyu Meng and Junlan Feng},
booktitle={Forty-third International Conference on Machine Learning},
year={2026},
sourceurl={https://icml.cc/virtual/2026/poster/60721}
}

@article{wang2026formalize,
  title={Formalize, Don't Optimize: The Heuristic Trap in LLM-Generated Combinatorial Solvers},
  author={Wang, Haoyu and Song, Yuliang and Li, Tao and Deng, Zhiwei and Wang, Yaqing and Ramachandran, Deepak and Cohen, Eldan and Roth, Dan},
  journal={arXiv preprint arXiv:2605.12421},
  year={2026}
}

@article{kong2026frontieror,
  title={FrontierOR: Benchmarking LLMs' Capacity for Efficient Algorithm Design in Large-Scale Optimization},
  author={Kong, Minwei and Jiang, Chonghe and Qu, Ao and Ouyang, Wenbin and Zeng, Zhaoming and Guo, Xiaotong and Li, Zhekai and Li, Junyi and Fan, Yi and Zheng, Xinshou and others},
  journal={arXiv preprint arXiv:2605.25246},
  year={2026}
}

@inproceedings{astorga2024autoformulation,
  title={Autoformulation of Mathematical Optimization Models Using {LLMs}},
  author={Astorga, Nicol{\'a}s and Liu, Tennison and Xiao, Yuanzhang and Van Der Schaar, Mihaela},
  booktitle={Proceedings of the 42nd International Conference on Machine Learning},
  volume={267},
  series={Proceedings of Machine Learning Research},
  pages={1864--1886},
  publisher={PMLR},
  year={2025},
  sourceurl={https://proceedings.mlr.press/v267/astorga25a.html}
}

@inproceedings{zhou2026steporlm,
  title={{StepORLM}: A Self-Evolving Framework With Generative Process Supervision For Operations Research Language Models},
  author={Zhou, Chenyu and Xu, Tianyi and Lin, Jianghao and Ge, Dongdong},
  booktitle={International Conference on Learning Representations},
  volume={2026},
  pages={6914--6940},
  year={2026},
  sourceurl={https://proceedings.iclr.cc/paper_files/paper/2026/hash/0bcfb525c8f8f07ae10a93d0b2a40e00-Abstract-Conference.html}
}

@inproceedings{chen2026sirl,
  title={Solver-Informed {RL}: Grounding Large Language Models for Authentic Optimization Modeling},
  author={Chen, Yitian and Xia, Jingfan and Shao, Siyu and Ge, Dongdong and Ye, Yinyu},
  booktitle={Advances in Neural Information Processing Systems},
  volume={38},
  pages={106027--106069},
  year={2025},
  sourceurl={https://proceedings.neurips.cc/paper_files/paper/2025/hash/98555b92e5dedbdad5921049281995bc-Abstract-Conference.html}
}

@inproceedings{liu2026optitree,
  title={{OptiTree}: Hierarchical Thoughts Generation with Tree Search for {LLM} Optimization Modeling},
  author={Liu, Haoyang and Wang, Jie and Cai, Yuyang and Han, Xiongwei and Kuang, Yufei and Hao, Jianye},
  booktitle={Advances in Neural Information Processing Systems},
  volume={38},
  pages={120713--120781},
  year={2025},
  sourceurl={https://proceedings.neurips.cc/paper_files/paper/2025/hash/aeaf9b454f4670d5bf2f02b4b6e5bffd-Abstract-Conference.html}
}

@article{zhang2025optimind,
  title={OptiMind: Teaching LLMs to think like optimization experts},
  author={Zhang, Xinzhi and Chen, Zeyi and Zope, Humishka and Barbalho, Hugo and Mellou, Konstantina and Molinaro, Marco and Kulkarni, Janardhan and Menache, Ishai and Li, Sirui},
  journal={arXiv preprint arXiv:2509.22979},
  year={2025}
}

@inproceedings{liu2026optverifier,
  title={Opt-Verifier: Unleashing the Power of LLMs for Optimization Modeling via Dual-Side Verification},
  author={Liu, Haoyang and Wang, Jie and Niu, Boxuan and Han, Xiongwei and Xu, Yian and Ye, Mingxuan and Geng, Zijie and Zhu, Fangzhou and Zhong, Tao and Yuan, Mingxuan and others},
  booktitle={Proceedings of the 43rd International Conference on Machine Learning},
  year={2026},
  sourceurl={https://icml.cc/Downloads/2026}
}

@inproceedings{ding2026orr1,
  title={{OR-R1}: Automating modeling and solving of operations research optimization problem via test-time reinforcement learning},
  author={Ding, Zezhen and Tan, Zhen and Zhang, Jiheng and Chen, Tianlong},
  booktitle={Proceedings of the AAAI Conference on Artificial Intelligence},
  volume={40},
  pages={228--236},
  year={2026},
  doi={10.1609/aaai.v40i1.36983},
  sourceurl={https://ojs.aaai.org/index.php/AAAI/article/view/36983}
}

@inproceedings{
liu2026optminer,
title={Opt-Miner: Empowering Information-Seeking Agent with Tree-Guided Data Synthesis for Optimization Modeling},
author={Haoyang Liu and Yuyang Cai and Jie Wang and Xiongwei Han and Minyang Hu and Shuqi LIU and Mingxuan Yuan and Jianye HAO and Feng Wu},
booktitle={Forty-third International Conference on Machine Learning},
year={2026}
}

@inproceedings{kong2026alphaopt,
  title={Alphaopt: Formulating optimization programs with self-improving llm experience library},
  author={Kong, Minwei and Qu, Ao and Guo, Xiaotong and Ouyang, Wenbin and Jiang, Chonghe and Zheng, Han and Ma, Yining and Zhuang, Dingyi and Tang, Yuhan and Li, Junyi and others},
  booktitle={Proceedings of the 32nd ACM SIGKDD Conference on Knowledge Discovery and Data Mining V. 2},
  pages={2389--2400},
  year={2026}
}

@inproceedings{li2026constructing,
  title={Constructing industrial-scale optimization modeling benchmark},
  author={Li, Zhong and Lu, Hongliang and Wei, Tao and Chen, Yuxuan and Liu, Wenyu and Lan, Yuan and Zhang, Fan and Wen, Zaiwen},
  booktitle={Proceedings of the 43rd International Conference on Machine Learning},
  year={2026},
  sourceurl={https://icml.cc/virtual/2026/poster/66779}
}

@article{yang2026orthought,
  title={ORThought: Benchmarking and automating logistics optimization modeling via structured LLM reasoning},
  author={Yang, Beinuo and Zhou, Qishen and Li, Junyi and Su, Chenxing and Angeloudis, Panagiotis and Hu, Simon},
  journal={Artificial Intelligence for Transportation},
  volume={6},
  pages={100059},
  year={2026},
  publisher={Elsevier}
}

@article{refai2026component,
  title={A component-level evaluation framework for diagnosing LLM errors in optimization modelling},
  author={Refai, Dania and Ahmed, Moataz},
  journal={Neurocomputing},
  pages={133981},
  year={2026},
  publisher={Elsevier}
}

@inproceedings{huang2025llms,
  title={LLMs for mathematical modeling: Towards bridging the gap between natural and mathematical languages},
  author={Huang, Xuhan and Shen, Qingning and Hu, Yan and Gao, Anningzhe and Wang, Benyou},
  booktitle={Findings of the Association for Computational Linguistics: NAACL 2025},
  pages={2678--2710},
  year={2025}
}

@misc{wang2024optibench,
  title={OptiBench: Benchmarking large language models in optimization modeling with equivalence-detection evaluation},
  author={Wang, Zhuohan and Zhu, Ziwei and Han, Yizhou and Lin, Yufeng and Lin, Zhihang and Sun, Ruoyu and Ding, Tian},
  year={2024},
  howpublished={OpenReview manuscript},
  sourceurl={https://openreview.net/forum?id=KD9F5Ap878}
}

@article{wang2025orgeval,
  title={ORGEval: Graph-Theoretic Evaluation of LLMs in Optimization Modeling},
  author={Wang, Zhuohan and Zhu, Ziwei and Li, Ziniu and Chen, Congliang and Han, Yizhou and Lin, Yufeng and Lin, Zhihang and Gu, Angyang and Hu, Xinglin and Sun, Ruoyu and others},
  journal={arXiv preprint arXiv:2510.27610},
  year={2025}
}

@misc{gurobi_variability,
  author = {{Gurobi Optimization}},
  title = {What is performance variability?},
  year = {2026},
  sourceurl = {https://support.gurobi.com/hc/en-us/articles/26989876720657-What-is-performance-variability},
  note = {Official support documentation; accessed 2026-09-17}
}

@misc{gurobi_model_attributes,
  author = {{Gurobi Optimization}},
  title = {{Gurobi Optimizer Reference Manual}: Model Attributes},
  year = {2026},
  sourceurl = {https://docs.gurobi.com/projects/optimizer/en/current/reference/attributes/model.html},
  note = {Runtime, Work, MemUsed, and MaxMemUsed; accessed 2026-09-17}
}

@misc{gurobi_start,
  author = {{Gurobi Optimization}},
  title = {{Gurobi Optimizer Reference Manual}: Variable Attributes},
  year = {2026},
  sourceurl = {https://docs.gurobi.com/projects/optimizer/en/current/reference/attributes/variable.html},
  note = {Starts, hints, priorities, and basis attributes; accessed 2026-09-17}
}

@misc{gurobi_numerics,
  author = {{Gurobi Optimization}},
  title = {{Gurobi} Numerical Guidelines: Tolerances and User-Scaling},
  year = {2026},
  sourceurl = {https://docs.gurobi.com/projects/optimizer/en/current/concepts/numericguide/tolerances_scaling.html},
  note = {Accessed 2026-09-17}
}

@incollection{lodi2013variability,
  author = {Lodi, Andrea and Tramontani, Andrea},
  title = {Performance Variability in Mixed-Integer Programming},
  booktitle = {Theory Driven by Influential Applications},
  series = {INFORMS TutORials in Operations Research},
  pages = {1--12},
  year = {2013},
  publisher = {INFORMS},
  doi = {10.1287/educ.2013.0112},
  sourceurl = {https://doi.org/10.1287/educ.2013.0112}
}

@inproceedings{snell2025scaling,
  title = {Scaling LLM Test-Time Compute Optimally Can be More Effective than Scaling Parameters for Reasoning},
  author = {Charlie Snell and Jaehoon Lee and Kelvin Xu and Aviral Kumar},
  year = {2025},
  sourceurl = {https://proceedings.iclr.cc/paper_files/paper/2025/hash/1b623663fd9b874366f3ce019fdfdd44-Abstract-Conference.html},
  booktitle = {The Thirteenth International Conference on Learning Representations}
}

@inproceedings{wu2025inference,
  title = {Inference Scaling Laws: An Empirical Analysis of Compute-Optimal Inference for LLM Problem-Solving},
  author = {Yangzhen Wu and Zhiqing Sun and Shanda Li and Sean Welleck and Yiming Yang},
  year = {2025},
  sourceurl = {https://proceedings.iclr.cc/paper_files/paper/2025/hash/8c3caae2f725c8e2a55ecd600563d172-Abstract-Conference.html},
  booktitle = {The Thirteenth International Conference on Learning Representations}
}

@inproceedings{chen2025overthinking,
  title = {Do NOT Think That Much for 2+3=? On the Overthinking of Long Reasoning Models},
  author = {Xingyu Chen and Jiahao Xu and Tian Liang and Zhiwei He and Jianhui Pang and Dian Yu and Linfeng Song and Qiuzhi Liu and Mengfei Zhou and Zhuosheng Zhang and Rui Wang and Zhaopeng Tu and Haitao Mi and Dong Yu},
  year = {2025},
  sourceurl = {https://proceedings.mlr.press/v267/chen25bx.html},
  booktitle = {Proceedings of the 42nd International Conference on Machine Learning},
  volume = {267},
  series = {Proceedings of Machine Learning Research},
  publisher = {PMLR},
  pages = {9487--9499}
}

@inproceedings{yi2025shorterbetter,
  title = {ShorterBetter: Guiding Reasoning Models to Find Optimal Inference Length for Efficient Reasoning},
  author = {Jingyang Yi and Jiazheng Wang and Sida Li},
  year = {2025},
  sourceurl = {https://proceedings.neurips.cc/paper_files/paper/2025/hash/377ca194437916417e3dbe0dc67254f1-Abstract-Conference.html},
  booktitle = {Advances in Neural Information Processing Systems},
  volume = {38},
  doi = {10.52202/085713-1302}
}

@article{vielma2015formulation,
  author={Vielma, Juan Pablo},
  title={Mixed Integer Linear Programming Formulation Techniques},
  journal={SIAM Review},
  volume={57},
  number={1},
  pages={3--57},
  year={2015},
  doi={10.1137/130915303},
  sourceurl={https://doi.org/10.1137/130915303}
}

@article{vanderhulst2026implied,
  author={van der Hulst, Rolf and Walter, Matthias},
  title={Implied Integrality in Mixed-Integer Optimization},
  journal={Mathematical Programming},
  year={2026},
  doi={10.1007/s10107-026-02389-3},
  sourceurl={https://doi.org/10.1007/s10107-026-02389-3}
}

@incollection{margot2010symmetry,
  author={Margot, Fran{\c{c}}ois},
  title={Symmetry in Integer Linear Programming},
  booktitle={50 Years of Integer Programming 1958--2008: From the Early Years to the State-of-the-Art},
  editor={J{\"u}nger, Michael and Liebling, Thomas M. and Naddef, Denis and Nemhauser, George L. and Pulleyblank, William R. and Reinelt, Gerhard and Rinaldi, Giovanni and Wolsey, Laurence A.},
  publisher={Springer},
  address={Berlin, Heidelberg},
  pages={647--686},
  year={2010},
  doi={10.1007/978-3-540-68279-0_17},
  sourceurl={https://doi.org/10.1007/978-3-540-68279-0_17}
}

@article{dantzig1960decomposition,
  author={Dantzig, George B. and Wolfe, Philip},
  title={Decomposition Principle for Linear Programs},
  journal={Operations Research},
  volume={8},
  number={1},
  pages={101--111},
  year={1960},
  doi={10.1287/opre.8.1.101},
  sourceurl={https://doi.org/10.1287/opre.8.1.101}
}

@article{klotz2013linear,
  author={Klotz, Ed and Newman, Alexandra M.},
  title={Practical Guidelines for Solving Difficult Linear Programs},
  journal={Surveys in Operations Research and Management Science},
  volume={18},
  number={1--2},
  pages={1--17},
  year={2013},
  doi={10.1016/j.sorms.2012.11.001},
  sourceurl={https://doi.org/10.1016/j.sorms.2012.11.001}
}

@article{yang2026optskills,
  title={{OptSkills}: Learning Generalizable Optimization Skills from Problem Archetypes via Cluster-Based Distillation},
  author={Yang, Haochen and Zhao, Ke and Ma, Mengyuan and Lu, Xingyu and Wang, Xiangfeng and Qian, Hong},
  journal={arXiv preprint arXiv:2605.29829},
  year={2026}
}

@article{li2026optargus,
  title={{OptArgus}: A Multi-Agent System to Detect Hallucinations in {LLM}-based Optimization Modeling},
  author={Li, Zhong and Guo, Zihan and Lu, Xiaohan and Wang, Juntao and Song, Jie and Shen, Chao and Wu, Jiageng and Sun, Mingyang},
  journal={arXiv preprint arXiv:2605.11738},
  year={2026},
  doi={10.48550/arXiv.2605.11738},
  sourceurl={https://arxiv.org/abs/2605.11738}
}

@book{bertsimas1997linear,
  author = {Bertsimas, Dimitris and Tsitsiklis, John N.},
  title = {Introduction to Linear Optimization},
  publisher = {Athena Scientific},
  year = {1997},
  isbn = {9781886529199},
  sourceurl = {https://ocw.mit.edu/courses/6-251j-introduction-to-mathematical-programming-fall-2009/pages/readings/}
}

@book{nemhauser1988integer,
  author = {Nemhauser, George L. and Wolsey, Laurence A.},
  title = {Integer and Combinatorial Optimization},
  publisher = {John Wiley \& Sons},
  year = {1988},
  isbn = {9780471828198},
  doi = {10.1002/9781118627372},
  sourceurl = {https://onlinelibrary.wiley.com/doi/book/10.1002/9781118627372}
}

@book{boyd2004convex,
  author = {Boyd, Stephen and Vandenberghe, Lieven},
  title = {Convex Optimization},
  publisher = {Cambridge University Press},
  year = {2004},
  isbn = {9780521833783},
  doi = {10.1017/CBO9780511804441},
  sourceurl = {https://web.stanford.edu/~boyd/cvxbook/}
}

@book{birge2011stochastic,
  author = {Birge, John R. and Louveaux, Fran{\c{c}}ois},
  title = {Introduction to Stochastic Programming},
  edition = {2nd},
  publisher = {Springer},
  year = {2011},
  isbn = {9781461402367},
  doi = {10.1007/978-1-4614-0237-4},
  sourceurl = {https://link.springer.com/book/10.1007/978-1-4614-0237-4}
}

@article{benders1962partitioning,
  author = {Benders, J. F.},
  title = {Partitioning Procedures for Solving Mixed-Variables Programming Problems},
  journal = {Numerische Mathematik},
  volume = {4},
  pages = {238--252},
  year = {1962},
  doi = {10.1007/BF01386316},
  sourceurl = {https://link.springer.com/article/10.1007/BF01386316}
}

@article{mccormick1976computability,
  author = {McCormick, Garth P.},
  title = {Computability of Global Solutions to Factorable Nonconvex Programs: Part {I}---Convex Underestimating Problems},
  journal = {Mathematical Programming},
  volume = {10},
  pages = {147--175},
  year = {1976},
  doi = {10.1007/BF01580665},
  sourceurl = {https://link.springer.com/article/10.1007/BF01580665}
}

@misc{gurobi_constraints,
  author = {{Gurobi Optimization}},
  title = {{Gurobi Optimizer Reference Manual}: Constraints},
  year = {2026},
  sourceurl = {https://docs.gurobi.com/projects/optimizer/en/current/concepts/modeling/constraints.html},
  note = {SOS, indicator, piecewise-linear, and general constraints; accessed 2026-09-21}
}

@misc{cplex_indicator_practices,
  author = {{IBM}},
  title = {Best Practices with Indicator Constraints},
  year = {n.d.},
  sourceurl = {https://www.ibm.com/docs/en/cofz/12.9.0?topic=optimization-best-practices-indicator-constraints},
  note = {CPLEX 12.9.0 documentation; accessed 2026-09-21}
}

@manual{mosek_modeling_cookbook,
  author = {{MOSEK ApS}},
  title = {{MOSEK} Modeling Cookbook},
  year = {2025},
  sourceurl = {https://docs.mosek.com/MOSEKModelingCookbook-a4paper.pdf},
  note = {Release 3.4.0, November 5, 2025; accessed 2026-09-21}
}

@article{li2026mmoptbench,
  title={{MM-OptBench}: A Solver-Grounded Benchmark for Multimodal Optimization Modeling},
  author={Li, Zhong and Huang, Qi and Zhu, Yuxuan and Amiri, Mohammad Mohammadi and {van Stein}, Niki and B{\"a}ck, Thomas and {van Leeuwen}, Matthijs and Wen, Zaiwen and Yang, Lincen},
  journal={arXiv preprint arXiv:2605.12154},
  year={2026},
  doi={10.48550/arXiv.2605.12154},
  sourceurl={https://arxiv.org/abs/2605.12154}
}

@misc{ortools_job_shop,
  author = {{Google}},
  title = {The Job Shop Problem},
  year = {2024},
  howpublished = {OR-Tools documentation},
  sourceurl = {https://developers.google.com/optimization/scheduling/job_shop},
  note = {CP-SAT interval variables and NoOverlap constraints; accessed 2026-09-21}
}

@article{boyd2011admm,
  author = {Boyd, Stephen and Parikh, Neal and Chu, Eric and Peleato, Borja and Eckstein, Jonathan},
  title = {Distributed Optimization and Statistical Learning via the Alternating Direction Method of Multipliers},
  journal = {Foundations and Trends in Machine Learning},
  volume = {3},
  number = {1},
  pages = {1--122},
  year = {2011},
  sourceurl = {https://web.stanford.edu/~boyd/papers/admm_distr_stats.html}
}
\bibliographystyle{iclr2027_conference}

\appendix
\setcounter{topnumber}{3}
\setcounter{bottomnumber}{2}
\setcounter{totalnumber}{5}
\renewcommand{\topfraction}{0.95}
\renewcommand{\bottomfraction}{0.9}
\renewcommand{\textfraction}{0.05}
\renewcommand{\floatpagefraction}{0.8}
\setlength{\intextsep}{6pt plus 1pt minus 1pt}
\setlength{\textfloatsep}{10pt plus 2pt minus 2pt}
\setlength{\floatsep}{10pt plus 2pt minus 2pt}
\makeatletter
\setlength{\@fptop}{0pt}
\setlength{\@fpsep}{14pt}
\setlength{\@fpbot}{0pt plus 1fil}
\makeatother

\addtocontents{toc}{\protect\setcounter{tocdepth}{2}}
\begingroup
\makeatletter
\setlength{\parskip}{0pt}
\pdfbookmark[0]{Appendix}{appendix.contents}
{\fontsize{14}{16}\selectfont\bfseries Appendix\par}
\vspace{10pt}
{\fontsize{11}{13}\selectfont\bfseries Table of Contents\par}
\vspace{7pt}
\fontsize{9}{11}\selectfont
\renewcommand{\l@section}[2]{\ifnum\c@tocdepth>0\vspace{1.5pt}\fi\@dottedtocline{1}{0em}{1.8em}{#1}{#2}}
\renewcommand{\l@subsection}[2]{\@dottedtocline{2}{1.8em}{2.5em}{#1}{#2}}
\renewcommand{\l@subsubsection}[2]{} 
\setcounter{tocdepth}{-10}
\@starttoc{toc}
\setcounter{tocdepth}{3} 
\makeatother
\endgroup

\section{Extended Related Work}
\label{app:extended-related-work}

We situate EfficientOpt within four connected research directions: generating optimization models, using reusable modeling knowledge, designing benchmarks and verification methods, and evaluating computational efficiency. These directions overlap: trained models can also use search, retrieval, and solver feedback during inference.

\subsection{Generating and Refining Optimization Models}

Inference workflows organize the steps between a problem description and executable solver code. Chain-of-Experts \citep{xiao2024chain} coordinates specialized agents, while OptiMUS \citep{ahmaditeshnizi2024optimus} uses a modular workflow to formulate models, generate and debug code, and evaluate solutions. AutoFormulation \citep{astorga2024autoformulation} and OptiTree \citep{liu2026optitree} use tree search to explore candidate modeling decisions. These methods address how to construct and revise a candidate when a single generation is insufficient.

Learning-based methods improve the modeler's parameters through specialized data and feedback. ORLM \citep{huang2025orlm} trains on synthetic optimization-modeling instructions; LLMOPT \citep{jiang2025llmopt} uses multi-instruction tuning for mathematical formalization and code generation. OptMATH \citep{lu2025optmath} and ReSocratic \citep{yang2025optibench} synthesize mathematical models and corresponding natural-language problems to support training. Solver feedback provides another source of supervision: SIRL \citep{chen2026sirl} derives rewards from executable code and instance-level model verification; StepORLM \citep{zhou2026steporlm} combines solver outcomes with generative process supervision. OR-R1 \citep{ding2026orr1} combines supervised fine-tuning with test-time policy optimization, while PEARL \citep{lu2026pearl} learns interactive execution and revision using Python and solver feedback. Together, these approaches motivate evaluating both the mathematical model and its implemented solution procedure.

\subsection{Expert Knowledge and Reusable Modeling Skills}

Knowledge-augmented approaches differ in what they store and how it guides modeling. Opt-Miner \citep{liu2026optminer} studies information seeking for optimization modeling. AlphaOPT \citep{kong2026alphaopt} maintains a library of solver-verified insights with applicability conditions, explanations, and examples. OptiMind \citep{zhang2025optimind} uses expert analyses of errors within problem classes to develop guidance for training and inference. In concurrent work, OptSkills \citep{yang2026optskills} distills modeling and solving trajectories into workflows and pitfalls organized by problem archetype, then retrieves these skills during inference. Its examples include activation variables, disjunctive constraints, and Big-$M$ formulations, demonstrating that reusable workflows already contain concrete modeling knowledge.

Our OptTips organizes knowledge at the level of individual modeling techniques for task design and formulation inspection. A card connects applicability conditions and inefficiency symptoms to modeling actions and a mathematical example. Our OptDachshund uses these specifications to redesign tasks and construct ordinary and expert reference implementations on the same numerical instance. This use of technique-level knowledge for benchmark construction complements its use as assistance during generation; the evaluated LLMs receive neither the target-technique cards nor the reference implementations.

\subsection{Benchmark Coverage and Semantic Verification}

Benchmarks differ in both the problems they cover and the evidence used to assess a generated model. NL4Opt \citep{ramamonjison2023nl4opt} studies the translation of natural-language descriptions into linear-programming formulations. ComplexOR \citep{xiao2024chain} and NLP4LP \citep{ahmaditeshnizi2024optimus} extend evaluation to more complex modeling requirements and descriptions. IndustryOR \citep{huang2025orlm} and MAMO \citep{huang2025llms} broaden application and mathematical-modeling coverage. OptiBench \citep{yang2025optibench} includes linear and nonlinear problems with or without tabular data, while OptMATH \citep{lu2025optmath} contributes challenging instances through its synthesis and filtering process.

Recent benchmarks extend scale and input modality. MIPLIB-NL \citep{li2026constructing} reconstructs 223 industrial MILP instances from MIPLIB~2017 as natural-language specifications with separate numerical data, using independent reconstruction and expert review to check fidelity. MM-OptBench \citep{li2026mmoptbench} combines text with structured visuals in 780 solver-verified instances across six problem families. Its diagnostics separate input-extraction errors from formulation and implementation failures. These resources broaden the settings in which modeling capabilities can be tested.

Verification also extends beyond matching an objective value. The equivalence-oriented OptiBench \citep{wang2024optibench} and the graph-based ORGEval framework and Bench4Opt dataset \citep{wang2025orgeval} examine structural relationships between formulations. Component-level evaluation \citep{refai2026component} diagnoses variables, objectives, and constraints separately. OptArgus \citep{li2026optargus} audits consistency across descriptions, symbolic formulations, and code, measuring false alarms, error localization, and detection. Opt-Verifier \citep{liu2026optverifier} combines structure-side and solution-side verification. These distinctions inform our separation of numerical agreement from mathematical inspection: a matching objective alone does not establish that a modeling transformation preserves the task's required outputs.

\subsection{Evaluating Computational Efficiency}

Efficiency can concern generating an answer, checking it, or executing the resulting solver program. Test-time scaling and overthinking studies examine how reasoning effort translates into answer quality \citep{snell2025scaling,wu2025inference,chen2025overthinking,yi2025shorterbetter}. In optimization modeling, ORThought \citep{yang2026orthought} reports token efficiency, and component-level evaluation \citep{refai2026component} includes latency and token usage. OptArgus \citep{li2026optargus} measures the cost of running its auditor and discusses efficiency-related modeling issues in its error taxonomy. These resource scopes differ from the construction and solution of the generated optimization model.

Several works directly examine downstream computation. SAGE \citep{zhao2026sage} trains with correctness and solver-efficiency feedback from alternative modeling strategies and evaluates solver time, iterations, and formulation size. Experiments on CP-SynC-XL \citep{wang2026formalize} show that efficiency-oriented prompting can yield small or inconsistent speedups and can compromise correctness through unreliable search modifications. In concurrent work, FrontierOR \citep{kong2026frontieror} evaluates solution quality and runtime for LLM-designed optimization algorithms against a representative formulation and Gurobi implementation for each task. Its scope includes reformulation and decomposition, and its references are also hidden from evaluated models.

Our EfficientOpt focuses this evaluation on designated modeling techniques. Each task is constructed around an applicable technique and paired with ordinary and expert references for the same task and data. The pair makes the conventional and technique-based implementations explicit comparison points. We inspect technique adoption separately from numerical correctness and recorded generation, model-construction, solving, and memory costs. This design supports distinguishing non-adoption of the designated technique, costly implementations that adopt it, and effective alternatives. The contribution is this combination of technique-guided task construction, paired references, and separate assessments, building on existing work on modeling knowledge and computational efficiency.

\section{Benchmark Composition and Evaluation Coverage}
\label{app:dataset_stats}

EfficientOpt contains 561 tasks covering 29 of the 50 OptTips techniques. The remaining 21 techniques have 63 supplementary tasks, three per technique, documented in Appendix~\ref{app:opttips-catalog} and excluded from the reported evaluation. This appendix describes task classification, reference verification, measurement coverage, and code generation.

\subsection{Task Composition and Classification}
\label{app:composition-labels}

\leavevmode\vspace{-2\baselineskip}
\begin{wraptable}{r}{0.6\linewidth}
\vspace{-8pt}
\centering\small
\setlength{\tabcolsep}{3pt}
\caption{\footnotesize Task counts by problem category (left) and application field (right), each totaling 561. Figure~\ref{fig:benchmark_profile}(a) combines fitting and estimation with layout planning as \emph{Other} (25 tasks).}
\label{tab:dataset-composition}
\resizebox{\linewidth}{!}{%
\begin{tabular}{@{}p{0.40\textwidth}r@{\hspace{1.3em}}p{0.39\textwidth}r@{}}
\toprule
\textbf{Primary problem category} & \textbf{$n$} &
\textbf{Application field} & \textbf{$n$} \\
\midrule
Resource allocation \& production & 177 & Mobility \& logistics & 114 \\
Routing \& network flow & 121 & Manufacturing \& industry & 106 \\
Packing \& selection & 88 & Unspecified & 76 \\
Covering \& facility planning & 68 & Public services \& research & 61 \\
Assignment \& matching & 45 & Business \& resource planning & 47 \\
Scheduling \& control & 37 & Health \& life sciences & 46 \\
Fitting \& estimation & 21 & Computing \& data analysis & 43 \\
Layout planning & 4 & Energy \& utilities & 37 \\
 & & Environment \& agriculture & 31 \\
\midrule
\textbf{Total} & \textbf{561} & \textbf{Total} & \textbf{561} \\
\bottomrule
\end{tabular}}
\vspace{-6pt}
\end{wraptable}

\noindent\textbf{Problem categories and application fields.} Each task receives one primary problem category and one application label based on its statement and numerical instance (Table~\ref{tab:dataset-composition}). Tasks without a clear application context are labeled \emph{Unspecified}. These categories describe the decision problem and its context; the target modeling technique is recorded separately.

\noindent\textbf{Formulation types.} We classify each task's ordinary reference by its objective, constraints, and variable domains before presolve. The 561 references fall into seven mutually exclusive classes: MILP (225), binary IP (158), nonbinary pure IP (76), LP (65), MIQCP (27), MIQP (8), and QP (2), as shown in Figure~\ref{fig:benchmark_profile}(b).

The classification follows the implemented ordinary formulation. Integer variables restricted to $\{0,1\}$ count as binary. The MILP class includes 13 ordinary reference models with linearly \mbox{representable} general constraints. Explicit products of decision variables retain a quadratic classification, even when a linear reformulation is possible. Stochastic and robust models can belong to these same algebraic classes.

These classifications complement the 29 target-technique groups, which contain 5--46 tasks each. Appendix~\ref{app:group-outcomes} provides the group descriptions and corresponding accuracy results.

\finishwraptable

\FloatBarrier
\subsection{Evaluation Coverage and Reference Verification}
\label{app:evaluation-populations}

We evaluate 11 general-purpose LLMs on all 561 tasks, retaining one outcome per model--task pair, including generation and execution failures. Each model's numerical accuracy uses all 561 tasks, yielding 6,171 retained outcomes in total. Cost comparisons require a correct numerical answer and the measurements needed for the relevant metric, as detailed in Appendix~\ref{app:measurement-detail}.

\noindent\textbf{Reference answers and cost coverage.} All 561 tasks have verified reference objectives. For 543 tasks, the selected ordinary and expert executions reach optimality and agree on the objective under the cost-measurement protocol. The remaining 18 tasks are verified through supplementary reference executions, analytic or exact solutions, or optimality certificates on the same numerical inputs. These checks supply reference answers for accuracy; their timings are not added to the cost analysis, which retains the 543-task subset. Appendix~\ref{app:quality-control} describes the validation criteria.

\noindent\textbf{Reference model sizes.} Figure~\ref{fig:benchmark_profile}(d,e) covers all 561 reference pairs. For each implementation, variable and total constraint counts are separate maxima over models submitted to the solver before presolve; the two maxima need not come from the same model or solver call. Repeated solves are not summed. Total constraints include native linear, quadratic, general, and SOS constraints, counting each once; variable bounds, integrality declarations, and solver-generated cuts are excluded. Counts are measured during execution or derived exactly from the implementation and fixed input where needed. Appendix~\ref{app:structure-diagnostic} instead relates runtime to the variable, linear-constraint, and nonzero counts recorded in the corresponding execution.

\subsection{Code Generation Protocol}
\label{app:generation-provenance}

Each prompt provides the problem statement and input-field descriptions. It requests a function that constructs and returns a Gurobi model; the evaluation runner supplies the complete numerical instance at execution time and invokes the solver. Some prompts also request a symbolic formulation summary. Code-only prompts include a general efficiency instruction without specifying the target technique.

\FloatBarrier

\section{Detailed Experimental Results}
\label{app:full-results-561}

This appendix expands the main findings with accuracy breakdowns, cost distributions, and analyses of model size, preparation, generation, and memory use. It also reports fine-tuned-model failures (Section~\ref{app:specialized-open-models}) and identifies the slow tasks driving cumulative solver costs for all 11 general-purpose models (Section~\ref{app:slow-task-analysis}). Measurement definitions and comparison rules are in Appendix~\ref{app:measurement-detail}; technique-use assessments are in Appendix~\ref{app:judge-full}.

\subsection{Numerical Outcomes (RQ1)}
\label{app:numerical-outcomes}

All 11 general-purpose models are evaluated on the same 561 tasks using the verified reference objectives and the correctness criterion in Section~\ref{sec:end-to-end}. Table~\ref{tab:outcome-decomposition} separates correct answers, incorrect objectives returned with \texttt{OPTIMAL} status, and other generation or execution outcomes.

\begin{table}[!htbp]
\centering\footnotesize
\setlength{\tabcolsep}{3.5pt}
\caption{\footnotesize Numerical outcomes on all 561 tasks per model. Correct, Objective mismatch, and Other outcome sum to 561. The \texttt{OPTIMAL} column is the sum of Correct and Objective mismatch.}
\label{tab:outcome-decomposition}
\begin{tabular}{lrrrr}\toprule
 & \multicolumn{3}{c}{Mutually exclusive outcomes} & Solver status \\
\cmidrule(lr){2-4}\cmidrule(lr){5-5}
Model & Correct & Objective mismatch & Other outcome & OPTIMAL \\\midrule
Gemini 3.1 Pro & 512 & 10 & 39 & 522 \\
GPT-5.5 & 447 & 23 & 91 & 470 \\
Claude Opus 4.6 & 442 & 36 & 83 & 478 \\
DeepSeek-V4 Flash & 441 & 25 & 95 & 466 \\
Kimi K2.6 & 433 & 23 & 105 & 456 \\
Qwen 3.6 27B & 394 & 27 & 140 & 421 \\
GLM-5.1 & 390 & 40 & 131 & 430 \\
Qwen 3.5 122B & 369 & 28 & 164 & 397 \\
MiniMax M2.5 & 315 & 33 & 213 & 348 \\
Qwen 3 32B & 313 & 56 & 192 & 369 \\
Qwen 3.6 Plus & 208 & 37 & 316 & 245 \\
\midrule
Total & 4,264 & 338 & 1,569 & 4,602 \\
\bottomrule\end{tabular}
\end{table}

\noindent\textbf{Solver status can give a different ranking from accuracy.} Across 6,171 evaluations, 4,264 return correct objective values (69.1\%). Of the 4,602 programs reporting \texttt{OPTIMAL}, 338 (7.3\%) return incorrect values. For example, Claude Opus 4.6 produces more \texttt{OPTIMAL} results than GPT-5.5 (478 versus 470), but fewer correct answers (442 versus 447). Checking the objective therefore matters both for identifying errors and for comparing models.

\noindent\textbf{Most unsuccessful evaluations end without an \texttt{OPTIMAL} result.} The 1,569 records in Other outcome account for 82.3\% of the 1,907 unsuccessful evaluations. They include generation failures, execution errors, and nonoptimal solver termination. Table~\ref{tab:outcome-decomposition} distinguishes these outcomes from objective errors after an optimal solve. Section~\ref{app:group-outcomes} examines how accuracy and objective errors vary across task groups.

\noindent\textbf{Accuracy and cost comparisons use different task sets.} Of the 4,264 correct evaluations, 4,202 belong to the 543-task reference cost subset; the other 62 contribute only to accuracy (Appendix~\ref{app:evaluation-populations}). All 4,202 have paired \texttt{Runtime}, \texttt{Work}, and size measurements, while 4,148 have complete \texttt{Build+opt} measurements. Table~\ref{tab:main-results-561} uses the same 4,148 pairs for all four cost ratios; per-model counts appear in Table~\ref{tab:construction-effects}. Missing values are excluded. The model-size analysis additionally requires compatible counting scopes and positive measurements (Section~\ref{app:structure-diagnostic}).

\subsection{Accuracy Across Task Groups (RQ1)}
\label{app:group-outcomes}

Table~\ref{tab:groups-561} reports accuracy across all 11 models, grouped by target OptTips technique; Figure~\ref{fig:families-561} shows each model separately. The 29 groups contain 5--46 tasks each and include the modeling variants listed in the table.

\begin{table}[!htb]
\centering\small
\setlength{\tabcolsep}{3pt}
\definecolor{lzgroupbar}{HTML}{57A57B}
\newcommand{\lzgroupaccuracy}[1]{%
  \begin{tikzpicture}[baseline=(value.base),x=1.8cm,y=1ex]
    \path[use as bounding box] (0,-.35) rectangle (1,1.85);
    \shade[draw=lzgroupbar,line width=.3pt,
      left color=lzgroupbar!60!white,right color=lzgroupbar!12!white]
      (0,-.35) rectangle ({#1/100},1.85);
    \node[anchor=base,inner sep=0pt,outer sep=0pt] (value) at (.5,0) {#1};
  \end{tikzpicture}%
}
\caption{Task groups and numerical accuracy. $n$ is the number of tasks. Correct gives the number of correctly solved cases out of all $11n$ model--task pairs; the last column shows the corresponding percentage. Descriptions list the modeling variants represented in each group.}
\label{tab:groups-561}
\resizebox{0.7\linewidth}{!}{%

\endgroup
\caption{\textbf{Numerical accuracy by model and task group.} Each cell shows the percentage of tasks solved correctly, with all tasks in the group included in the denominator. Model order follows Table~\ref{tab:main-results-561}.}
\label{fig:families-561}
\end{figure}


\noindent\textbf{Accuracy varies substantially across task groups.} Across the 11 models, accuracy ranges from 27.3\% (15/55) on T17 (column generation) to 93.9\% (248/264) on T15 (independent block decomposition). Each task contributes one outcome per model.

\noindent\textbf{Objective errors are concentrated in a few groups.} T08 (substitution and product linearization) and T17 contain 51 of the 561 tasks (9.1\%) but account for 109 of the 338 programs that report \texttt{OPTIMAL} with incorrect objective values (32.2\%). On T17's five tasks, 41 of the 55 model--task evaluations produce an \texttt{OPTIMAL} result, but only 15 of these match the verified reference objectives. Appendix~\ref{app:judge-full} reports automated technique-use assessments; Appendix~\ref{app:technique-cases} summarizes selected manual checks.

\subsection{Solver Costs Across Tasks and Models (RQ2)}
\label{app:cost-distributions}

We first describe solver costs on the tasks each model solves correctly, then compare models on shared tasks. All analyses use the 543-task reference cost subset defined in Section~\ref{app:numerical-outcomes}; Appendix~\ref{app:paired-aggregation} gives the comparison and aggregation rules. Table~\ref{tab:absolute-costs-561} summarizes measured times and solver work; Table~\ref{tab:main-results-561} reports costs relative to the two references.


\begin{table}[!htb]
\centering\footnotesize
\setlength{\tabcolsep}{3.1pt}
\caption{\footnotesize Recorded costs on correctly solved tasks within the reference cost subset. Each metric uses that model's available measurements; medians are calculated separately. The final column uses the same 47 tasks for all 11 models and the expert reference. Times are in seconds; \texttt{Work} is in solver work units.}
\label{tab:absolute-costs-561}
\begin{tabular}{lrrrrrrr}
\toprule
 & \multicolumn{3}{c}{\texttt{Runtime} (s)} & \texttt{Work} & \texttt{Build} & \texttt{Build+opt} & Shared \texttt{Runtime} \\
\cmidrule(lr){2-4}
Model & Median & Mean & P90 & Median & Median & Median & Mean ($n=47$) \\
\midrule
Gemini 3.1 Pro & 32.51 & 197.66 & 414.53 & 22.66 & 0.63 & 37.80 & 77.04 \\
GPT-5.5 & 28.84 & 208.50 & 671.57 & 21.57 & 0.97 & 37.77 & 197.44 \\
Claude Opus 4.6 & 23.52 & 177.59 & 369.99 & 18.84 & 0.87 & 37.38 & 130.45 \\
DeepSeek-V4 Flash & 23.11 & 190.92 & 430.92 & 18.19 & 0.82 & 32.71 & 109.45 \\
Kimi K2.6 & 29.70 & 210.28 & 599.96 & 20.36 & 0.76 & 36.10 & 166.83 \\
Qwen 3.6 27B & 32.23 & 215.00 & 453.29 & 20.77 & 1.65 & 49.24 & 144.98 \\
GLM-5.1 & 27.59 & 150.78 & 352.58 & 18.87 & 1.01 & 33.77 & 171.77 \\
Qwen 3.5 122B & 26.00 & 200.10 & 408.90 & 20.36 & 0.84 & 32.58 & 182.40 \\
MiniMax M2.5 & 30.60 & 232.36 & 474.85 & 23.98 & 0.91 & 38.86 & 181.27 \\
Qwen 3 32B & 35.95 & 247.32 & 698.13 & 22.53 & 1.25 & 53.98 & 125.19 \\
Qwen 3.6 Plus & 34.20 & 214.13 & 426.31 & 26.45 & 0.71 & 38.19 & 134.57 \\
\midrule
Expert reference & --- & --- & --- & --- & --- & --- & 43.05 \\
\bottomrule
\end{tabular}
\end{table}

\noindent\textbf{A small number of long runs raises average solver time.} Median \texttt{Runtime} is 23.1--36.0 seconds across models, whereas mean runtime is 150.8--247.3 seconds (Table~\ref{tab:absolute-costs-561}). The slowest $\sim$10\% of correctly solved runs with valid timings account for 68.4--76.2\% of total solver time (Figure~\ref{fig:model-comparison-561}(b)). Figure~\ref{fig:runtime-tail-distributions} illustrates the lowest, middle, and highest shares using Kimi K2.6, Qwen 3.6 27B, and MiniMax M2.5. Section~\ref{app:slow-task-analysis} identifies the slow tasks for every model and compares their runtimes with the expert references.

\begin{figure}[!htb]
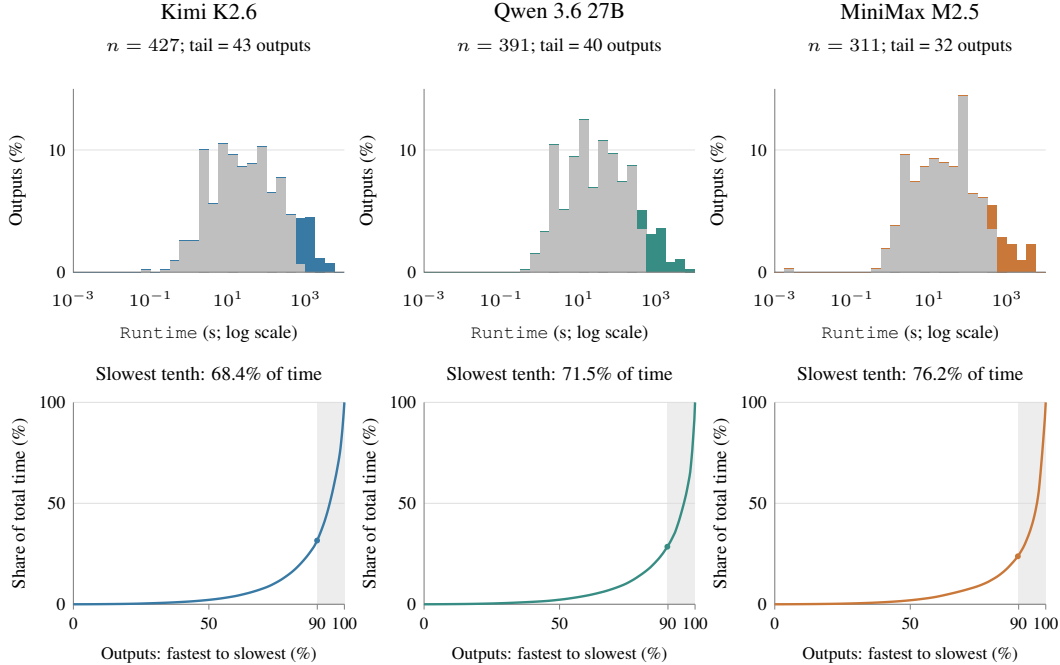

\centering
\begingroup
\definecolor{tailblue}{HTML}{397AA5}
\definecolor{tailteal}{HTML}{378D83}
\definecolor{tailorange}{HTML}{CA783A}

\endgroup
\caption{How long runs contribute to total solver time. The three models have the lowest, middle, and highest shares of time spent in their slowest tenth. Top: runtime histograms on a logarithmic time axis, with the slowest $\lceil n/10\rceil$ runs highlighted. Bottom: cumulative share of solver time as runs are ordered from fastest to slowest; shading marks the same slowest runs. Each model uses its correctly solved tasks with valid timings.}
\label{fig:runtime-tail-distributions}
\end{figure}

\noindent\textbf{Direct model comparisons use the same tasks on both sides.} Each model's runtime distribution in Table~\ref{tab:absolute-costs-561} includes a different set of correctly solved tasks. Figure~\ref{fig:model-comparison-561}(a) compares each pair on the tasks both models solve correctly, yielding 126--420 tasks per pair. A smaller set of 47 tasks is solved correctly by all 11 models. On this fixed set, mean \texttt{Runtime} is 43.05 seconds for the expert reference and 77.04--197.44 seconds for the LLMs (Table~\ref{tab:absolute-costs-561}). This comparison holds the task set fixed across every model, but covers only 8.4\% of the benchmark.

\begin{table}[!htb]
\centering\footnotesize
\setlength{\tabcolsep}{3.1pt}
\caption{\footnotesize\textbf{Solver costs by the number of models answering each task correctly.} The four groups contain 541 tasks in the reference cost subset, each solved correctly by at least one model. /Expert is the shifted geometric \texttt{Runtime} ratio (Appendix~\ref{app:paired-aggregation}). Slower (\%) is the percentage of valid LLM--expert pairs with longer LLM runtime.}
\label{tab:success-coverage-561}
\begin{tabular}{lrrrr}
\toprule
Models with correct answers & Tasks & \texttt{Runtime} pairs & /Expert $\downarrow$ & Slower (\%) \\
\midrule
1--3 & 26 & 67 & 1.69 & 58.2 \\
4--7 & 192 & 1,169 & 1.58 & 57.5 \\
8--10 & 276 & 2,449 & 1.67 & 66.8 \\
11 (all models) & 47 & 517 & 2.82 & 82.2 \\
\bottomrule
\end{tabular}
\end{table}

\noindent\textbf{The expert advantage also appears outside the 47 shared tasks.} On the other 494 tasks solved correctly by at least one LLM, 2,346 of 3,685 valid LLM--expert pairs (63.7\%) have longer LLM runtimes. Their pooled shifted geometric \texttt{Runtime} ratio is 1.64. Table~\ref{tab:success-coverage-561} groups these tasks by how many models solve them correctly: the ratio remains above one in each group. Thus, the observed efficiency gap is not confined to the small set solved correctly by every model.

\begin{figure}[!htb]
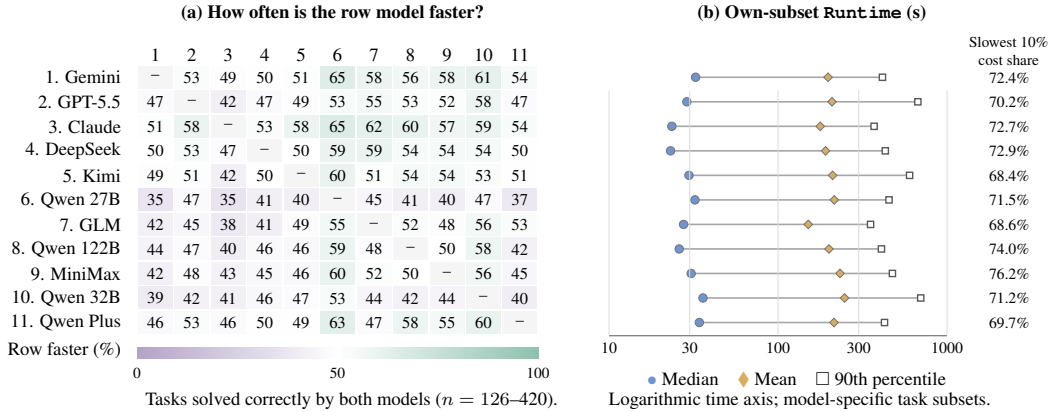

\centering
\begingroup
\definecolor{cmpmint}{HTML}{82B9A5}
\definecolor{cmplilac}{HTML}{AB99C2}
\definecolor{cmpblue}{HTML}{6F94CE}
\definecolor{cmpgold}{HTML}{D8AD60}
\resizebox{\textwidth}{!}{%
%
}
\endgroup
\caption{\textbf{Model comparisons on shared tasks and runtime distributions.} (a) Percentage of tasks solved correctly by both models with valid timings on which the row model is faster than the column model ($10^{-6}$~s tie tolerance; 126--420 tasks per pair). Opposite cells need not sum to 100\% because ties remain in the denominator. (b) Median, mean, and 90th-percentile \texttt{Runtime} on tasks solved correctly by each model, and the slowest tenth's share of measured \texttt{Runtime}. Whiskers summarize the distribution; they are not confidence intervals. Model order follows Table~\ref{tab:main-results-561}.}
\label{fig:model-comparison-561}
\end{figure}

\subsection{Model Size and Solver Time (RQ3)}
\label{app:structure-diagnostic}

We compare each generated model with the expert reference on the same task to test whether a smaller model also solves faster. Within the reference cost subset, we retain tasks whose archived ordinary and expert size summaries describe a single model. Requiring correct objective values and positive size counts and \texttt{Runtime} yields 3,788 LLM--expert pairs on 497 tasks. Dimensions and runtime come from the same recorded execution.

\noindent\textbf{Two ways to identify smaller models.} The first criterion compares the pre-presolve counts of variables, linear constraints, and nonzero coefficients in the linear constraint matrix: none of these three counts may increase, and at least one must decrease. The second requires fewer variables and fewer linear constraints, without restricting nonzeros. Both comparisons are relative to the expert. Table~\ref{tab:structure-results-561} reports the number of pairs satisfying each criterion and how often the LLM takes longer to solve the task.

\begin{table}[!htb]
\centering
\begin{minipage}{0.6\linewidth}
\centering\small
\setlength{\tabcolsep}{4pt}
\caption{\footnotesize Model size and solver time relative to the expert reference. Eligible counts pairs before applying the size criteria; $n$ counts pairs meeting each criterion. Slower counts pairs where LLM \texttt{Runtime} exceeds the expert's runtime by more than $10^{-6}$ seconds; percentages use $n$ as the denominator. Rows are linear constraints; nonzeros are nonzero coefficients in their constraint matrix.}
\label{tab:structure-results-561}
\resizebox{\linewidth}{!}{%
\begin{tabular}{lrrrrrrr}
\toprule
& & \multicolumn{3}{c}{No larger on three counts} & \multicolumn{3}{c}{Fewer variables and rows} \\
\cmidrule(lr){3-5}\cmidrule(lr){6-8}
Model & Eligible & $n$ & Slower & \% & $n$ & Slower & \% \\
\midrule
Gemini 3.1 Pro & 465 & 100 & 55 & 55.0 & 45 & 24 & 53.3 \\
GPT-5.5 & 401 & 45 & 26 & 57.8 & 19 & 11 & 57.9 \\
Claude Opus 4.6 & 395 & 82 & 44 & 53.7 & 41 & 19 & 46.3 \\
DeepSeek-V4 Flash & 395 & 59 & 31 & 52.5 & 29 & 10 & 34.5 \\
Kimi K2.6 & 390 & 70 & 40 & 57.1 & 28 & 13 & 46.4 \\
Qwen 3.6 27B & 352 & 80 & 56 & 70.0 & 31 & 23 & 74.2 \\
GLM-5.1 & 342 & 71 & 36 & 50.7 & 27 & 12 & 44.4 \\
Qwen 3.5 122B & 329 & 74 & 52 & 70.3 & 28 & 21 & 75.0 \\
MiniMax M2.5 & 277 & 56 & 31 & 55.4 & 29 & 14 & 48.3 \\
Qwen 3 32B & 270 & 66 & 52 & 78.8 & 31 & 25 & 80.6 \\
Qwen 3.6 Plus & 172 & 32 & 26 & 81.2 & 9 & 9 & 100.0 \\
\midrule
Total & 3,788 & 735 & 449 & 61.1 & 317 & 181 & 57.1 \\
\bottomrule\end{tabular}}
\end{minipage}
\end{table}

\noindent\textbf{Smaller reported models often solve more slowly.} Under the first criterion, 449 of 735 pairs (61.1\%) have longer LLM runtimes. The per-model proportion ranges from 50.7\% to 81.2\%, so the pattern appears in every model (Figure~\ref{fig:diagnostic-costs}(a)). Under the second criterion, 181 of 317 pairs (57.1\%) are slower. Reducing both variables and linear constraints therefore does not by itself establish a runtime improvement.

\noindent\textbf{The differences are not limited to nearly tied runtimes.} Of the 449 slower cases under the first criterion, 346 take at least twice as long as the expert, with at least one additional second of solver time. These cases span all 11 models. Including zero-valued counts and runtimes changes the slower proportions to 60.6\% (449/741) and 56.0\% (181/323) under the first and second criteria, respectively. The finding is therefore similar with or without zero-valued observations.

\subsection{Costs Beyond Solver Time (RQ4)}
\label{app:resource-distributions}

We examine model preparation, code generation, and memory use. Preparation and execution comparisons use correctly solved tasks in the 543-task reference cost subset. Generation summaries include all available responses. Appendix~\ref{app:measurement-detail} defines the measurements.

\noindent\textbf{Preparing a model can take longer than solving it.} Among 4,202 LLM runs with valid preparation and solver-call timings, \texttt{Build} exceeds solver-call time in 729 (17.3\%). The proportion ranges from 11.5\% to 22.5\% across models (Table~\ref{tab:construction-effects}), even though the median preparation share of \texttt{Build+opt} is only 2.8\%.

\noindent\textbf{Preparation can reverse the comparison with the expert.} On the 4,148 pairs with both LLM and expert timings, we first compare the externally measured solver-call times, then compare \texttt{Build+opt}. Adding preparation reverses which implementation is faster in 357 pairs (8.6\%): 122 LLM programs change from faster to slower than the expert, and 235 change from slower to faster. These reversals occur in all 11 models. Thus, a shorter solver call does not necessarily mean a faster implementation once preparation is included.

\begin{table}[!htbp]
\centering\footnotesize
\setlength{\tabcolsep}{3.5pt}
\caption{\footnotesize Preparation time and comparisons with the expert. LLM runs have recorded preparation and solver-call times; LLM--expert pairs also have the corresponding expert measurements. \texttt{Build} longer counts runs where preparation exceeds solver-call time, with percentages over LLM runs. The last two columns count changes in the LLM's faster/slower relation to the expert when preparation is included.}
\label{tab:construction-effects}
\begin{tabular}{@{}lrrrrr@{}}
\toprule
Model & \shortstack{LLM\\runs} & \shortstack{\texttt{Build} longer\\$n$ (\%)} & \shortstack{LLM--expert\\pairs} & \shortstack{Faster to\\slower} & \shortstack{Slower to\\faster} \\
\midrule
Gemini 3.1 Pro & 504 & 58 (11.5) & 499 & 0 & 49 \\
GPT-5.5 & 441 & 73 (16.6) & 435 & 8 & 33 \\
Claude Opus 4.6 & 435 & 76 (17.5) & 430 & 21 & 29 \\
DeepSeek-V4 Flash & 433 & 88 (20.3) & 428 & 14 & 18 \\
Kimi K2.6 & 427 & 71 (16.6) & 423 & 14 & 22 \\
Qwen 3.6 27B & 391 & 88 (22.5) & 386 & 24 & 15 \\
GLM-5.1 & 382 & 60 (15.7) & 375 & 5 & 22 \\
Qwen 3.5 122B & 367 & 70 (19.1) & 361 & 14 & 14 \\
MiniMax M2.5 & 311 & 61 (19.6) & 308 & 13 & 11 \\
Qwen 3 32B & 309 & 57 (18.4) & 303 & 6 & 14 \\
Qwen 3.6 Plus & 202 & 27 (13.4) & 200 & 3 & 8 \\
\midrule
Total & 4,202 & 729 (17.3) & 4,148 & 122 & 235 \\
\bottomrule
\end{tabular}
\end{table}

\noindent\textbf{The reversals include substantial time differences.} In 160 of the 357 reversals, the final \texttt{Build+opt} difference is at least 10 seconds. Our default comparison treats differences within $10^{-6}$ seconds as ties. Increasing this tolerance to the larger of 0.01 seconds and 1\% of the longer compared time still gives 317 reversals (7.6\%).

\begin{table}[!htb]
\centering
\begin{minipage}{0.6\linewidth}
\centering\small
\setlength{\tabcolsep}{3.0pt}
\caption{\footnotesize Generation latency and token counts for all available responses, including incorrect programs. Values are medians. $n_A$, $n_T$, and $n_R$ count records with latency, input/output-token, and reasoning-token measurements, respectively. Prompt and Completion denote input and output tokens. Token definitions vary by provider; reasoning tokens may already be included in Completion, and a reported zero does not establish that no internal reasoning occurred.}
\label{tab:generation-profile-561}
\resizebox{\linewidth}{!}{%
\begin{tabular}{lrrrrrr}
\toprule
Model & $n_A/n_T$ & \texttt{API latency} (s) & \texttt{Prompt} & \texttt{Completion} & \texttt{Reasoning} & $n_R$\\
\midrule
Gemini 3.1 Pro & 559/558 & 55.4 & 743.5 & 6,714.5 & 5,827 & 558 \\
GPT-5.5 & 535/534 & 61.2 & 715.5 & 1,431 & 516 & 522 \\
Claude Opus 4.6 & 561/561 & 17.2 & 937 & 748 & 0 & 561 \\
DeepSeek-V4 Flash & 561/561 & 81.9 & 714 & 11,327 & 10,562 & 561 \\
Kimi K2.6 & 561/561 & 82.2 & 714 & 11,680 & 10,714 & 561 \\
Qwen 3.6 27B & 558/558 & 194.9 & 724 & 8,319 & 8,193 & 190 \\
GLM-5.1 & 561/561 & 27.9 & 690 & 1,671 & 1,245 & 531 \\
Qwen 3.5 122B & 561/561 & 62.2 & 724 & 6,727 & 2,401 & 12 \\
MiniMax M2.5 & 560/560 & 55.9 & 689 & 4,837.5 & 0 & 13 \\
Qwen 3 32B & 560/560 & 231.2 & 702 & 8,247.5 & 7,686.5 & 560 \\
Qwen 3.6 Plus & 561/561 & 14.0 & 726 & 1,135 & -- & 0 \\
\bottomrule
\end{tabular}}
\end{minipage}
\end{table}

\noindent\textbf{Generating code often takes longer than running it.} Table~\ref{tab:generation-profile-561} summarizes the final recorded API call for each response, including those that do not lead to a correct solution. Median \texttt{API latency} ranges from 14.0 to 231.2 seconds across models. Among the 4,202 correct runs with both \texttt{API latency} and \texttt{Build+opt} recorded, generation takes longer in 2,336 (55.6\%). The proportion reaches 76.2\% for Qwen 3.6 27B and 74.1\% for Qwen 3 32B, using each model's own correctly solved tasks in the cost subset. Reducing solver time alone therefore leaves a substantial part of the recorded waiting time unchanged. These final-call waiting times exclude earlier generation or repair calls; Appendix~\ref{sec:timing-contract} details their measurement scope.

\noindent\textbf{More output tokens do not necessarily mean longer waits.} Reported median prompt lengths vary from 689 to 937 tokens, while median completion lengths span 748--11,680 (Table~\ref{tab:generation-profile-561}). Kimi has a larger median completion count than Qwen 3 32B (11,680 versus 8,247.5), yet a much shorter median wait (82.2 versus 231.2 seconds). Output volume and elapsed waiting time thus capture different aspects of generation cost. These summaries do not isolate the causes: latency includes service and network waiting, and token definitions vary by provider. Reasoning-token coverage also differs sharply; Qwen 3.5 122B's median uses only 12 records, so that column does not support a uniform comparison of reasoning effort.

\noindent\textbf{The upper tail matters more than small differences in median memory.} Table~\ref{tab:matched-memory-561} covers all 4,202 correct cost-analysis runs after unit harmonization. Per-model median \texttt{Gurobi peak} values cluster between 107.5 and 139.1 MiB, but each model's P90 is 12.6--21.4 times its own median. Across all runs, the median is 128.5 MiB and P90 is 2,206.7 MiB; 224 runs (5.3\%) reach at least 4 GB. The common pattern is therefore a long upper tail, which a median-based memory budget would miss. These summaries use each model's own correctly solved tasks, so they do not isolate differences between models on identical workloads.

\noindent\textbf{The process-memory subset has a different distribution.} The 1,769 runs with all three \texttt{RSS} snapshots have a median solver allocation of 61.7 MiB, less than half the full-set median of 128.5 MiB. Their P90 remains similar, however: 2,238.2 versus 2,206.7 MiB. Thus, the matched subset retains demanding cases but does not represent the full distribution's center. We use the full set to describe solver-memory allocations and the matched subset to compare process and solver measurements on the same runs.

\noindent\textbf{Process snapshots and solver peaks answer different questions.} In the matched subset, median \texttt{RSS} snapshot maximum is 74.8 MiB and P90 is 1,328.6 MiB. Post-solve \texttt{RSS} is at least twice its pre-solve value in 412 runs (23.3\%), showing that memory held at the end of solving can substantially exceed the pre-solve level. These boundary snapshots can miss temporary peaks during optimization; \texttt{Gurobi peak} instead tracks solver-environment allocations. Neither their difference nor the difference between their quantiles measures Python memory overhead (Appendix~\ref{sec:timing-contract}).

\begin{table}[!htb]
\centering
\begin{minipage}{0.95\linewidth}
\centering\small
\setlength{\tabcolsep}{3.5pt}
\caption{\footnotesize Solver and process memory in MiB. The full set contains all 4,202 correct runs used for cost analysis. The matched subset contains 1,769 runs with both solver memory and process \texttt{RSS} sampled before preparation, before solving, and after solving; \texttt{RSS} is the largest of these three snapshots. \texttt{Gurobi peak} measures peak allocation within the solver environment. P90 is the 90th percentile. Overall pools runs across models; each model uses its own correctly solved tasks.}
\label{tab:matched-memory-561}
\resizebox{\linewidth}{!}{%
\begin{tabular}{lrrrrrrrr}
\toprule
 & \multicolumn{3}{c}{All correct cost-analysis runs} & \multicolumn{5}{c}{Matched subset with process \texttt{RSS}}\\
\cmidrule(lr){2-4}\cmidrule(lr){5-9}
 & & \multicolumn{2}{c}{\texttt{Gurobi peak}} & & \multicolumn{2}{c}{\texttt{RSS} snapshot max} & \multicolumn{2}{c}{\texttt{Gurobi peak}}\\
\cmidrule(lr){3-4}\cmidrule(lr){6-7}\cmidrule(lr){8-9}
Model & $n$ & Median & P90 & $n$ & Median & P90 & Median & P90\\
\midrule
Gemini 3.1 Pro & 504 & 124.7 & 2,397.9 & 210 & 79.2 & 2,011.6 & 72.0 & 2,403.3 \\
GPT-5.5 & 441 & 136.3 & 2,198.6 & 186 & 70.8 & 2,069.4 & 58.7 & 2,440.1 \\
Claude Opus 4.6 & 435 & 118.4 & 2,510.0 & 176 & 78.8 & 2,082.5 & 57.8 & 2,507.5 \\
DeepSeek-V4 Flash & 433 & 130.4 & 1,938.5 & 184 & 75.9 & 1,234.1 & 69.6 & 1,660.2 \\
Kimi K2.6 & 427 & 117.2 & 2,085.0 & 183 & 73.0 & 1,204.1 & 57.6 & 1,723.6 \\
Qwen 3.6 27B & 391 & 110.7 & 1,389.6 & 166 & 74.0 & 890.9 & 41.8 & 1,262.2 \\
GLM-5.1 & 382 & 136.1 & 2,402.8 & 157 & 80.0 & 1,937.9 & 67.1 & 2,475.4 \\
Qwen 3.5 122B & 367 & 132.6 & 1,727.3 & 134 & 73.7 & 1,147.5 & 56.1 & 1,918.8 \\
MiniMax M2.5 & 311 & 137.8 & 1,807.7 & 160 & 77.6 & 1,225.3 & 53.9 & 1,874.7 \\
Qwen 3 32B & 309 & 139.1 & 2,331.5 & 124 & 84.9 & 1,343.5 & 60.4 & 2,502.1 \\
Qwen 3.6 Plus & 202 & 107.5 & 2,305.8 & 89 & 73.4 & 1,051.0 & 53.0 & 2,037.1 \\
\midrule
Overall & 4,202 & 128.5 & 2,206.7 & 1,769 & 74.8 & 1,328.6 & 61.7 & 2,238.2 \\
\bottomrule
\end{tabular}}
\end{minipage}
\end{table}

\FloatBarrier

\subsection{Results for Fine-Tuned Models}
\label{app:specialized-open-models}

We evaluate five fine-tuned models on all 561 EfficientOpt tasks: LLMOPT (14B) \citep{jiang2025llmopt}, OptMATH (7B and 32B) \citep{lu2025optmath}, and SIRL (7B and 32B) \citep{chen2026sirl}. This section examines their accuracy and where their programs fail.

\noindent\textbf{Evaluation setup.} We execute the saved programs on the fixed task instances, using the correctness criterion and verified reference objectives from Section~\ref{sec:end-to-end}. Programs call Gurobi directly or through Pyomo's \texttt{gurobi\_direct} interface. Each run uses one solver thread, \texttt{Seed=0}, \texttt{MIPGap=0}, a 2\,GiB process-memory limit, and Gurobi \texttt{SoftMemLimit=1.5} (GB). Up to ten programs run concurrently. These runs contribute only to the accuracy and failure analysis; their timings are excluded from efficiency comparisons.

\begin{table}[!htb]
\centering\small
\setlength{\tabcolsep}{3pt}
\caption{Accuracy and execution outcomes for five fine-tuned models, with 561 tasks per model. Correct counts programs that finish with \texttt{OPTIMAL} status and the verified objective value. It is a subset of \texttt{OPTIMAL}, not an additional outcome category. The six columns from \texttt{OPTIMAL} through Other sum to 561 in each row.}
\label{tab:specialized-open-results}
\begin{tabular*}{\linewidth}{@{\extracolsep{\fill}}lccccccc@{}}
\toprule
Model & \shortstack{Correct\\$n$ (\%)} & \texttt{OPTIMAL} & \shortstack{Static\\failure} & \shortstack{Code\\error} & \shortstack{Time\\limit} & \shortstack{Memory\\limit} & Other\\
\midrule
OptMATH (32B) & 137 (24.42) & 189 & 11 & 212 & 51 & 57 & 41 \\
SIRL (32B) & 118 (21.03) & 188 & 34 & 164 & 77 & 72 & 26 \\
LLMOPT (14B) & 36 (6.42) & 77 & 20 & 377 & 28 & 35 & 24 \\
OptMATH (7B) & 1 (0.18) & 4 & 401 & 146 & 3 & 6 & 1 \\
SIRL (7B) & 0 (0.00) & 0 & 504 & 57 & 0 & 0 & 0 \\
\bottomrule
\end{tabular*}
\end{table}

\noindent\textbf{Correct solutions remain limited.} OptMATH (32B) and SIRL (32B) achieve the highest accuracy in this group, at 24.42\% and 21.03\%, respectively; the smaller models range from 0 to 6.42\% \mbox{(Table~\ref{tab:specialized-open-results})}. These small sets of correct solutions limit efficiency comparisons, motivating the separate accuracy and failure analysis.

\noindent\textbf{Optimal solver status does not ensure a correct answer.} Across the five models, 458 programs finish with \texttt{OPTIMAL} status, but 166 (36.2\%) return an objective that differs from the verified answer. As in Section~\ref{app:numerical-outcomes}, a solver can optimize the generated model successfully even when that model gives the wrong answer to the task.

\noindent\textbf{Failure patterns differ across models.} OptMATH (7B) and SIRL (7B) have 401 and 504 static failures, respectively: these programs cannot be compiled or lack the required modeling function. LLMOPT instead has 377 code errors, its most common outcome. Code errors include invalid data access and incorrect use of the modeling or solver API. For the two 32B models, time and memory limits also account for many outcomes: 108 runs for OptMATH and 149 for SIRL. These categories distinguish programs that cannot execute, programs that return incorrect objectives, and runs stopped by resource limits.

Time and memory limits include termination by either the execution process or Gurobi. Other covers the remaining solver and execution failures, including policy and logging interruptions.

\subsection{Slow Runs and Total Solver Time (RQ2)}
\label{app:slow-task-analysis}

We examine which runs account for most solver time and how the expert implementations perform on those same tasks. As in Appendix~\ref{app:cost-distributions}, we use 4,202 runs that pass the numerical checks and have valid \texttt{Runtime} measurements within the 543-task reference cost subset. For each LLM, we rank its $n$ eligible runs by \texttt{Runtime} and select the slowest $k=\lceil n/10\rceil$, approximately 10\%. Table~\ref{tab:slow-task-summary} summarizes their contribution to the model's total. Here, total time is the sum of recorded solver runtimes. It does not represent total program time or the elapsed time of a parallel batch.

\begin{table}[!htbp]
\centering\footnotesize
\setlength{\tabcolsep}{3pt}
\caption{Solver-time concentration for each LLM. Time share is the selected $k$ runs' percentage of total \texttt{Runtime}. Slower gives the number exceeding expert time by more than $10^{-6}$ seconds, out of $k$. LLM/expert divides summed LLM time by summed expert time on these same $k$ tasks; it is not the shifted geometric ratio in Table~\ref{tab:main-results-561}.}
\label{tab:slow-task-summary}
\begin{tabular}{lrrrrrrr}
\toprule
& \multicolumn{3}{c}{All eligible runs} & \multicolumn{4}{c}{Slowest $\sim$10\%} \\
\cmidrule(lr){2-4}\cmidrule(lr){5-8}
Model & $n$ & Median (s) & Total (h) & $k$ & Time share (\%) & Slower & LLM/expert \\
\midrule
Gemini 3.1 Pro & 504 & 32.5 & 27.67 & 51 & 72.4 & 48/51 & 5.43 \\
GPT-5.5 & 441 & 28.8 & 25.54 & 45 & 70.2 & 43/45 & 7.71 \\
Claude Opus 4.6 & 435 & 23.5 & 21.46 & 44 & 72.7 & 42/44 & 5.31 \\
DeepSeek-V4 Flash & 433 & 23.1 & 22.96 & 44 & 72.9 & 43/44 & 5.80 \\
Kimi K2.6 & 427 & 29.7 & 24.94 & 43 & 68.4 & 42/43 & 4.97 \\
Qwen 3.6 27B & 391 & 32.2 & 23.35 & 40 & 71.5 & 37/40 & 6.23 \\
GLM-5.1 & 382 & 27.6 & 16.00 & 39 & 68.6 & 37/39 & 3.45 \\
Qwen 3.5 122B & 367 & 26.0 & 20.40 & 37 & 74.0 & 33/37 & 6.96 \\
MiniMax M2.5 & 311 & 30.6 & 20.07 & 32 & 76.2 & 31/32 & 5.78 \\
Qwen 3 32B & 309 & 36.0 & 21.23 & 31 & 71.2 & 31/31 & 5.96 \\
Qwen 3.6 Plus & 202 & 34.2 & 12.02 & 21 & 69.7 & 20/21 & 4.84 \\
\bottomrule
\end{tabular}
\end{table}

\noindent\textbf{A few runs account for most solver time.} Across models, the slowest $\sim$10\% of runs consume 68.4--76.2\% of total solver time. Median runtimes are only 23.1--36.0 seconds, yet totals reach 12.0--27.7 hours, with just 11--24 runs per model accounting for at least half. These totals describe different task sets; Appendix~\ref{app:cost-distributions} compares models on shared tasks. Together, the selected sets contain 427 model--task runs covering 101 distinct tasks. T10 (convex hulls, envelopes, and convexification) contributes the largest share of selected-run time for every model: 24.6--50.5\%.

\noindent\textbf{Expert implementations are often faster on these slow tasks.} On each LLM's selected tasks, its summed \texttt{Runtime} is 3.45--7.71 times the expert total. Of the 427 selected runs, 407 are slower than the expert; 359 take at least twice as long, with a gap of at least one second. Table~\ref{tab:slow-task-cases} shows the slowest run for each LLM. Qwen 3.6 27B takes 7,065.35 seconds on \texttt{T07\_003}, versus 15.21 seconds for the expert. Other tasks are slow for both: Kimi K2.6 takes 4,324.49 seconds on \texttt{T06\_001}, versus 2,905.46 seconds for the expert. These comparisons describe tasks selected for high LLM runtime; they neither estimate the overall LLM--expert gap nor isolate the effect of a particular technique.

\begin{table}[!htbp]
\centering\footnotesize
\setlength{\tabcolsep}{5pt}
\caption{The slowest eligible run for each LLM and the expert's runtime on the same task. Times are Gurobi \texttt{Runtime}. Share is the selected run's percentage of that LLM's total eligible solver time. Runs are selected by LLM time, irrespective of the expert time.}
\label{tab:slow-task-cases}
\begin{tabular}{llrrr}
\toprule
Model & Task & LLM (s) & Expert (s) & Share (\%) \\
\midrule
Gemini 3.1 Pro & \texttt{T24\_002} & 4,677.80 & 303.66 & 4.7 \\
GPT-5.5 & \texttt{T25\_019} & 4,071.28 & 16.38 & 4.4 \\
Claude Opus 4.6 & \texttt{T48\_009} & 4,718.84 & 22.63 & 6.1 \\
DeepSeek-V4 Flash & \texttt{T15\_012} & 5,431.07 & 18.95 & 6.6 \\
Kimi K2.6 & \texttt{T06\_001} & 4,324.49 & 2,905.46 & 4.8 \\
Qwen 3.6 27B & \texttt{T07\_003} & 7,065.35 & 15.21 & 8.4 \\
GLM-5.1 & \texttt{T05\_020} & 4,736.70 & 285.27 & 8.2 \\
Qwen 3.5 122B & \texttt{T07\_003} & 6,501.74 & 15.21 & 8.9 \\
MiniMax M2.5 & \texttt{T21\_022} & 5,617.01 & 71.17 & 7.8 \\
Qwen 3 32B & \texttt{T48\_004} & 5,077.62 & 164.01 & 6.6 \\
Qwen 3.6 Plus & \texttt{T10\_010} & 3,079.45 & 236.71 & 7.1 \\
\bottomrule
\end{tabular}
\end{table}

\noindent\textbf{The slowest runs dominate total solver time without affecting the median.}
For each model, halving the recorded runtimes of its slowest 10\% of runs would reduce total solver time by 34.2--38.1\%, while leaving the median unchanged. Halving the runtimes of all other runs instead would reduce total solver time by only 11.9--15.8\%. Thus, the median alone obscures how a small number of slow runs drive total solver time.

\FloatBarrier

\section{Evaluation Metrics and Comparison Protocol}
\label{app:measurement-detail}

This appendix defines the measurement and comparison rules used in Section~\ref{sec:experiments} and Appendix~\ref{app:full-results-561}. Appendix~\ref{sec:timing-contract} explains what each metric captures; Appendix~\ref{app:paired-aggregation} specifies eligible comparisons, aggregation, and the limits of their interpretation.

\subsection{Metrics and Measurement Scope}
\label{sec:timing-contract}

We measure numerical accuracy, technique use, and computational cost separately. Table~\ref{tab:pipeline-metrics} summarizes the metrics; the following paragraphs explain their measurement boundaries.

\begin{table}[!htbp]
\centering\small
\setlength{\tabcolsep}{4pt}
\caption{Evaluation metrics and what they measure.}
\label{tab:pipeline-metrics}
\begin{tabular}{>{\raggedright\arraybackslash}p{.16\linewidth}>{\raggedright\arraybackslash}p{.32\linewidth}>{\raggedright\arraybackslash}p{.45\linewidth}}
\toprule
Quantity & Recorded measurement & What it captures \\
\midrule
Numerical check & \texttt{OPTIMAL} status and a finite objective matching the verified reference value & Numerical agreement on the tested instance; full task semantics and required decisions need separate verification. \\
Technique use & Automated labels from code and task context & Identifies target, alternative, none, or N/A; does not establish correctness or efficiency (Appendix~\ref{app:judge-full}). \\
Solver cost & Gurobi \texttt{Runtime} (s), \texttt{Work} (units) & Time and solver effort for the recorded optimization calls. \\
\texttt{Build+opt} & Preparation time + separately timed solver calls (s) & An execution subtotal; excludes work outside the recorded intervals. \\
Generation & \texttt{API latency} (s); provider-reported \texttt{Tokens} & Waiting time for the final recorded call; token usage reported for its response. Earlier calls are excluded. \\
Model size & Variables, constraints, linear-matrix nonzeros & Dimensions before presolve. Figure~\ref{fig:benchmark_profile} and Appendix~\ref{app:structure-diagnostic} use different counting scopes, explained below. \\
Memory & Process \texttt{RSS} snapshots; \texttt{Gurobi peak} & Sampled process-resident memory and peak solver-environment allocation, respectively. \\
\bottomrule
\end{tabular}
\end{table}
\noindent\emph{Preparation and solving.} \texttt{Build} measures preparation, including preprocessing, enumeration, and any solves or between-solve computation within the recorded intervals. \texttt{Build+opt} adds the elapsed time around separately timed solver calls, without adding Gurobi's \texttt{Runtime} again. This subtotal can omit input loading, imports, result extraction, and other work outside the timed intervals. Appendix~\ref{app:efficiency-definition} distinguishes it from complete execution and request time.

\noindent\emph{Solver time and effort.} Gurobi's \texttt{Runtime} and \texttt{Work} attributes describe the most recent optimization call \citep{gurobi_model_attributes}. We report the returned model's solve for single-model runs and sum the tracked calls for multi-solve procedures. Untracked solves inside model-building code may be absent from these totals even when their time is included in \texttt{Build}. \texttt{Work} measures solver effort in arbitrary units, not seconds or Python operations; it depends on the model, hardware, and solver settings.

\noindent\emph{Code generation.} \texttt{API latency} measures the final recorded client call, including service and network waiting and any retries and backoff within that call. It excludes earlier generation or repair calls. Token counts describe the returned response under each provider's conventions, rather than cumulative usage over retries. Reasoning tokens may already be included in output tokens, so we do not add those categories together. Unavailable counts remain missing.

\noindent\emph{Model size.} Figure~\ref{fig:benchmark_profile}(d,e) reports separate maxima of variable and total constraint counts across each reference's submitted models (Appendix~\ref{app:evaluation-populations}); the maxima need not come from the same model. Total constraints include linear, quadratic, general, and SOS constraints. The size--runtime analysis instead uses the variables, linear constraints, and linear-matrix nonzeros recorded with the timed solve (Appendix~\ref{app:structure-diagnostic}). All counts are before presolve.

\noindent\emph{Memory.} In the matched memory analysis, we report the maximum of three process \texttt{RSS} snapshots: before preparation, before the outer solver call, and after that call. These samples can miss intervening peaks. \texttt{Gurobi peak} is the maximum allocation within the solver environment, which may be shared by several models \citep{gurobi_model_attributes}; it is neither whole-process memory nor a sum of per-model peaks. Both are reported in MiB, converting native Gurobi GB values by $10^9/2^{20}$. Appendix~\ref{app:resource-distributions} gives the distributions and measurement coverage.

\subsection{Task Selection and Result Aggregation}
\label{app:paired-aggregation}

\noindent\emph{Accuracy and cost coverage.} Numerical accuracy uses all 561 tasks. Paired execution-cost comparisons use the 543 tasks whose recorded ordinary and expert runs reach agreeing optimal objectives (Appendix~\ref{app:evaluation-populations}).

\noindent\emph{Tasks included in each comparison.} Table~\ref{tab:main-results-561} requires numerically correct outputs and complete measurements for all four cost ratios. The ratios within each row use the same tasks, yielding 4,148 model--task pairs in total; task sets can differ between LLMs. Single-metric analyses require only the relevant measurements within the reference cost subset, so the solver-runtime analysis includes 4,202 correct runs.

Direct LLM comparisons use tasks solved correctly by both models, or by all 11 for the all-model comparison, with valid measurements of the relevant metric (Appendix~\ref{app:cost-distributions}).

\noindent\emph{Missing measurements.} Each comparison excludes missing or invalid measurements; these are never replaced with zero. We do not substitute time limits for incomplete or timed-out runs.

\Needspace{7\baselineskip}
\noindent\emph{Cost and size ratios.} Table~\ref{tab:main-results-561} uses shifted geometric ratios. For a cost or size metric $X$, let $X_{im}$ and $X_{ir}$ denote the measurements for LLM $m$ and reference $r$ on task $i$. The shifted geometric ratio over the eligible paired task set $S$ is
\begin{equation}
R_{m,r}^{X}(S)=\exp\!\left[\frac{1}{|S|}\sum_{i\in S}\log\frac{X_{im}+s_X}{X_{ir}+s_X}\right].
\end{equation}
This shifted geometric ratio gives each task equal weight in log space. The shift $s_X$ is one second for time, one solver work unit for \texttt{Work}, and one for counts. It accommodates zeros and limits the influence of tiny denominators; any unit conversion must rescale the shift as well as the measurements. A ratio below one indicates lower LLM cost or size under this aggregation. It is neither a ratio of summed costs nor an unshifted multiplicative speedup. Size ratios describe model structure and do not establish faster solving.

\noindent\emph{Runtime distributions and faster/slower counts.} Faster/slower counts compare unshifted measurements. Unless stated otherwise in a sensitivity analysis, differences of at most $10^{-6}$ in the metric's unit are ties; ties remain in the denominator of win fractions. Appendix~\ref{app:slow-task-analysis} defines the slowest-tenth selection and summed-time comparisons. 

\noindent\emph{Execution conditions.} All paired timing measurements were obtained on the same machine for the same task and numerical instance.

\noindent\emph{Interpreting cost differences.} Equal model-size counts do not establish formulation identity. Even algebraically equivalent formulations can differ in input order, starting solutions, or solver parameters that affect performance \citep{lodi2013variability,gurobi_variability}. Appendix~\ref{app:case-formulations} examines formulation differences; Appendix~\ref{app:case-construction-costs} shows how preparation affects costs.

\section{Dataset Quality Control and Expert Verification}
\label{app:quality-control}

This appendix describes the acceptance protocol in Figure~\ref{fig:quality-control-workflow}: independent expert review, mathematical and execution checks, and re-verification after repairs. Acceptance requires valid tasks and references and justified technique use; observed speedup is not a criterion.

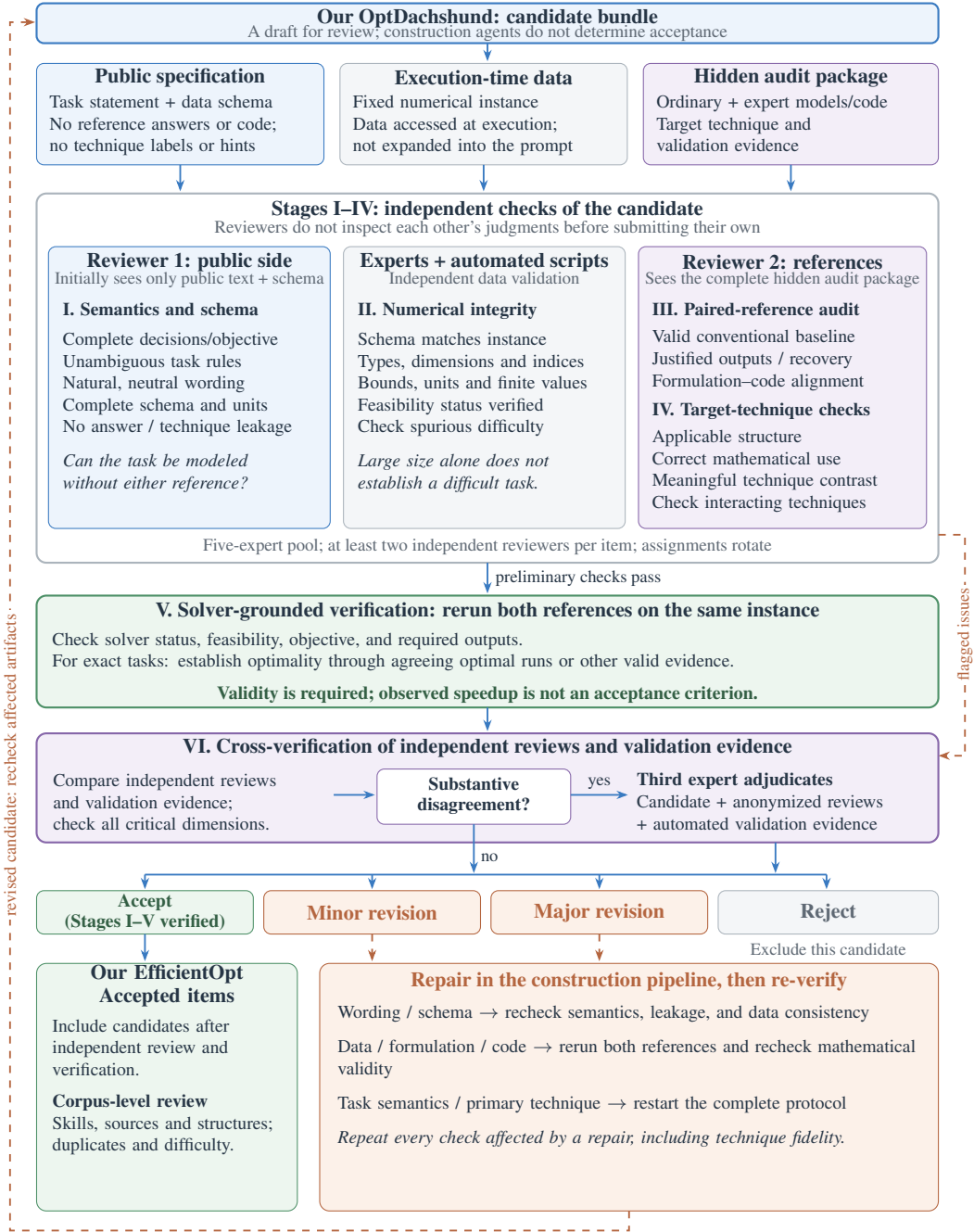
\begin{figure}[!htb]
\centering
\begingroup
\definecolor{qcink}{HTML}{253444}
\definecolor{qcmuted}{HTML}{5E6873}
\definecolor{qcblue}{HTML}{397AC1}
\definecolor{qcbluefill}{HTML}{EDF4FC}
\definecolor{qcpurple}{HTML}{8666A9}
\definecolor{qcpurplefill}{HTML}{F3EEF8}
\definecolor{qcgreen}{HTML}{4D9067}
\definecolor{qcgreenfill}{HTML}{EDF6EF}
\definecolor{qcrepair}{HTML}{B4653F}
\definecolor{qcrepairfill}{HTML}{FCF1E9}
\definecolor{qcrule}{HTML}{ABB4BC}
\definecolor{qcgrayfill}{HTML}{F4F6F8}
\newcommand{\qcfont}[2]{\fontsize{#1}{#2}\selectfont}
\resizebox{\textwidth}{!}{%
\begin{tikzpicture}[x=1cm,y=1cm,
  every node/.style={text=qcink,inner sep=0pt},
  qctitle/.style={font=\bfseries\qcfont{8.4}{9.5},align=center},
  qctext/.style={font=\qcfont{7.4}{8.9},align=left,anchor=north west},
  qcarrow/.style={-{Stealth[length=1.65mm,width=1.2mm]},draw=qcblue,line width=.7pt},
  qcreturn/.style={-{Stealth[length=1.65mm,width=1.2mm]},draw=qcrepair,dashed,line width=.7pt}]
\path[use as bounding box] (0,0) rectangle (14,-17.79);

\draw[rounded corners=3pt,draw=qcblue,fill=qcbluefill,line width=.8pt]
  (.55,-.03) rectangle (13.45,-.60);
\node[qctitle] at (7,-.24) {Our OptDachshund: candidate bundle};
\node[font=\qcfont{7.2}{8.2},text=qcmuted] at (7,-.46)
  {A draft for review; construction agents do not determine acceptance};
\draw[qcarrow] (7,-.60) -- (7,-.86);

\draw[rounded corners=3pt,draw=qcblue,fill=qcbluefill] (.55,-.89) rectangle (4.66,-2.34);
\node[qctitle] at (2.605,-1.09) {Public specification};
\node[qctext,text width=3.78cm] at (.73,-1.34)
  {Task statement + data schema\\No reference answers or code;\\no technique labels or hints};
\draw[rounded corners=3pt,draw=qcrule,fill=qcgrayfill] (4.89,-.89) rectangle (9.00,-2.34);
\node[qctitle] at (6.945,-1.09) {Execution-time data};
\node[qctext,text width=3.77cm] at (5.07,-1.34)
  {Fixed numerical instance\\Data accessed at execution;\\not expanded into the prompt};
\draw[rounded corners=3pt,draw=qcpurple,fill=qcpurplefill] (9.23,-.89) rectangle (13.45,-2.34);
\node[qctitle] at (11.34,-1.09) {Hidden audit package};
\node[qctext,text width=3.88cm] at (9.41,-1.34)
  {Ordinary + expert models/code\\Target technique and\\validation evidence};

\draw[qcarrow] (2.605,-2.34) -- (2.605,-2.72);
\draw[qcarrow] (6.945,-2.34) -- (6.945,-2.72);
\draw[qcarrow] (11.34,-2.34) -- (11.34,-2.72);
\draw[rounded corners=4pt,draw=qcrule,fill=white,line width=.8pt]
  (.55,-2.75) rectangle (13.45,-8.03);
\node[qctitle] at (7,-2.99) {Stages I--IV: independent checks of the candidate};
\node[font=\qcfont{7.2}{8.2},text=qcmuted] at (7,-3.25)
  {Reviewers do not inspect each other's judgments before submitting their own};

\draw[rounded corners=3pt,draw=qcblue,fill=qcbluefill] (.72,-3.51) rectangle (4.75,-7.54);
\node[qctitle] at (2.735,-3.73) {Reviewer 1: public side};
\node[font=\qcfont{7.0}{8.0},text=qcmuted] at (2.735,-3.98) {Initially sees only public text + schema};
\node[qctext,text width=3.64cm] at (.92,-4.31)
  {\textbf{I. Semantics and schema}\\[3pt]
   Complete decisions/objective\\
   Unambiguous task rules\\
   Natural, neutral wording\\
   Complete schema and units\\
   No answer / technique leakage\\[5pt]
   \textit{Can the task be modeled\\without either reference?}};

\draw[rounded corners=3pt,draw=qcrule,fill=qcgrayfill] (4.94,-3.51) rectangle (8.97,-7.54);
\node[qctitle] at (6.955,-3.73) {Experts + automated scripts};
\node[font=\qcfont{7.0}{8.0},text=qcmuted] at (6.955,-3.98) {Independent data validation};
\node[qctext,text width=3.64cm] at (5.14,-4.31)
  {\textbf{II. Numerical integrity}\\[3pt]
   Schema matches instance\\
   Types, dimensions and indices\\
   Bounds, units and finite values\\
   Feasibility status verified\\
   Check spurious difficulty\\[5pt]
   \textit{Large size alone does not\\establish a difficult task.}};

\draw[rounded corners=3pt,draw=qcpurple,fill=qcpurplefill] (9.16,-3.51) rectangle (13.28,-7.54);
\node[qctitle] at (11.22,-3.73) {Reviewer 2: references};
\node[font=\qcfont{7.0}{8.0},text=qcmuted] at (11.22,-3.98) {Sees the complete hidden audit package};
\node[qctext,text width=3.75cm] at (9.35,-4.31)
  {\textbf{III. Paired-reference audit}\\[2pt]
   Valid conventional baseline\\
   Justified outputs / recovery\\
   Formulation--code alignment\\[3pt]
   \textbf{IV. Target-technique checks}\\[2pt]
   Applicable structure\\
   Correct mathematical use\\
   Meaningful technique contrast\\
   Check interacting techniques};
\node[font=\qcfont{7.15}{8.2},text=qcmuted] at (7,-7.80)
  {Five-expert pool; at least two independent reviewers per item; assignments rotate};

\draw[qcarrow] (7,-8.03) -- node[right,font=\qcfont{7.0}{8.0},xshift=3pt]
  {preliminary checks pass} (7,-8.47);
\draw[qcreturn] (13.45,-7.63) -- (13.83,-7.63) -- (13.83,-10.79) -- (13.45,-10.79);
\node[rotate=90,font=\qcfont{6.8}{7.8},text=qcrepair,fill=white,inner sep=1pt]
  at (13.83,-9.05) {flagged issues};

\draw[rounded corners=4pt,draw=qcgreen,fill=qcgreenfill,line width=.8pt]
  (.55,-8.50) rectangle (13.45,-10.10);
\node[qctitle] at (7,-8.73) {V. Solver-grounded verification: rerun both references on the same instance};
\node[qctext,text width=12.0cm] at (.77,-9.02)
  {Check solver status, feasibility, objective, and required outputs.\\For exact tasks: establish optimality through agreeing optimal runs or other valid evidence.};
\node[font=\bfseries\qcfont{7.5}{8.5},text=qcgreen!65!black] at (7,-9.91)
  {Validity is required; observed speedup is not an acceptance criterion.};

\draw[qcarrow] (7,-10.10) -- (7,-10.44);
\draw[rounded corners=4pt,draw=qcpurple,fill=qcpurplefill,line width=.8pt]
  (.55,-10.47) rectangle (13.45,-12.03);
\node[qctitle] at (7,-10.68) {VI. Cross-verification of independent reviews and validation evidence};
\node[qctext,text width=4.15cm] at (.78,-11.04)
  {Compare independent reviews\\and validation evidence;\\check all critical dimensions.};
\draw[rounded corners=3pt,draw=qcpurple,fill=white] (5.42,-10.98) rectangle (8.19,-11.77);
\node[font=\bfseries\qcfont{7.4}{8.6},align=center] at (6.805,-11.36)
  {Substantive\\disagreement?};
\draw[qcarrow] (4.80,-11.35) -- (5.39,-11.35);
\draw[qcarrow] (8.19,-11.35) -- node[above,font=\qcfont{7.0}{8.0},yshift=2pt] {yes} (9.01,-11.35);
\node[qctext,text width=3.95cm] at (9.13,-11.03)
  {\textbf{Third expert adjudicates}\\Candidate + anonymized reviews\\+ automated validation evidence};
\draw[qcarrow] (6.805,-11.77) -- (6.805,-12.47);
\node[anchor=west,font=\qcfont{7.0}{8.0}] at (6.90,-12.24) {no};
\draw[qcarrow] (11.12,-12.03) -- (11.12,-12.47);

\draw[draw=qcblue,line width=.7pt] (2.10,-12.49) -- (11.85,-12.49);
\foreach \x in {2.10,5.35,8.60,11.85}{\draw[qcarrow] (\x,-12.49) -- (\x,-12.70);}
\draw[rounded corners=3pt,draw=qcgreen,fill=qcgreenfill] (.55,-12.72) rectangle (3.65,-13.34);
\node[font=\bfseries\qcfont{7.5}{8.4},align=center,text=qcgreen!65!black] at (2.10,-13.03) {Accept\\(Stages I--V verified)};
\draw[rounded corners=3pt,draw=qcrepair,fill=qcrepairfill] (3.80,-12.72) rectangle (6.90,-13.34);
\node[qctitle,text=qcrepair] at (5.35,-13.03) {Minor revision};
\draw[rounded corners=3pt,draw=qcrepair,fill=qcrepairfill] (7.05,-12.72) rectangle (10.15,-13.34);
\node[qctitle,text=qcrepair] at (8.60,-13.03) {Major revision};
\draw[rounded corners=3pt,draw=qcrule,fill=qcgrayfill] (10.30,-12.72) rectangle (13.45,-13.34);
\node[qctitle,text=qcmuted] at (11.875,-13.03) {Reject};
\draw[qcarrow] (2.10,-13.34) -- (2.10,-13.73);
\draw[qcreturn] (5.35,-13.34) -- (5.35,-13.73);
\draw[qcreturn] (8.60,-13.34) -- (8.60,-13.73);
\node[font=\qcfont{7.0}{8.0},text=qcmuted] at (11.875,-13.55) {Exclude this candidate};

\draw[rounded corners=4pt,draw=qcgreen,fill=qcgreenfill]
  (.55,-13.76) rectangle (4.28,-17.30);
\node[qctitle,text width=3.33cm] at (2.415,-14.08) {Our EfficientOpt\\Accepted items};
\node[qctext,text width=3.30cm] at (.77,-14.55)
  {Include candidates after\\independent review and\\verification.\\[4pt]
   \textbf{Corpus-level review}\\Skills, sources and structures;\\duplicates and difficulty.};

\draw[rounded corners=4pt,draw=qcrepair,fill=qcrepairfill]
  (4.60,-13.76) rectangle (13.45,-17.30);
\node[qctitle,text=qcrepair] at (9.025,-14.01) {Repair in the construction pipeline, then re-verify};
\node[qctext,text width=8.35cm] at (4.85,-14.36)
  {Wording / schema $\rightarrow$ recheck semantics, leakage, and data consistency\\[5pt]
   Data / formulation / code $\rightarrow$ rerun both references and recheck mathematical validity\\[5pt]
   Task semantics / primary technique $\rightarrow$ restart the complete protocol\\[5pt]
   \textit{Repeat every check affected by a repair, including technique fidelity.}};
\draw[qcreturn] (9.025,-17.30) -- (9.025,-17.58) -- (.17,-17.58) -- (.17,-.30) -- (.53,-.30);
\node[rotate=90,font=\qcfont{7.0}{8.0},text=qcrepair,fill=white,inner sep=1.5pt]
  at (.17,-10.92) {revised candidate: recheck affected artifacts};
\end{tikzpicture}%
}
\endgroup
\caption{\textbf{Dataset quality control and expert verification.} Checks cover required decision recovery and valid termination, not objective agreement alone. Repairs trigger all affected checks. Appendix~\ref{app:evaluation-populations} summarizes reference-answer verification and cost coverage.}
\label{fig:quality-control-workflow}
\end{figure}

\subsection{Independent Review and Information Access}

OptDachshund produces candidates; independent experts determine acceptance. The protocol draws from five experts in mathematical optimization, modeling, and solver \mbox{implementation}, assigning at least two to each candidate and rotating assignments across techniques and source groups. One reviewer initially sees only the public statement and schema, checking whether the task is understandable without its intended solution. The other sees both reference models and implementations, the target OptTips card, and validation evidence. They submit their assessments independently before comparison.
The review panel comprised five researchers, including doctoral students, postdoctoral researchers, and faculty members. Their collective expertise covered mathematical optimization, solver implementation, and LLM-based optimization modeling.

Evaluated LLMs receive the public statement and schema; their code accesses the numerical instance at execution. Reference solutions, objective values, technique labels, and expert hints remain hidden.

\subsection{Acceptance Criteria}

\noindent\textbf{Task and data validity (Stages I--II).} The public specification must define the decisions, objective, domains, constraints, and required outputs without relying on a hidden reference; material ambiguity requires revision or rejection. Automated checks and expert review verify schema--instance consistency, including dimensions, indices, units, bounds, and intended feasibility. The task may expose structure but must not prescribe the target technique. For example, stating that storage cartridges are \mbox{identical} describes the task; prescribing an order for their use variables reveals a symmetry-breaking formulation. Reviewers assess structural difficulty, including trivial reductions and whether recognizing them is the intended skill.

\noindent\textbf{Valid and comparable references (Stage III).} Both references must solve the same task on identical data and satisfy its output requirements. The ordinary reference must be a plausible conventional formulation, without artificial delays, invalid bounds, or numerical pathologies introduced to favor the expert. Experts check formulation--code consistency and the mathematical justification for each answer. Different feasible sets are permitted when the transformation is valid, including recovery of requested decisions and justified termination for decompositions (Appendix~\ref{app:modeling-definition}).

\noindent\textbf{Applicable and faithful technique use (Stage IV).} The expert reference must apply the OptTips mechanism to a structure that supports it. A tighter Big-$M$ constant, for example, must follow from valid bounds; a valid inequality must preserve feasible integer solutions. Reviewers check for a meaningful modeling contrast on the same task, data, and objective, accounting for techniques that interact with the primary target.

\noindent\textbf{Solver-grounded verification (Stage V).} The protocol requires executing both references on the same instance to check status, feasibility, objectives, and requested decisions. For exact tasks, optimality can be established by agreeing optimal runs, analytic or exact solutions, primal--dual certificates, or a checked feasible solution with a matching bound from an equivalent formulation. These alternatives allow answer verification when an ordinary run exceeds its time or memory budget. Multi-solve procedures require justified termination, such as a final pricing check for column generation. Appendix~\ref{app:evaluation-populations} summarizes reference-answer verification and cost coverage.

\subsection{Decisions, Repair, and Corpus Review}

Reviewers compare their independent findings with validation evidence (Stage VI). A third expert adjudicates substantive disagreements using the candidate, anonymized reviews, and validation results. The outcomes are acceptance, minor revision, major revision, or rejection. Minor revisions preserve task semantics; changes to data, formulations, code, or the target technique require major revision. Acceptance requires resolving critical findings and completing the relevant checks.

Repairs trigger renewed checks of affected content: wording and schema changes require semantic, leakage, and data-consistency review; data or implementation changes require rerunning both references and rechecking mathematical claims. Changes to task semantics or the primary technique restart the full protocol. Review decisions and supporting evidence remain linked to each task.

Corpus review checks duplicates and examines the distributions of techniques, source problems, mathematical structures, and task difficulty.

\section{Assessing Modeling Technique Use}
\label{app:technique-adjudication}

This appendix separates technique recognition from numerical correctness and computational cost. Appendix~\ref{app:judge-full} defines the automated labels and reports results by LLM and OptTips group; Appendix~\ref{app:judge-combined} combines these labels with execution and timing checks; Appendix~\ref{app:technique-cases} presents manual inspections of selected formulations and code.

\subsection{Automated Assessment of Technique Use}
\label{app:judge-full}

\noindent\textbf{Assessment protocol.} We report technique-use assessments for all 6,171 model--task pairs: 561 tasks for each of the 11 LLMs. A frontier LLM serves as judge. The recorded protocol supplies the problem statement, OptTips knowledge, candidate code, both reference implementations, and execution measurements.

\noindent\textbf{Label definitions.} Each record receives one of four labels. \emph{Target} indicates the designated technique or an implementation judged equivalent. \emph{Alternative} indicates another modeling technique; \emph{none} means neither is identified. \emph{N/A} covers cases judged inapplicable or unassessable. Labels include failed executions and incorrect objectives, so a target label does not certify a valid implementation or lower computational cost.

\noindent\textbf{Results across models and task groups.} Target, alternative, none, and N/A account for 46.3\%, 24.6\%, 21.3\%, and 7.8\% of all records, respectively (Figure~\ref{fig:diagnostic-costs}(b)). Target-label rates range from 33.9\% to 55.4\% across models, each using all 561 tasks as the denominator (Table~\ref{tab:judge-models}). Table~\ref{tab:judge-groups} gives counts for the 29 OptTips groups; Table~\ref{tab:groups-561} describes their task variants.

\begin{table}[!htbp]
\centering\small
\setlength{\tabcolsep}{3.1pt}
\caption{Technique-use assessments by LLM. Each model row classifies the same 561 records in two ways: technique labels (left) and combined diagnostic categories (right; Appendix~\ref{app:judge-combined}). The two column blocks each sum to 561 and must not be added together.}
\label{tab:judge-models}
\begin{tabular}{lrrrrrrrrr}
\toprule
& \multicolumn{4}{c}{LLM technique label} & \multicolumn{5}{c}{Combined diagnostic category} \\
\cmidrule(lr){2-5}\cmidrule(lr){6-10}
Model & Target & Alt. & None & N/A & C1 & C2 & C3 & C4 & C5 \\
\midrule
Claude Opus 4.6 & 311 & 131 & 98 & 21 & 253 & 52 & 66 & 68 & 122 \\
Gemini 3.1 Pro & 297 & 173 & 79 & 12 & 272 & 67 & 99 & 73 & 50 \\
DeepSeek-V4 Flash & 296 & 139 & 100 & 26 & 247 & 62 & 62 & 71 & 119 \\
Qwen 3.6 27B & 286 & 132 & 128 & 15 & 196 & 44 & 75 & 79 & 167 \\
GPT-5.5 & 284 & 147 & 106 & 24 & 223 & 68 & 71 & 82 & 117 \\
Kimi K2.6 & 282 & 138 & 85 & 56 & 233 & 59 & 74 & 65 & 130 \\
GLM-5.1 & 275 & 123 & 142 & 21 & 198 & 44 & 60 & 88 & 171 \\
Qwen 3 32B & 231 & 129 & 173 & 28 & 132 & 45 & 54 & 80 & 250 \\
MiniMax M2.5 & 204 & 138 & 109 & 110 & 147 & 40 & 78 & 50 & 246 \\
Qwen 3.5 122B & 202 & 133 & 137 & 89 & 164 & 48 & 78 & 78 & 193 \\
Qwen 3.6 Plus & 190 & 135 & 158 & 78 & 67 & 45 & 50 & 42 & 357 \\
\midrule
Total & 2,858 & 1,518 & 1,315 & 480 & 2,132 & 574 & 767 & 776 & 1,922 \\
\bottomrule\end{tabular}
\end{table}

\begin{table}[!htbp]
\centering\small
\setlength{\tabcolsep}{3.2pt}
\caption{Automated technique labels by OptTips group. $n$ includes all model--task records from 11 LLMs, including failed executions. Target gives count (\% of $n$); other columns give counts.}
\label{tab:judge-groups}
\begin{tabular}{llrrrrr}
\toprule
ID & Target OptTips group & $n$ & Target (\%) & Alt. & None & N/A \\
\midrule
T01 & Equality/inequality transformations & 352 & 283 (80.4) & 10 & 39 & 20 \\
T02 & Auxiliary variables and epigraphs & 165 & 156 (94.5) & 0 & 1 & 8 \\
T03 & Bound tightening and scaling & 209 & 12 (5.7) & 115 & 79 & 3 \\
T04 & LP relaxation & 275 & 23 (8.4) & 94 & 147 & 11 \\
T05 & Big-$M$ modeling & 209 & 171 (81.8) & 7 & 10 & 21 \\
T06 & Indicators, semicontinuous variables, SOS & 231 & 58 (25.1) & 149 & 15 & 9 \\
T07 & Valid inequalities and cuts & 176 & 134 (76.1) & 17 & 24 & 1 \\
T08 & Substitution and product linearization & 506 & 382 (75.5) & 54 & 43 & 27 \\
T09 & Piecewise-linear modeling & 88 & 78 (88.6) & 0 & 3 & 7 \\
T10 & Convex hulls and convexification & 330 & 137 (41.5) & 109 & 63 & 21 \\
T12 & Dual transformation & 66 & 1 (1.5) & 64 & 0 & 1 \\
T13 & Lagrangian relaxation & 165 & 0 (0.0) & 46 & 116 & 3 \\
T15 & Block decomposition & 264 & 0 (0.0) & 151 & 112 & 1 \\
T16 & Benders decomposition & 143 & 0 (0.0) & 75 & 68 & 0 \\
T17 & Dantzig--Wolfe / column generation & 55 & 0 (0.0) & 25 & 28 & 2 \\
T18 & Network flow and matching & 231 & 80 (34.6) & 72 & 70 & 9 \\
T19 & Covering, packing, and knapsack & 154 & 85 (55.2) & 51 & 15 & 3 \\
T20 & Scheduling and temporal formulations & 473 & 393 (83.1) & 22 & 19 & 39 \\
T21 & Routing and path formulations & 231 & 195 (84.4) & 0 & 8 & 28 \\
T22 & Robust counterparts & 297 & 174 (58.6) & 29 & 78 & 16 \\
T24 & Preprocessing and dominance & 198 & 4 (2.0) & 69 & 109 & 16 \\
T25 & Symmetry breaking & 209 & 114 (54.5) & 29 & 26 & 40 \\
T30 & Warm starts & 176 & 2 (1.1) & 139 & 14 & 21 \\
T34 & Slack and surplus variables & 176 & 135 (76.7) & 0 & 16 & 25 \\
T35 & Variable elimination & 143 & 14 (9.8) & 11 & 89 & 29 \\
T38 & Convex conjugacy and Fenchel duality & 77 & 49 (63.6) & 15 & 0 & 13 \\
T44 & Transportation and network simplex & 154 & 34 (22.1) & 38 & 54 & 28 \\
T48 & Subtour, cover, clique, and capacity cuts & 165 & 84 (50.9) & 27 & 9 & 45 \\
T49 & Linking and activation strengthening & 253 & 60 (23.7) & 100 & 60 & 33 \\
\midrule
Total & & 6,171 & 2,858 (46.3) & 1,518 & 1,315 & 480 \\
\bottomrule\end{tabular}
\end{table}

\noindent\textbf{Interpreting the labels.} Target labels can cover different implementations. In T20\_011, both Kimi's cumulative auxiliary variables and Qwen 3.5 122B's direct coverage constraints receive target labels (Appendix~\ref{app:technique-cases}).

\subsection{Combining Technique Labels with Execution Outcomes}
\label{app:judge-combined}

The C1--C5 columns in Table~\ref{tab:judge-models} combine technique labels with execution outcomes and solver times using the rules below. They are distinct from the judge's raw labels.

\noindent\textbf{Assignment rules.} A record must report \texttt{OPTIMAL} or \texttt{SOLVED} and pass the archived objective check to enter C1--C4. All other records, including errors and timeouts, enter \textbf{C5}.
\Needspace{5\baselineskip}
\begin{list}{\textbullet}{\setlength{\leftmargin}{1.2em}\setlength{\labelsep}{0.4em}\setlength{\topsep}{3pt}\setlength{\partopsep}{0pt}\setlength{\itemsep}{0pt}\setlength{\parsep}{0pt}}
\item \textbf{C1:} a target label, with no requirement for a runtime advantage.
\item \textbf{C2:} an alternative label meeting the timing criterion below.
\item \textbf{C3:} an alternative label not meeting that criterion.
\item \textbf{C4:} a none or N/A label.
\end{list}

\noindent\textbf{Timing criterion for C2.} Let $r=t_{\mathrm{LLM}}/t_{\mathrm{expert}}$ and $s=t_{\mathrm{LLM}}/t_{\mathrm{ordinary}}$ be unshifted solver \texttt{Runtime} ratios. An alternative enters C2 if $r\leq1.5$ or $(r<2\ \text{and}\ s<1)$. The first condition requires LLM and expert timings; the second also requires the ordinary-reference timing. Meeting this solver-time threshold does not guarantee a speedup over either reference and says nothing about generation or construction cost.

\noindent\textbf{Relation to the main results.} These categories retain the archived runs and objective checks, yielding 4,249 records in C1--C4. The main evaluation uses its own selection of runs and verified reference objectives, yielding 4,264 numerically correct records. The two sets are not interchangeable: these diagnostics do not replace the accuracy results in Appendix~\ref{app:numerical-outcomes} or the cost comparisons defined in Appendix~\ref{app:paired-aggregation}.

\subsection{Manual Inspection of Selected Modeling Choices}
\label{app:technique-cases}

We inspected variables, constraints, and objectives in selected formulations and executable code to identify the implemented modeling choices. Solver status, model size, and runtime alone cannot identify a technique; valid alternatives may differ from the expert reference.
Table~\ref{tab:inspected-techniques} summarizes three examples and points to their detailed analyses.

\begin{table}[!htbp]
\centering\small
\setlength{\tabcolsep}{4pt}
\caption{Selected modeling choices (RQ3). The cases are illustrative, not a random sample.}
\label{tab:inspected-techniques}
\begin{tabular}{p{.17\linewidth}p{.37\linewidth}p{.39\linewidth}}
\toprule
Task and output & Implemented choice & What the evidence supports \\
\midrule
T10\_011, GPT-5.5 & Binary pairwise products with separate product inequalities; the expert uses continuous interactions with row--column marginals. & Both formulations enforce the same products at integer assignments, but their LP relaxations differ. The GPT model does not use the expert's marginal formulation. \\
\addlinespace
T13\_006, Gemini & Primal resource-allocation LP with explicit implied upper bounds; the expert uses a dual LP. & The approaches attain the same objective with similar recorded \texttt{Build+opt} times. \\
\addlinespace
T20\_011, Kimi / Qwen 3.5 122B & Kimi uses cumulative auxiliary variables; Qwen uses direct coverage constraints. & Eliminating Kimi's auxiliaries preserves feasible integer shift-start decisions and the objective. Qwen's longer construction time largely offsets its solver-time advantage. \\
\bottomrule
\end{tabular}
\end{table}

\section{Case Studies on Modeling Choices and Efficiency}
\label{app:efficiency-audit}

These cases examine what alternative formulations preserve and how construction affects their recorded costs. See Appendix~\ref{sec:timing-contract} for timing scopes and Appendix~\ref{app:modeling-definition} for required outputs.

\subsection{What the Compared Formulations Preserve}
\label{app:case-formulations}

\noindent\textbf{Technology mixing (T13\_006).} The primal LP uses nonnegative technology shares summing to one per block, giving a bounded feasible region. On the evaluated instance, choosing each block's least-resource technology satisfies the resource cap, establishing feasibility. LP strong duality therefore gives equal optimal values for the primal and expert dual. The original technology shares correspond to the dual multipliers of the expert LP's mode constraints $u_i+\lambda r_{ik}\ge v_{ik}$, providing a direct recovery map.

\noindent\textbf{Department relocation (T10\_011).} For departments $p,q$ assigned to cities $i^*,j^*$, the expert's nonnegative interaction variables have row and column sums fixed by the assignments. This forces $y^{pq}_{i^*j^*}=1$ and all other entries to zero. Both formulations therefore enforce $y^{pq}_{ij}=x_{pi}x_{qj}$ at integer assignments and preserve the common objective. Their LP relaxations differ: for a single department pair with fractional assignments $(1/2,1/2)$ over two cities, the product inequalities allow $y=0$, but the marginal equalities do not. The product links already make interactions integral under binary assignments. The expert also changes the linking constraints, so the comparison does not isolate the effect of removing binary declarations.

\subsection{How Construction Changes the Cost Comparison}
\label{app:case-construction-costs}

Here, \texttt{Runtime} is solver time; \texttt{Build+opt} includes construction and externally timed optimization (Appendix~\ref{sec:timing-contract}). Appendix~\ref{app:resource-distributions} gives the corresponding benchmark-wide analysis.

\noindent\textbf{Shift scheduling (T20\_011).} Eliminating Kimi's cumulative auxiliary variables preserves the feasible integer shift-start decisions and objective of Qwen 3.5 122B's direct coverage model. Their solver times are 566.443 and 2.071 seconds, respectively, while their recorded \texttt{Build} times are 6.927 and 547.507 seconds. Their recorded \texttt{Build+opt} times are 573.362 and 549.577 seconds. The Kimi/Qwen ratio therefore shrinks from $273.512\times$ for solver time to $1.043\times$ for \texttt{Build+opt}: Qwen remains faster, but construction offsets most of its solver-time advantage.

\noindent\textbf{Transit duty scheduling (T20\_013).} The ordinary formulation introduces one continuous active-duty variable and one equality link for each feasible start--period pair. The expert formulation instead uses the integer start variables directly in the period-coverage constraints. The two formulations encode the same feasible start decisions and objective, and both attain an objective value of 68,835,006. The ordinary model has 1,620,420 variables and 1,620,420 constraints, compared with 180,000 of each in the expert model; their constraint matrices have 4,321,260 and 1,440,420 nonzeros, respectively. Their Gurobi \texttt{Runtime} values are 5.420 and 2.989 seconds, while their recorded \texttt{Build} times are 26.382 and 3.345 seconds. Their recorded \texttt{Build+opt} times are 31.803 and 6.334 seconds. The ordinary/expert ratio thus increases from 1.813 for \texttt{Runtime} to 5.021 for \texttt{Build+opt}. In this pair, the expert formulation is faster to build as well as to solve, so including Build widens its measured advantage.


\section{OptDachshund: Agent Roles and Information Flow}
\label{app:agents}

OptDachshund uses six agents to redesign source problems around OptTips techniques and construct paired references. Their roles follow the three stages in Section~\ref{subsec:multi_agent_framework} and Figure~\ref{fig:multi-agent}.

\begingroup\brokenpenalty=10000
\noindent\textbf{Technique Matcher.} The matcher selects OptTips techniques whose applicability conditions fit the source problem and proposes a redesign that exposes the relevant modeling challenge. Expected difficulty and efficiency gains guide this design; they remain hypotheses to test.
\par\endgroup

\noindent\textbf{Reformulation Designer.} Using the source problem and matching plan, the designer creates a task that requires a substantive modeling choice beyond rewording or numerical substitution. It writes the public statement and data schema, together with an internal formulation and intended solution.

\noindent\textbf{Quality Auditor.} The auditor checks mathematical validity and consistency across the candidate's statement, schema, formulation, and intended solution. It also checks applicability against the selected OptTips card, source traceability, technique leakage, duplicate risk, and diversity. A candidate proceeds to reference implementation after passing this design audit.

\noindent\textbf{Repair Agent.} This agent addresses the auditor's findings and returns the candidate for re-audit. Local repairs preserve the source problem and primary technique; changes to task semantics or the technique trigger the full review in Appendix~\ref{app:quality-control}.

\noindent\textbf{Reference Implementation Agent.} This agent builds ordinary and expert implementations for the same final task, fixed instance, and required outputs. The expert applies the selected technique. Reference implementations support both model-building and algorithmic interfaces, with cost measurements defined in Appendix~\ref{sec:timing-contract}.

\noindent\textbf{Reference Audit Agent.} This agent checks both references against the final task and selected OptTips card, using their formulations, code, and execution evidence. It examines executability, required outputs, task consistency, and valid technique use. Failures require revision and re-audit.

Target techniques, references, and review records remain hidden from evaluated LLMs, which receive the public task and data through the interface in Section~\ref{subsec:efficientopt}. Final acceptance requires independent expert review and solver-grounded verification, repeated after relevant repairs (Appendix~\ref{app:quality-control}). Agent approval alone is insufficient, and observed speedup is not required.


\subsection*{Example agent prompt (excerpt)}
\label{app:agent-prompts}

The excerpt below retains the Technique Matcher's main inputs and requested outputs, using OptTips terminology and descriptive runtime placeholders in double braces. Full prompts, schemas, and configurations are available in the public repository.\footnote{\url{https://github.com/ZhongLIFR/EfficientOpt}}

\Needspace{9\baselineskip}

\begin{optdprompt}{\href{https://github.com/ZhongLIFR/EfficientOpt}{Technique Matcher}}
[System message]
You are the Technique Matching Agent in a multi-agent framework for optimization-modeling benchmark reformulation. Select suitable OptTips techniques for one raw benchmark problem. Return only valid JSON.

[User message: selected fields]
{
  "source_problem": "{{source_problem}}",
  "required_candidate_count": "{{techniques_per_problem}}",
  "technique_matching_rules": "{{matching_rules}}",
  "coverage_rules": "{{coverage_rules}}",
  "available_techniques": "{{OptTips_cards}}",
  "output_schema": {
    "candidate_techniques": [
      {
        "tech_id": "Txx",
        "reason": "why this source naturally supports this technique",
        "expected_general_llm_failure": "e.g. enumeration, loose Big-M, no dualization",
        "transformation_operator": "short operator name"
      }
    ]
  }
}
\end{optdprompt}


\FloatBarrier

\section{Task Validity and Computational Cost}
\label{app:formal-task}

This appendix expands Definitions~1 and~2 by explaining what a solution procedure must preserve and which stages contribute to its cost.

\subsection{Valid Solution Procedures}
\label{app:modeling-definition}

For a task $p=(q,I,\mathcal O)$, validity is judged against the original problem specification $q$, the fixed input $I$, and the required outputs and accuracy $\mathcal O$. If the task requests decisions, the returned decisions must satisfy the original domains and constraints and attain the required objective quality. Reporting the optimal value alone does not meet that requirement; any requested certificate must also be provided.

The solution procedure $\widehat{\mathcal A}=(\widehat{\mathcal M},\widehat C)$ combines a mathematical representation with executable code. Its variables and feasible set may differ from those of the original task, provided the transformation is justified and the code recovers the required answer. For example, an extended formulation may introduce auxiliary variables, with projection recovering the original decisions. A dual formulation requires the relevant duality conditions to justify equality of optimal values and, when original decisions are requested, a procedure for recovering them. A decomposition requires coordinated subproblem solves and a stopping rule that establishes the required solution quality for the full task.

\subsection{Resource Accounting and Comparison Scope}
\label{app:efficiency-definition}

An efficiency comparison concerns the same task, numerical input, and required answer quality, with an explicit resource and accounting scope. Time and memory are assessed separately. Costs depend on implementation and execution conditions as well as the formulation.

\Needspace{9\baselineskip}
\noindent\textbf{Complete request time.} For a fully traced serial request, time can be partitioned into nonoverlapping phases, accumulated over all attempts and repairs:
\begin{equation}
\begin{aligned}
T_{\rm request}&=T_{\rm API}+T_{\rm orchestration}+T_{\rm execution}+T_{\rm check},\\
T_{\rm execution}&=T_{\rm setup/data}+T_{\rm build}+T_{\rm opt}+T_{\rm between/recover}.
\end{aligned}
\label{eq:e2e-boundary}
\end{equation}
Here, $T_{\rm API}$ sums generation-call latency, including service and network waiting. Orchestration and checking cover work outside generation and execution. Within execution, $T_{\rm build}$ is model construction; $T_{\rm opt}$ includes every solver call, including those invoked during construction or within a decomposition. Computation between calls and answer recovery belong to $T_{\rm between/recover}$. A recorded \texttt{Build} interval may span several of these phases; solver time already included in that interval must not be counted again.

\noindent\textbf{Memory and modeling tradeoffs.} Memory measurements require a stated scope, such as process memory or solver-managed allocation, and a sampling method; stagewise peaks cannot simply be summed. A technique may reduce solver time while increasing construction time or memory. Model-size counts help describe such changes but do not measure efficiency, and an expert reference need not be the least costly valid implementation.

\noindent\textbf{Scope of the reported evidence.} The final-call generation measurements and recorded execution and memory observations do not reconstruct the complete request budget in Equation~\ref{eq:e2e-boundary}. Appendix~\ref{sec:timing-contract} defines their measurement scopes; Appendix~\ref{app:paired-aggregation} specifies eligible comparisons, aggregation, and execution conditions.

\section{OptTips: Technique Catalog, Coverage, and Extension}
\label{app:opttips-catalog}

This appendix describes the coverage and structure of OptTips and lists all 50 techniques in Table~\ref{tab:opttips-catalog}. The complete cards are available in our public repository.

\begingroup
\setlength{\parskip}{3pt}
\subsection{Rationale and Coverage}

\begin{table}[!htb]
\centering
\begin{minipage}{0.6\linewidth}
\centering\small
\setlength{\tabcolsep}{4pt}
\caption{\footnotesize Primary reporting families of OptTips. Codes identify the assignments in Table~\ref{tab:opttips-catalog}; techniques can apply across families.}
\label{tab:opttips-families}
\resizebox{\linewidth}{!}{%
\begin{tabular}{cl r}
\toprule
Code & Methodological family & Cards \\
\midrule
A & Formulation standardization and dimension reduction & 6 \\
B & Relaxation strengthening and convexification & 5 \\
C & Duality, optimality, and equilibrium & 4 \\
D & Specialized combinatorial and structured models & 8 \\
E & Discrete and logical modeling & 4 \\
F & Convex, nonlinear, and structure-preserving modeling & 9 \\
G & Decomposition and large-scale optimization & 6 \\
H & Robust, stochastic, multiobjective, and solver-aware modeling & 8 \\
\midrule
& Total & 50 \\
\bottomrule
\end{tabular}}
\end{minipage}
\end{table}

The collection combines three kinds of coverage. \emph{Foundational coverage} includes recurring LP/MIP choices such as bounds, Big-$M$, linearization, cuts, and variable elimination. \emph{Mechanism diversity} includes changes to representations, relaxation strength, and the organization of computation through duality and decomposition. \emph{Broader methodological scope} includes uncertainty, equilibrium, nonsmooth, and matrix/tensor modeling. Table~\ref{tab:opttips-families} groups the cards into eight families; the collection is a starting point, not an exhaustive catalog.

Each card receives one \emph{primary reporting assignment} for counting, even when it applies across families. These counts describe knowledge coverage, not equally frequent or independent competencies. The reported Gurobi evaluation uses 561 tasks covering 29 cards. The remaining 21 cards have 63 supplementary tasks, three per card, outside the reported performance analysis. This count excludes extension tasks for already-covered cards.

\subsection{Card Structure and Extension}

Every card uses the same seven components shown in Figure~\ref{fig:opttips-structure}: \emph{identifier}, \emph{core idea}, \emph{applicable patterns}, \emph{inefficiency symptoms}, \emph{expert actions}, \emph{mathematical example}, and \emph{benchmark guidance}. Applicable patterns state the required structure and conditions; symptoms identify the modeling issue; expert actions specify the transformation. The mathematical example illustrates the transformation; benchmark guidance covers paired implementation, validation, and performance measurement.

Figures~\ref{fig:opttips-eight-cards} and~\ref{fig:opttips-cards-eh} give one worked example from each family, including applicability conditions, shared model terms, mathematical justification, and benchmark guidance. Case identifiers and recorded solver \texttt{Runtime} values refer to the illustrated instances.

New cards can extend the collection by following this schema and stating their mathematical justification. Associated tasks and paired references follow the validation process in Appendix~\ref{app:quality-control}.

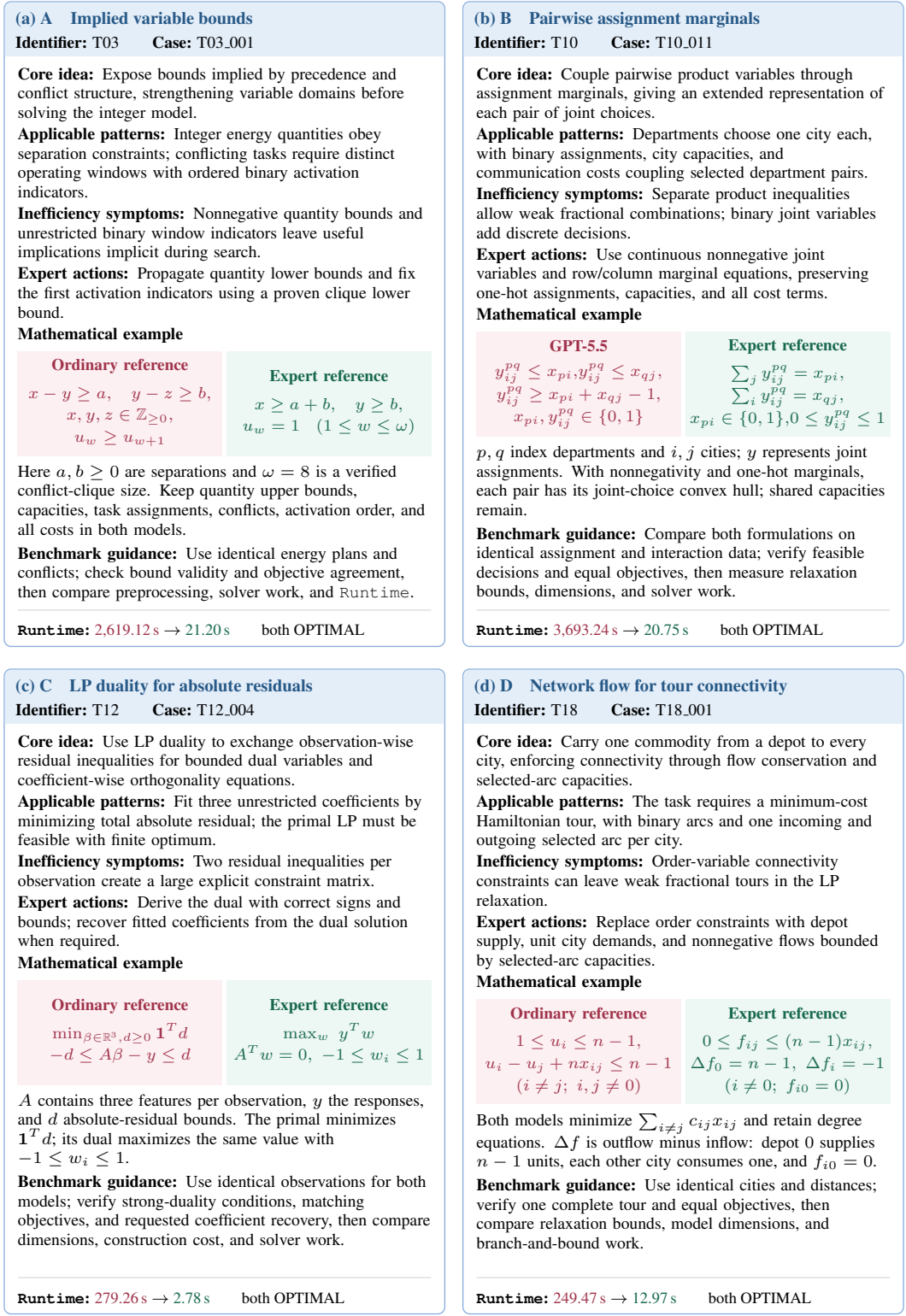
\begin{figure}[!htbp]
\centering
\begingroup
\definecolor{cardblue}{HTML}{255785}
\definecolor{cardline}{HTML}{89AEDD}
\definecolor{cardhead}{HTML}{E4EEF9}
\definecolor{cardbefore}{HTML}{FCF0F2}
\definecolor{cardafter}{HTML}{EDF7F3}
\definecolor{cardred}{HTML}{9F3048}
\definecolor{cardgreen}{HTML}{14664D}
\newcommand{\expandedfield}[2]{\noindent\textbf{#1:} #2\par\vspace{1pt}}
\newcommand{\expandedmath}{\fontsize{7.1}{8.3}\selectfont\renewcommand{\arraystretch}{1.16}\setlength{\arraycolsep}{0pt}}
\begin{tikzpicture}[x=1pt,y=1pt,every node/.style={inner sep=0pt,outer sep=0pt}]

\begin{scope}[shift={(0,0)}]
\path[draw=cardline,line width=.5pt,fill=white,rounded corners=3pt] (0,0) rectangle (194,-284);
\begin{scope}\clip[rounded corners=3pt] (.3,-.3) rectangle (193.7,-283.7);
\fill[cardhead] (0,0) rectangle (194,-24);\end{scope}
\node[anchor=west,font=\fontsize{7.7}{8.6}\selectfont\bfseries,text=cardblue] at (5,-8) {(a) A\quad Implied variable bounds};
\node[anchor=west,font=\fontsize{7.3}{8.2}\selectfont] at (5,-18) {\textbf{Identifier:} T03\qquad\textbf{Case:} T03\_001};
\node[anchor=north west,text width=182pt,align=left,font=\fontsize{7.4}{8.5}\selectfont] (expandedbody0) at (6,-29) {%
\begin{minipage}{182pt}\raggedright\setlength{\parindent}{0pt}\setlength{\parskip}{0pt}
\expandedfield{Core idea}{Expose bounds implied by precedence and conflict structure, strengthening variable domains before solving the integer model.}
\expandedfield{Applicable patterns}{Integer energy quantities obey separation constraints; conflicting tasks require distinct operating windows with ordered binary activation indicators.}
\expandedfield{Inefficiency symptoms}{Nonnegative quantity bounds and unrestricted binary window indicators leave useful implications implicit during search.}
\expandedfield{Expert actions}{Propagate quantity lower bounds and fix the first activation indicators using a proven clique lower bound.}
\textbf{Mathematical example}\par\vspace{3pt}
\begingroup\setlength{\fboxsep}{1pt}%
\colorbox{cardbefore}{\begin{minipage}[c][44pt][c]{88pt}\centering
{\fontsize{7.2}{8.2}\selectfont\bfseries\color{cardred}Ordinary reference}\par\vspace{3pt}
{\expandedmath\color{cardred}$\begin{array}{c}x-y\ge a,\quad y-z\ge b,\\x,y,z\in\mathbb Z_{\ge0},\\u_w\ge u_{w+1}\end{array}$}\end{minipage}}%
\hfill%
\colorbox{cardafter}{\begin{minipage}[c][44pt][c]{88pt}\centering
{\fontsize{7.2}{8.2}\selectfont\bfseries\color{cardgreen}Expert reference}\par\vspace{3pt}
{\expandedmath\color{cardgreen}$\begin{array}{c}x\ge a+b,\quad y\ge b,\\u_w=1\quad(1\le w\le\omega)\end{array}$}\end{minipage}}\endgroup\par\vspace{3pt}
Here $a,b\ge0$ are separations and $\omega=8$ is a verified conflict-clique size. Keep quantity upper bounds, capacities, task assignments, conflicts, activation order, and all costs in both models.\par\vspace{2pt}
\expandedfield{Benchmark guidance}{Use identical energy plans and conflicts; check bound validity and objective agreement, then compare preprocessing, solver work, and \texttt{Runtime}.}
\end{minipage}};
\path let \p1=(expandedbody0.south),\p2=(expandedbody0.north) in \pgfextra{\typeout{CARD-BODY-0-HEIGHT=\the\dimexpr\y2-\y1\relax}};
\draw[gray!35,line width=.4pt] (6,-268)--(188,-268);
\node[anchor=west,font=\fontsize{7.0}{8.0}\selectfont] at (6,-277) {\textbf{\texttt{Runtime}:} \textcolor{cardred}{2,619.12\,s}\ $\rightarrow$\ \textcolor{cardgreen}{21.20\,s}\qquad both OPTIMAL};
\end{scope}
\begin{scope}[shift={(202,0)}]
\path[draw=cardline,line width=.5pt,fill=white,rounded corners=3pt] (0,0) rectangle (194,-284);
\begin{scope}\clip[rounded corners=3pt] (.3,-.3) rectangle (193.7,-283.7);
\fill[cardhead] (0,0) rectangle (194,-24);\end{scope}
\node[anchor=west,font=\fontsize{7.7}{8.6}\selectfont\bfseries,text=cardblue] at (5,-8) {(b) B\quad Pairwise assignment marginals};
\node[anchor=west,font=\fontsize{7.3}{8.2}\selectfont] at (5,-18) {\textbf{Identifier:} T10\qquad\textbf{Case:} T10\_011};
\node[anchor=north west,text width=182pt,align=left,font=\fontsize{7.4}{8.5}\selectfont] (expandedbody1) at (6,-29) {%
\begin{minipage}{182pt}\raggedright\setlength{\parindent}{0pt}\setlength{\parskip}{0pt}
\expandedfield{Core idea}{Couple pairwise product variables through assignment marginals, giving an extended representation of each pair of joint choices.}
\expandedfield{Applicable patterns}{Departments choose one city each, with binary assignments, city capacities, and communication costs coupling selected department pairs.}
\expandedfield{Inefficiency symptoms}{Separate product inequalities allow weak fractional combinations; binary joint variables add discrete decisions.}
\expandedfield{Expert actions}{Use continuous nonnegative joint variables and row/column marginal equations, preserving one-hot assignments, capacities, and all cost terms.}
\textbf{Mathematical example}\par\vspace{3pt}
\begingroup\setlength{\fboxsep}{1pt}%
\colorbox{cardbefore}{\begin{minipage}[c][44pt][c]{88pt}\centering
{\fontsize{7.2}{8.2}\selectfont\bfseries\color{cardred}GPT-5.5}\par\vspace{3pt}
{\expandedmath\color{cardred}$\begin{array}{c}y_{ij}^{pq}\le x_{pi},\!y_{ij}^{pq}\le x_{qj},\\y_{ij}^{pq}\ge x_{pi}+x_{qj}-1,\\x_{pi},y_{ij}^{pq}\in\{0,1\}\end{array}$}\end{minipage}}%
\hfill%
\colorbox{cardafter}{\begin{minipage}[c][44pt][c]{88pt}\centering
{\fontsize{7.2}{8.2}\selectfont\bfseries\color{cardgreen}Expert reference}\par\vspace{3pt}
{\expandedmath\color{cardgreen}$\begin{array}{c}\sum_j y_{ij}^{pq}=x_{pi},\\\sum_i y_{ij}^{pq}=x_{qj},\\x_{pi}\in\{0,1\},\!0\le y_{ij}^{pq}\le1\end{array}$}\end{minipage}}\endgroup\par\vspace{3pt}
$p,q$ index departments and $i,j$ cities; $y$ represents joint assignments. With nonnegativity and one-hot marginals, each pair has its joint-choice convex hull; shared capacities remain.\par\vspace{2pt}
\expandedfield{Benchmark guidance}{Compare both formulations on identical assignment and interaction data; verify feasible decisions and equal objectives, then measure relaxation bounds, dimensions, and solver work.}
\end{minipage}};
\path let \p1=(expandedbody1.south),\p2=(expandedbody1.north) in \pgfextra{\typeout{CARD-BODY-1-HEIGHT=\the\dimexpr\y2-\y1\relax}};
\draw[gray!35,line width=.4pt] (6,-268)--(188,-268);
\node[anchor=west,font=\fontsize{7.0}{8.0}\selectfont] at (6,-277) {\textbf{\texttt{Runtime}:} \textcolor{cardred}{3,693.24\,s}\ $\rightarrow$\ \textcolor{cardgreen}{20.75\,s}\qquad both OPTIMAL};
\end{scope}
\begin{scope}[shift={(0,-294)}]
\path[draw=cardline,line width=.5pt,fill=white,rounded corners=3pt] (0,0) rectangle (194,-284);
\begin{scope}\clip[rounded corners=3pt] (.3,-.3) rectangle (193.7,-283.7);
\fill[cardhead] (0,0) rectangle (194,-24);\end{scope}
\node[anchor=west,font=\fontsize{7.7}{8.6}\selectfont\bfseries,text=cardblue] at (5,-8) {(c) C\quad LP duality for absolute residuals};
\node[anchor=west,font=\fontsize{7.3}{8.2}\selectfont] at (5,-18) {\textbf{Identifier:} T12\qquad\textbf{Case:} T12\_004};
\node[anchor=north west,text width=182pt,align=left,font=\fontsize{7.4}{8.5}\selectfont] (expandedbody2) at (6,-29) {%
\begin{minipage}{182pt}\raggedright\setlength{\parindent}{0pt}\setlength{\parskip}{0pt}
\expandedfield{Core idea}{Use LP duality to exchange observation-wise residual inequalities for bounded dual variables and coefficient-wise orthogonality equations.}
\expandedfield{Applicable patterns}{Fit three unrestricted coefficients by minimizing total absolute residual; the primal LP must be feasible with finite optimum.}
\expandedfield{Inefficiency symptoms}{Two residual inequalities per observation create a large explicit constraint matrix.}
\expandedfield{Expert actions}{Derive the dual with correct signs and bounds; recover fitted coefficients from the dual solution when required.}
\textbf{Mathematical example}\par\vspace{3pt}
\begingroup\setlength{\fboxsep}{1pt}%
\colorbox{cardbefore}{\begin{minipage}[c][44pt][c]{88pt}\centering
{\fontsize{7.2}{8.2}\selectfont\bfseries\color{cardred}Ordinary reference}\par\vspace{3pt}
{\expandedmath\color{cardred}$\begin{array}{c}\min_{\beta\in\mathbb R^3,d\ge0}\mathbf1^Td\\-d\le A\beta-y\le d\end{array}$}\end{minipage}}%
\hfill%
\colorbox{cardafter}{\begin{minipage}[c][44pt][c]{88pt}\centering
{\fontsize{7.2}{8.2}\selectfont\bfseries\color{cardgreen}Expert reference}\par\vspace{3pt}
{\expandedmath\color{cardgreen}$\begin{array}{c}\max_w\ y^Tw\\A^Tw=0,\ -1\le w_i\le1\end{array}$}\end{minipage}}\endgroup\par\vspace{3pt}
$A$ contains three features per observation, $y$ the responses, and $d$ absolute-residual bounds. The primal minimizes $\mathbf1^Td$; its dual maximizes the same value with $-1\le w_i\le1$.\par\vspace{2pt}
\expandedfield{Benchmark guidance}{Use identical observations for both models; verify strong-duality conditions, matching objectives, and requested coefficient recovery, then compare dimensions, construction cost, and solver work.}
\end{minipage}};
\path let \p1=(expandedbody2.south),\p2=(expandedbody2.north) in \pgfextra{\typeout{CARD-BODY-2-HEIGHT=\the\dimexpr\y2-\y1\relax}};
\draw[gray!35,line width=.4pt] (6,-268)--(188,-268);
\node[anchor=west,font=\fontsize{7.0}{8.0}\selectfont] at (6,-277) {\textbf{\texttt{Runtime}:} \textcolor{cardred}{279.26\,s}\ $\rightarrow$\ \textcolor{cardgreen}{2.78\,s}\qquad both OPTIMAL};
\end{scope}
\begin{scope}[shift={(202,-294)}]
\path[draw=cardline,line width=.5pt,fill=white,rounded corners=3pt] (0,0) rectangle (194,-284);
\begin{scope}\clip[rounded corners=3pt] (.3,-.3) rectangle (193.7,-283.7);
\fill[cardhead] (0,0) rectangle (194,-24);\end{scope}
\node[anchor=west,font=\fontsize{7.7}{8.6}\selectfont\bfseries,text=cardblue] at (5,-8) {(d) D\quad Network flow for tour connectivity};
\node[anchor=west,font=\fontsize{7.3}{8.2}\selectfont] at (5,-18) {\textbf{Identifier:} T18\qquad\textbf{Case:} T18\_001};
\node[anchor=north west,text width=182pt,align=left,font=\fontsize{7.4}{8.5}\selectfont] (expandedbody3) at (6,-29) {%
\begin{minipage}{182pt}\raggedright\setlength{\parindent}{0pt}\setlength{\parskip}{0pt}
\expandedfield{Core idea}{Carry one commodity from a depot to every city, enforcing connectivity through flow conservation and selected-arc capacities.}
\expandedfield{Applicable patterns}{The task requires a minimum-cost Hamiltonian tour, with binary arcs and one incoming and outgoing selected arc per city.}
\expandedfield{Inefficiency symptoms}{Order-variable connectivity constraints can leave weak fractional tours in the LP relaxation.}
\expandedfield{Expert actions}{Replace order constraints with depot supply, unit city demands, and nonnegative flows bounded by selected-arc capacities.}
\textbf{Mathematical example}\par\vspace{3pt}
\begingroup\setlength{\fboxsep}{1pt}%
\colorbox{cardbefore}{\begin{minipage}[c][44pt][c]{88pt}\centering
{\fontsize{7.2}{8.2}\selectfont\bfseries\color{cardred}Ordinary reference}\par\vspace{3pt}
{\expandedmath\color{cardred}$\begin{array}{c}1\le u_i\le n-1,\\u_i-u_j+n x_{ij}\le n-1\\(i\ne j;\ i,j\ne0)\end{array}$}\end{minipage}}%
\hfill%
\colorbox{cardafter}{\begin{minipage}[c][44pt][c]{88pt}\centering
{\fontsize{7.2}{8.2}\selectfont\bfseries\color{cardgreen}Expert reference}\par\vspace{3pt}
{\expandedmath\color{cardgreen}$\begin{array}{c}0\le f_{ij}\le(n-1)x_{ij},\\\Delta f_0=n-1,\ \Delta f_i=-1\\(i\ne0;\ f_{i0}=0)\end{array}$}\end{minipage}}\endgroup\par\vspace{3pt}
Both models minimize $\sum_{i\ne j}c_{ij}x_{ij}$ and retain degree equations. $\Delta f$ is outflow minus inflow: depot $0$ supplies $n-1$ units, each other city consumes one, and $f_{i0}=0$.\par\vspace{2pt}
\expandedfield{Benchmark guidance}{Use identical cities and distances; verify one complete tour and equal objectives, then compare relaxation bounds, model dimensions, and branch-and-bound work.}
\end{minipage}};
\path let \p1=(expandedbody3.south),\p2=(expandedbody3.north) in \pgfextra{\typeout{CARD-BODY-3-HEIGHT=\the\dimexpr\y2-\y1\relax}};
\draw[gray!35,line width=.4pt] (6,-268)--(188,-268);
\node[anchor=west,font=\fontsize{7.0}{8.0}\selectfont] at (6,-277) {\textbf{\texttt{Runtime}:} \textcolor{cardred}{249.47\,s}\ $\rightarrow$\ \textcolor{cardgreen}{12.97\,s}\qquad both OPTIMAL};
\end{scope}
\end{tikzpicture}
\endgroup
\caption{\footnotesize \textbf{OptTips cards for families A--D.} Each expanded card includes all seven components of the schema. Paired models highlight the transformation, with shared terms described in the card. Panel (b) retains the GPT-5.5--expert comparison from Figure~\ref{fig:introduction}; the other panels use ordinary and expert references. Recorded solver \texttt{Runtime} values have matching reported objective values.}
\label{fig:opttips-eight-cards}
\end{figure}

\begin{figure}[!htbp]
\centering
\begingroup
\definecolor{cardblue}{HTML}{255785}
\definecolor{cardline}{HTML}{89AEDD}
\definecolor{cardhead}{HTML}{E4EEF9}
\definecolor{cardbefore}{HTML}{FCF0F2}
\definecolor{cardafter}{HTML}{EDF7F3}
\definecolor{cardred}{HTML}{9F3048}
\definecolor{cardgreen}{HTML}{14664D}
\newcommand{\expandedfield}[2]{\noindent\textbf{#1:} #2\par\vspace{1pt}}
\newcommand{\expandedmath}{\fontsize{7.1}{8.3}\selectfont\renewcommand{\arraystretch}{1.16}\setlength{\arraycolsep}{0pt}}
\begin{tikzpicture}[x=1pt,y=1pt,every node/.style={inner sep=0pt,outer sep=0pt}]

\begin{scope}[shift={(0,0)}]
\path[draw=cardline,line width=.5pt,fill=white,rounded corners=3pt] (0,0) rectangle (194,-284);
\begin{scope}\clip[rounded corners=3pt] (.3,-.3) rectangle (193.7,-283.7);
\fill[cardhead] (0,0) rectangle (194,-24);\end{scope}
\node[anchor=west,font=\fontsize{7.7}{8.6}\selectfont\bfseries,text=cardblue] at (5,-8) {(e) E\quad Native semicontinuous domains};
\node[anchor=west,font=\fontsize{7.3}{8.2}\selectfont] at (5,-18) {\textbf{Identifier:} T06\qquad\textbf{Case:} T06\_010};
\node[anchor=north west,text width=182pt,align=left,font=\fontsize{7.4}{8.5}\selectfont] (expandedbody4) at (6,-29) {%
\begin{minipage}{182pt}\raggedright\setlength{\parindent}{0pt}\setlength{\parskip}{0pt}
\expandedfield{Core idea}{Encode zero-or-positive production directly with a native semicontinuous domain, replacing explicit activation binaries and their linking inequalities.}
\expandedfield{Applicable patterns}{Laboratory reagent batches have positive minimum and finite maximum sizes; activation appears only in quantity links, without separate costs.}
\expandedfield{Inefficiency symptoms}{Each product-shift pair adds one binary and two linking rows to the model.}
\expandedfield{Expert actions}{Declare production semicontinuous with the same bounds; preserve inventory balances, shared capacities, and production and holding costs.}
\textbf{Mathematical example}\par\vspace{3pt}
\begingroup\setlength{\fboxsep}{1pt}%
\colorbox{cardbefore}{\begin{minipage}[c][44pt][c]{88pt}\centering
{\fontsize{7.2}{8.2}\selectfont\bfseries\color{cardred}Ordinary reference}\par\vspace{3pt}
{\expandedmath\color{cardred}$\begin{array}{c}L_{pt}y_{pt}\le q_{pt}\le U_{pt}y_{pt},\\y_{pt}\in\{0,1\}\end{array}$}\end{minipage}}%
\hfill%
\colorbox{cardafter}{\begin{minipage}[c][44pt][c]{88pt}\centering
{\fontsize{7.2}{8.2}\selectfont\bfseries\color{cardgreen}Expert reference}\par\vspace{3pt}
{\expandedmath\color{cardgreen}$\begin{array}{c}q_{pt}\in\{0\}\cup[L_{pt},U_{pt}]\\\text{(native semicontinuous)}\end{array}$}\end{minipage}}\endgroup\par\vspace{3pt}
Here $q_{pt}$ is production and $y_{pt}$ activation, with $0<L_{pt}\le U_{pt}$. Both minimize production plus inventory cost. The zero-or-interval domain retains the discrete decision.\par\vspace{2pt}
\expandedfield{Benchmark guidance}{Build both representations on identical data; verify feasibility and objective agreement, then compare variable counts, solver work, and \texttt{Runtime}.}
\end{minipage}};
\path let \p1=(expandedbody4.south),\p2=(expandedbody4.north) in \pgfextra{\typeout{CARD-BODY-4-HEIGHT=\the\dimexpr\y2-\y1\relax}};
\draw[gray!35,line width=.4pt] (6,-268)--(188,-268);
\node[anchor=west,font=\fontsize{7.0}{8.0}\selectfont] at (6,-277) {\textbf{\texttt{Runtime}:} \textcolor{cardred}{411.77\,s}\ $\rightarrow$\ \textcolor{cardgreen}{31.68\,s}\qquad both OPTIMAL};
\end{scope}
\begin{scope}[shift={(202,0)}]
\path[draw=cardline,line width=.5pt,fill=white,rounded corners=3pt] (0,0) rectangle (194,-284);
\begin{scope}\clip[rounded corners=3pt] (.3,-.3) rectangle (193.7,-283.7);
\fill[cardhead] (0,0) rectangle (194,-24);\end{scope}
\node[anchor=west,font=\fontsize{7.7}{8.6}\selectfont\bfseries,text=cardblue] at (5,-8) {(f) F\quad Convex piecewise-linear costs};
\node[anchor=west,font=\fontsize{7.3}{8.2}\selectfont] at (5,-18) {\textbf{Identifier:} T09\qquad\textbf{Case:} T09\_008};
\node[anchor=north west,text width=182pt,align=left,font=\fontsize{7.4}{8.5}\selectfont] (expandedbody5) at (6,-29) {%
\begin{minipage}{182pt}\raggedright\setlength{\parindent}{0pt}\setlength{\parskip}{0pt}
\expandedfield{Core idea}{Represent convex piecewise-linear costs by supporting lines, removing discrete segment choices while preserving the minimum cost.}
\expandedfield{Applicable patterns}{Allocations have four ordered segments, nonnegative nondecreasing slopes, group quantity requirements, and shared resource limits.}
\expandedfield{Inefficiency symptoms}{Binary filling indicators introduce unnecessary integrality and enlarge the model for a convex cost function.}
\expandedfield{Expert actions}{Replace incremental quantities and filling indicators with one bounded quantity and one epigraph cost per unit.}
\textbf{Mathematical example}\par\vspace{3pt}
\begingroup\setlength{\fboxsep}{1pt}%
\colorbox{cardbefore}{\begin{minipage}[c][44pt][c]{88pt}\centering
{\fontsize{7.2}{8.2}\selectfont\bfseries\color{cardred}Ordinary reference}\par\vspace{3pt}
{\expandedmath\color{cardred}$\begin{array}{c}x=\sum_k y_k,\\y_k\ge\Delta_kz_k,\\y_{k+1}\le\Delta_{k+1}z_k\end{array}$}\end{minipage}}%
\hfill%
\colorbox{cardafter}{\begin{minipage}[c][44pt][c]{88pt}\centering
{\fontsize{7.2}{8.2}\selectfont\bfseries\color{cardgreen}Expert reference}\par\vspace{3pt}
{\expandedmath\color{cardgreen}$\begin{array}{c}c\ge m_kx+\beta_k\quad(\forall k),\\0\le x\le b_4\end{array}$}\end{minipage}}\endgroup\par\vspace{3pt}
For each unit, $\Delta_k=b_{k+1}-b_k$, $\beta_k=F(b_k)-m_kb_k$, and $0\le y_k\le\Delta_k$. Ordinary cost is $\sum_km_ky_k$; binary $z_k$ links segments. Minimize summed costs; $k=0,\ldots,3$, with links for $k<3$.\par\vspace{2pt}
\expandedfield{Benchmark guidance}{Preserve group requirements and resource capacities; verify convexity and objective agreement, then compare binary counts, matrix size, and solver work.}
\end{minipage}};
\path let \p1=(expandedbody5.south),\p2=(expandedbody5.north) in \pgfextra{\typeout{CARD-BODY-5-HEIGHT=\the\dimexpr\y2-\y1\relax}};
\draw[gray!35,line width=.4pt] (6,-268)--(188,-268);
\node[anchor=west,font=\fontsize{7.0}{8.0}\selectfont] at (6,-277) {\textbf{\texttt{Runtime}:} \textcolor{cardred}{378.62\,s}\ $\rightarrow$\ \textcolor{cardgreen}{42.75\,s}\qquad both OPTIMAL};
\end{scope}
\begin{scope}[shift={(0,-294)}]
\path[draw=cardline,line width=.5pt,fill=white,rounded corners=3pt] (0,0) rectangle (194,-284);
\begin{scope}\clip[rounded corners=3pt] (.3,-.3) rectangle (193.7,-283.7);
\fill[cardhead] (0,0) rectangle (194,-24);\end{scope}
\node[anchor=west,font=\fontsize{7.7}{8.6}\selectfont\bfseries,text=cardblue] at (5,-8) {(g) G\quad Lagrangian resource dualization};
\node[anchor=west,font=\fontsize{7.3}{8.2}\selectfont] at (5,-18) {\textbf{Identifier:} T13\qquad\textbf{Case:} T13\_006};
\node[anchor=north west,text width=182pt,align=left,font=\fontsize{7.4}{8.5}\selectfont] (expandedbody6) at (6,-29) {%
\begin{minipage}{182pt}\raggedright\setlength{\parindent}{0pt}\setlength{\parskip}{0pt}
\expandedfield{Core idea}{Price the shared labor resource to separate technology blocks, then express their adjusted values through an exact LP dual.}
\expandedfield{Applicable patterns}{Technology intensities form continuous simplices coupled by one labor budget; the primal LP must be feasible with finite optimum.}
\expandedfield{Inefficiency symptoms}{A monolithic model retains many intensity variables despite limited coupling between blocks.}
\expandedfield{Expert actions}{Dualize the labor limit with one nonnegative price and represent each block value using a free variable.}
\textbf{Mathematical example}\par\vspace{3pt}
\begingroup\setlength{\fboxsep}{1pt}%
\colorbox{cardbefore}{\begin{minipage}[c][44pt][c]{88pt}\centering
{\fontsize{7.2}{8.2}\selectfont\bfseries\color{cardred}GPT-5.5}\par\vspace{3pt}
{\expandedmath\color{cardred}$\begin{array}{c}\max_{x\ge0}\ \sum_{i,k}v_{ik}x_{ik}\\\sum_k x_{ik}=1\ (\forall i),\\\sum_{i,k}r_{ik}x_{ik}\le R\end{array}$}\end{minipage}}%
\hfill%
\colorbox{cardafter}{\begin{minipage}[c][44pt][c]{88pt}\centering
{\fontsize{7.2}{8.2}\selectfont\bfseries\color{cardgreen}Expert reference}\par\vspace{3pt}
{\expandedmath\color{cardgreen}$\begin{array}{c}\min_{\lambda,u}\ R\lambda+\sum_i u_i\\u_i+r_{ik}\lambda\ge v_{ik}\ (\forall i,k),\\\lambda\ge0,\quad u_i\in\mathbb R\end{array}$}\end{minipage}}\endgroup\par\vspace{3pt}
$v_{ik},r_{ik}$ denote value and labor, and $R$ the budget. At price $\lambda$, each block maximizes $v_{ik}-\lambda r_{ik}$; minimizing $R\lambda+\sum_i u_i$ gives the primal optimum.\par\vspace{2pt}
\expandedfield{Benchmark guidance}{Use identical values, labor requirements, and budget; verify strong duality and objective agreement, recover requested mixing decisions, then compare model size and solver work.}
\end{minipage}};
\path let \p1=(expandedbody6.south),\p2=(expandedbody6.north) in \pgfextra{\typeout{CARD-BODY-6-HEIGHT=\the\dimexpr\y2-\y1\relax}};
\draw[gray!35,line width=.4pt] (6,-268)--(188,-268);
\node[anchor=west,font=\fontsize{7.0}{8.0}\selectfont] at (6,-277) {\textbf{\texttt{Runtime}:} \textcolor{cardred}{1,048.04\,s}\ $\rightarrow$\ \textcolor{cardgreen}{4.44\,s}\qquad both OPTIMAL};
\end{scope}
\begin{scope}[shift={(202,-294)}]
\path[draw=cardline,line width=.5pt,fill=white,rounded corners=3pt] (0,0) rectangle (194,-284);
\begin{scope}\clip[rounded corners=3pt] (.3,-.3) rectangle (193.7,-283.7);
\fill[cardhead] (0,0) rectangle (194,-24);\end{scope}
\node[anchor=west,font=\fontsize{7.7}{8.6}\selectfont\bfseries,text=cardblue] at (5,-8) {(h) H\quad Compact robust counterpart};
\node[anchor=west,font=\fontsize{7.3}{8.2}\selectfont] at (5,-18) {\textbf{Identifier:} T22\qquad\textbf{Case:} T22\_008};
\node[anchor=north west,text width=182pt,align=left,font=\fontsize{7.4}{8.5}\selectfont] (expandedbody7) at (6,-29) {%
\begin{minipage}{182pt}\raggedright\setlength{\parindent}{0pt}\setlength{\parskip}{0pt}
\expandedfield{Core idea}{Dualize each budgeted deviation problem to replace enumerated scenarios with a compact formulation of the robust feasible region.}
\expandedfield{Applicable patterns}{Production and deviations are nonnegative; each resource row has bounded uncertainty, with integer $\Gamma$ for the displayed subset enumeration.}
\expandedfield{Inefficiency symptoms}{Enumerating allowed deviation subsets creates combinatorially many constraints and increases construction workload.}
\expandedfield{Expert actions}{Dualize the uncertainty budget and component bounds with nonnegative prices; add the resulting counterpart to each resource row.}
\textbf{Mathematical example}\par\vspace{3pt}
\begingroup\setlength{\fboxsep}{1pt}%
\colorbox{cardbefore}{\begin{minipage}[c][44pt][c]{88pt}\centering
{\fontsize{7.2}{8.2}\selectfont\bfseries\color{cardred}Ordinary reference}\par\vspace{3pt}
{\expandedmath\color{cardred}$\begin{array}{c}a^Tx+\sum_{j\in Q}d_jx_j\le b\\\forall Q\subseteq S,\quad |Q|\le\Gamma\end{array}$}\end{minipage}}%
\hfill%
\colorbox{cardafter}{\begin{minipage}[c][44pt][c]{88pt}\centering
{\fontsize{7.2}{8.2}\selectfont\bfseries\color{cardgreen}Expert reference}\par\vspace{3pt}
{\expandedmath\color{cardgreen}$\begin{array}{c}a^Tx+\Gamma\mu+\sum_j\lambda_j\le b,\\\mu+\lambda_j\ge d_jx_j\ (\forall j\in S),\\\mu,\lambda_j\ge0\end{array}$}\end{minipage}}\endgroup\par\vspace{3pt}
For each row, $a,d,b$ give nominal coefficients, deviations, and capacity; $0\le z\le1$ and $\sum_{j\in S}z_j\le\Gamma$ define uncertainty. Preserve profit and other restrictions.\par\vspace{2pt}
\expandedfield{Benchmark guidance}{Use identical uncertainty sets, objectives, and production restrictions; check inner LP duality and robust-feasibility equivalence, then compare rows, construction cost, and \texttt{Runtime}.}
\end{minipage}};
\path let \p1=(expandedbody7.south),\p2=(expandedbody7.north) in \pgfextra{\typeout{CARD-BODY-7-HEIGHT=\the\dimexpr\y2-\y1\relax}};
\draw[gray!35,line width=.4pt] (6,-268)--(188,-268);
\node[anchor=west,font=\fontsize{7.0}{8.0}\selectfont] at (6,-277) {\textbf{\texttt{Runtime}:} \textcolor{cardred}{157.49\,s}\ $\rightarrow$\ \textcolor{cardgreen}{13.63\,s}\qquad both OPTIMAL};
\end{scope}
\end{tikzpicture}
\endgroup
\caption{\footnotesize \textbf{OptTips cards for families E--H.} The same seven-component schema continues with four further examples. Panel (g) retains the GPT-5.5--expert comparison from Figure~\ref{fig:introduction}; other panels use ordinary and expert references. Times report solver \texttt{Runtime} under the execution protocol in Appendix~\ref{app:measurement-detail}; panel (h) lacks historical input-hash verification.}
\label{fig:opttips-cards-eh}
\end{figure}
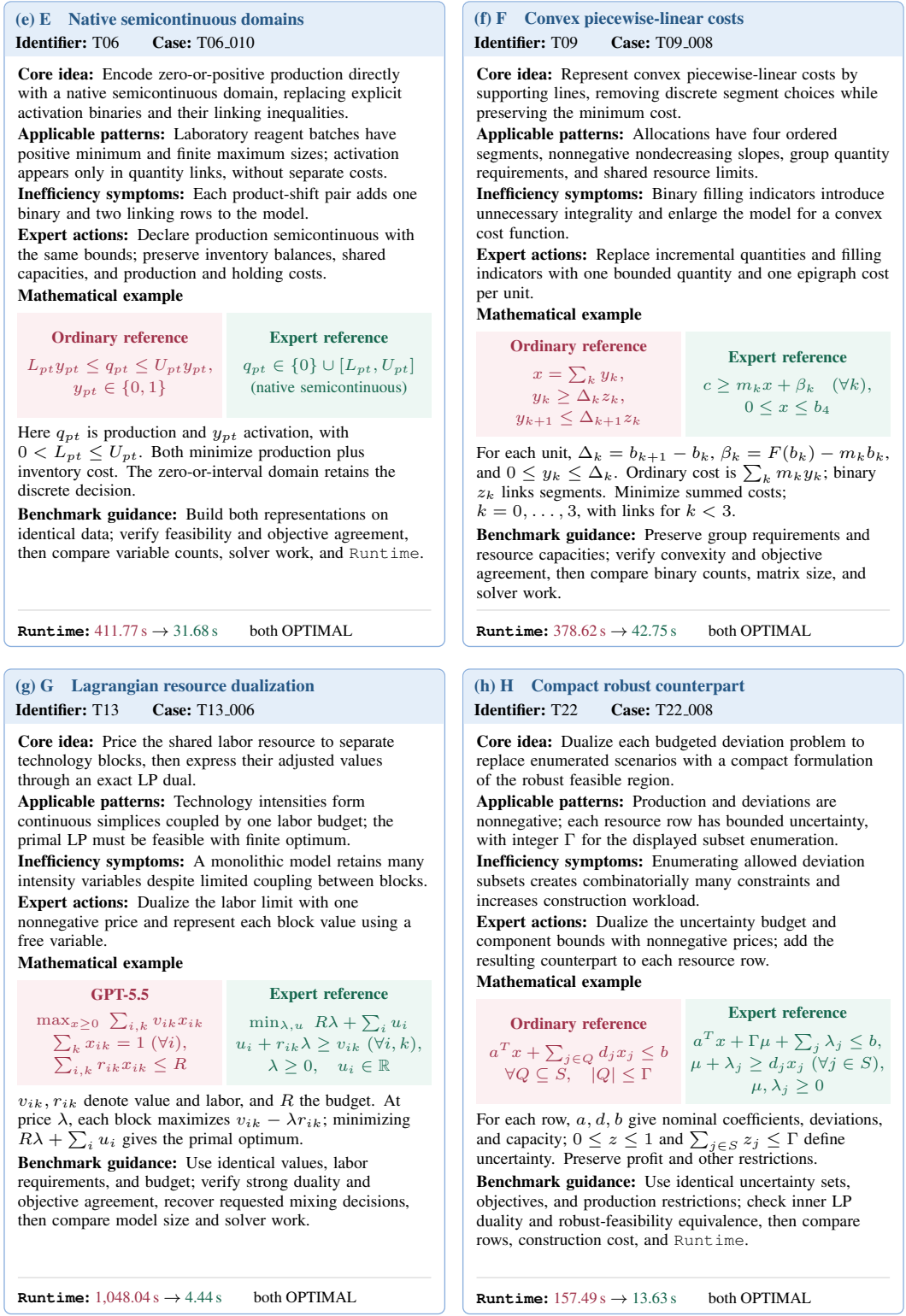


\noindent\emph{Future work: broader solver and algorithm coverage.} Evaluation could extend to additional solver interfaces and the complete iterative and factor-based procedures of T14, T47, and T50, at matched solution accuracy and with the same hidden-technique task interface. Some supplementary tasks already use constructs supported by Gurobi, including second-order-cone formulations (T11) and SOS, indicator, and piecewise-linear formulations (T29) \citep{gurobi_constraints}; their exclusion from the current evaluation does not imply solver incompatibility.

Broader variants include T47's nuclear-norm SDP with positive-semidefinite constraints \citep{mosek_modeling_cookbook} and T29's interval \texttt{NoOverlap} in CP-SAT \citep{ortools_job_shop}. Evaluating T14's ADMM requires its full multiplier-update iterations and residual-based stopping \citep{boyd2011admm}.


\par\endgroup

\begin{table}[!htbp]
\subsection{Complete Technique Inventory}
\label{app:opttips-inventory}
\centering\small
\setlength{\tabcolsep}{3.5pt}
\renewcommand{\arraystretch}{1.08}
\caption{All 50 OptTips techniques. Family codes refer to Table~\ref{tab:opttips-families}. The catalog includes formulation, algorithmic, and solver-aware choices; full cards are available in the public repository.}
\label{tab:opttips-catalog}
\resizebox{\linewidth}{!}{%
\begin{tabular}{@{}l l c@{\hspace{18pt}}l l c@{}}
\toprule
ID & Technique & Family & ID & Technique & Family \\
\cmidrule(r{9pt}){1-3}\cmidrule(l{9pt}){4-6}
T01 & Equality/inequality equivalent transformation & A & T26 & Complementarity and KKT/MPEC modeling & E \\
T02 & Auxiliary-variable epigraph/hypograph modeling & A & T27 & Multi-objective and lexicographic optimization & H \\
T03 & Bound tightening and scaling & A & T28 & Penalty, slack, and feasibility-relaxation modeling & H \\
T04 & LP relaxation of IP/MIP & E & T29 & Solver-native modeling constructs & H \\
T05 & Big-$M$ modeling & E & T30 & Warm starts and incumbent injection & H \\
T06 & Indicator, semicontinuous, and SOS constraints & E & T31 & Algorithm--structure matching & H \\
T07 & Valid inequalities and cutting planes & B & T32 & Solver-log-guided diagnosis & H \\
T08 & Variable substitution and product linearization & F & T33 & Standard/canonical-form transformation & A \\
T09 & Piecewise-linear modeling & F & T34 & Slack, surplus, and artificial variables & A \\
T10 & Convex-hull/envelope reformulation and convexification & B & T35 & Variable elimination and dimension reduction & A \\
T11 & Conic reformulation & F & T36 & Convexity recognition and DCP modeling & F \\
T12 & Dual transformation & C & T37 & Smooth--nonsmooth composite modeling & F \\
T13 & Lagrangian relaxation & G & T38 & Convex conjugacy and Fenchel duality & F \\
T14 & Augmented Lagrangian and ADMM & G & T39 & Subgradient, normal-cone, and optimality-condition modeling & F \\
T15 & Block decomposition & G & T40 & Projection and proximal-operator modeling & F \\
T16 & Benders decomposition & G & T41 & Variational-inequality and equilibrium-constraint modeling & C \\
T17 & Dantzig--Wolfe decomposition and column generation & G & T42 & Sensitivity and parametric optimization & C \\
T18 & Network-flow and matching formulations & D & T43 & Dynamic programming and Bellman recursion & G \\
T19 & Covering, partitioning, packing, and knapsack formulations & D & T44 & Transportation and network-simplex structure & D \\
T20 & Scheduling and temporal formulations & D & T45 & Constraint qualifications and KKT validity & C \\
T21 & Routing and path formulations & D & T46 & Interior-point-friendly modeling & F \\
T22 & Robust counterpart reformulation & H & T47 & Matrix-variable and low-rank modeling & D \\
T23 & Stochastic programming and scenario decomposition & H & T48 & Cover, clique, subtour, and capacity cuts & B \\
T24 & Preprocessing and dominance elimination & B & T49 & Linking constraints and activation strengthening & B \\
T25 & Symmetry breaking & D & T50 & Tensor-variable and low-rank modeling & D \\
\bottomrule
\end{tabular}%
}
\end{table}

\end{document}